\documentclass[english,11pt]{article}
\pdfoutput=1
\usepackage[sort,comma]{natbib}
\usepackage[pagebackref=false,colorlinks, linktocpage=true]{hyperref}
\hypersetup{pdffitwindow=true,linkcolor=blue,citecolor=blue,urlcolor=blue}
\usepackage{amsmath}
\usepackage{bbm}
\usepackage{multirow}

\usepackage{amsfonts}
\usepackage{mathdots}
\usepackage{amsthm}

\usepackage{multirow}
\usepackage{centernot}
\usepackage{todonotes}
\usepackage[ruled,linesnumbered]{algorithm2e}
\usepackage{float}
\usepackage{optidef}
\usepackage{dsfont}

\usepackage{lmodern}
\usepackage{amssymb,amsmath}
\usepackage{ifxetex,ifluatex}
\usepackage{fixltx2e} 
\ifnum 0\ifxetex 1\fi\ifluatex 1\fi=0 
  \usepackage[T1]{fontenc}
  \usepackage[utf8]{inputenc}
\else 
  \ifxetex
    \usepackage{mathspec}
  \else
    \usepackage{fontspec}
  \fi
  \defaultfontfeatures{Ligatures=TeX,Scale=MatchLowercase}
\fi
\IfFileExists{upquote.sty}{\usepackage{upquote}}{}
\IfFileExists{microtype.sty}{%
\usepackage{microtype}
\UseMicrotypeSet[protrusion]{basicmath} 
}{}
\usepackage{hyperref}
\hypersetup{unicode=true,
            pdftitle={pcaNet example output},
            pdfborder={0 0 0},
            breaklinks=true}
\usepackage[margin=1in]{geometry}
\usepackage{color}
\usepackage{fancyvrb}

\DefineVerbatimEnvironment{Highlighting}{Verbatim}{commandchars=\\\{\}}
\usepackage{framed}

\SetKwInOut{Input}{Input}
\SetKwInOut{Output}{Output\,}
\SetKwInOut{Data}{Data}

\definecolor{shadecolor}{RGB}{248,248,248}

\usepackage[font={footnotesize,it}, sc, bf]{caption}
\renewenvironment{abstract}{%
  \noindent\textbf\abstractname .\hspace{1pt}
}{
  \endlist \par\bigskip\bigskip
}

\usepackage{lastpage}
\RequirePackage[hyperpageref]{backref}
\renewcommand*{\backref}[1]{} 
\renewcommand*{\backrefalt}[4]{
    \ifcase #1
       No referred.
    \or
       \emph{Referred to on page #2.}
    \else
       \emph{Referred to on pages #2.}
    \fi}

\usepackage[shortlabels]{enumitem}

\usepackage[titletoc,title]{appendix}

\binoppenalty=\maxdimen
\relpenalty=\maxdimen

\mathtoolsset{showonlyrefs}

\usetikzlibrary{bayesnet}

\usepackage{xcolor}
\definecolor{cambridgebluecore}{RGB}{0, 176, 185}
\definecolor{cambridgebluedark}{RGB}{17, 94, 103}
\definecolor{cambridgebluelight}{RGB}{133, 176, 154}

\definecolor{cambridgeredcore}{RGB}{213, 0, 50}
\definecolor{cambridgereddark}{RGB}{138, 21, 56}
\definecolor{cambridgeredlight}{RGB}{232, 156, 174}

\definecolor{cambridgeblue2core}{RGB}{0, 114, 206}
\definecolor{cambridgeblue2dark}{RGB}{0, 60, 113}
\definecolor{cambridgeblue2light}{RGB}{108, 172, 228}

\definecolor{cambridgeorangecore}{RGB}{232, 119, 34}
\definecolor{cambridgeorangedark}{RGB}{190, 77, 0}
\definecolor{cambridgeorangelight}{RGB}{241, 190, 72}

\definecolor{cambridgegreencore}{RGB}{100, 167, 111}
\definecolor{cambridgegreendark}{RGB}{78, 91, 49}
\definecolor{cambridgegreenlight}{RGB}{183,	191, 16}

\usepackage{amsthm}
\usepackage{thmtools}

\usepackage[]{graphicx}
\graphicspath{{../figures/}}
\usepackage{subcaption}
\usepackage{epstopdf}
\usepackage{tikz}
\tikzset{font={\fontsize{9pt}{8}\selectfont}}
\usetikzlibrary{positioning}

\title{Multiple kernel learning for integrative consensus clustering of 'omic datasets}
\author{Alessandra Cabassi, Paul D. W. Kirk}

\begin{document}

\begin{center}
{\LARGE\bf Multiple kernel learning for integrative consensus clustering of 'omic datasets}
\end{center}
\medskip
\begin{center}
{\large Alessandra Cabassi$^{1}$ and Paul D. W. Kirk$^{1,2}$ \\[15pt]
\emph{$^{1}$MRC Biostatistics Unit}\\
\emph{$^{2}$Cambridge Institute of Therapeutic Immunology \& Infectious Disease\\
University of Cambridge, U.K.}\\
}
\end{center}

\bigskip

\begin{center}
Preprint, \today
\end{center}
\bigskip\bigskip

\begin{abstract}\newline
\textbf{Summary:} 
Diverse applications -- particularly in tumour subtyping -- have demonstrated the importance of integrative clustering techniques for combining information from multiple 
data sources.
Cluster Of Clusters Analysis (COCA) is one such approach that has been widely applied in the context of tumour subtyping. However, the properties of COCA have never been systematically explored, and its robustness to the inclusion of noisy datasets is unclear.\\
We rigorously benchmark COCA, and present Kernel Learning Integrative Clustering (KLIC) as an alternative strategy.  KLIC frames the challenge of combining clustering structures as a multiple kernel learning problem, in which different datasets each provide a {\em weighted} contribution to the final clustering.  This allows the contribution of noisy datasets to be down-weighted relative to more informative datasets.  
We compare the performances of KLIC and COCA in a variety of situations through simulation studies. We also present the output of KLIC and COCA in real data applications to cancer subtyping and  transcriptional module discovery.\\
\textbf{Availability:} R packages  \textit{klic} and \textit{coca} are available on the Comprehensive R Archive Network.\\
\textbf{Contact:} \href{alessandra.cabassi@mrc-bsu.cam.ac.uk}{alessandra.cabassi@mrc-bsu.cam.ac.uk}, \href{paul.kirk@mrc-bsu.cam.ac.uk}{paul.kirk@mrc-bsu.cam.ac.uk}\\
\end{abstract}

\section{Introduction}
Thanks to technological advances, both the availability and the diversity of 'omic datasets have hugely increased in recent years \citep{Manzoni2018}. These datasets provide information on multiple levels of biological systems, going from the genomic and epigenomic level, to gene and protein expression level, up to the metabolomic level, accompanied by phenotype information. Many publications have highlighted the importance of integrating information from diverse 'omic datasets in order to provide novel biomedical insight. For example, numerous studies by The Cancer Genome Atlas (TCGA) consortium have demonstrated the value of combining multiple 'omic datasets in order to define cancer subtypes (see e.g. \citealp{TCGA2011, TCGA2012}).

Many existing statistical and computational tools have been applied to this  problem and many others have been developed specifically for this. One of the first statistical methods applied to integrative clustering for cancer subtypes was {\em iCluster} \citep{Shen2009integrative, Shen2013sparse}. iCluster finds a partitioning of the tumours into different subtypes by projecting the available datasets onto a common latent space, maximising the correlation between data types.  Another statistical method for integrative clustering is {\em Multiple Dataset Integration} \citep[MDI; see][]{Kirk2012bayesian,Mason2016}. It is based on Dirichlet-multinomial mixture models in which the allocation of observations to clusters in one dataset influences the allocation of observations in another, while allowing different datasets to have different numbers of clusters. 
Similarly, {\em Bayesian Consensus Clustering} (BCC) is based on a Dirichlet mixture model that assigns a different probability model to each dataset. Again, samples belong to different partitions, each given by a different data type, but here they also adhere loosely to an overall clustering \citep{Lock2013bayesian}. 
More recently, \cite{Gabasova2017} developed {\em Clusternomics}, a mixture model over all possible combinations of cluster assignments on the level of individual datasets that allows to model different degrees of dependence between clusters across datasets.

Integrative clustering methods can be broadly classified as either {\em joint modelling} or {\em two-step} approaches. The former simultaneously consider all datasets together (e.g. MDI or BCC).  The latter, which we consider here, are composed of two steps: first, the clustering structure in each dataset is analysed independently; then an integration step is performed to find a common clustering structure that combines the individual ones. These approaches have sometimes also been referred to as {\em sequential analysis} methods \citep{Kristensen2014}. 

Cluster Of Clusters Analysis (COCA) is a particular two-step approach, which has grown in popularity since its first introduction in \cite{TCGA2012}. As we explain in Section \ref{sec:coca}, COCA proceeds by first clustering each of the datasets separately, and then building a binary matrix that encodes the cluster allocations of each observation in each dataset.
This binary matrix is then used as the input to a consensus clustering algorithm \citep{Monti2003,Wilkerson2010}, which returns a single, global clustering structure, together with an assessment of its stability.  The idea is that this global clustering structure both combines and summarises the clustering structures of the individual datasets. 
Despite its widespread use, to the best of our knowledge the COCA algorithm has never previously been systematically explored. In what follows, we elucidate the algorithm underlying COCA, and highlight some of its limitations.  We show that one key limitation is that the combination of the clustering structures from each dataset is {\em unweighted}, making the output of the algorithm sensitive to the inclusion of poor quality datasets.

An alternative class of approaches for integrating multiple 'omic datasets is provided by those based on {\em kernel methods} \citep[see, among others,][for 'omic dataset applications]{Lanckriet2004,Lewis2006}.  In these, a kernel function (which defines similarities between different units of observation) is associated with each dataset.  These may be straightforwardly combined in order to define an overall similarity between different units of observation, which incorporates similarity information from each dataset.  Determining an optimal (weighted) combination of kernels is known as {\em multiple kernel learning} (MKL); see, for example, \citet{Lanckriet2004b,Bach2004,Yu2010,Gonen2011,Wang2017,Strauss2019}.  A challenge associated with these approaches is how best to define the kernel function(s), for which there may be many choices.
  
Here we combine ideas from COCA and MKL in order to propose a new Kernel Learning Integrative Clustering (KLIC) method that addresses the limitations of COCA (Section \ref{sec:kernelmethods-kic}).  Key to our approach is the result that the {\em consensus matrix} returned by consensus clustering is a valid kernel matrix (Section \ref{sec:indentifying-co-clustering-matrices-as-kernels}).  This insight allows us to make use of the full range of multiple kernel learning approaches in order to combine consensus matrices derived from different 'omic datasets.  We perform simulation studies to illustrate our proposed approach and compare it to COCA. 
Finally, we show how KLIC and COCA compare in two practical applications: multiplatform tumour subtyping, where the goal is to stratify patients, and transcriptional module discovery, where genes are the statistical observations that we want to cluster.

\section{Methods}
\subsection{Cluster Of Clusters Analysis}
\label{sec:coca}

\emph{Cluster Of Clusters Analysis} (COCA; \citealp{TCGA2012}) is an integrative clustering method that was first introduced in a breast cancer study by \cite{TCGA2012} and quickly became a popular tool in cancer studies (see e.g. \citealp{Hoadley2014} and \citealp{Aure2017}). It makes use of \emph{Consensus Clustering} (CC; \citealp{Monti2003}), an algorithm that was originally developed to assess the stability of the clusters obtained with any clustering algorithm. 

\subsubsection{Consensus clustering}
We recall here the main features of CC in order to be able to explain the functioning of COCA. As originally formulated, CC is an approach for assessing the robustness of the clustering structure present in a single dataset \citep{Monti2003,Wilkerson2010}. The idea behind CC is that, by resampling multiple times the items that we want to cluster and then applying the same clustering algorithm to each of the subsets of items, we assess the robustness of the clustering structure that the algorithm detects, both to perturbations of the data and (where relevant) to the stochasticity of the clustering algorithm. To do this, CC makes use of the concepts of co-clustering matrix and consensus matrix, which we recall here.

Given a set of items $ X = [\boldsymbol{x}_1, \dots, \boldsymbol{x}_N]$ that we seek to cluster and a clustering $\boldsymbol{c}=[c_1, \dots, c_N]$ such that $c_i$ is the label of the cluster to which item $\boldsymbol{x}_i$ has been assigned, the corresponding \emph{co-clustering matrix} (or \emph{connectivity matrix}) is an $N \times N$  matrix $C$ such that the $ij$-th element $C_{ij}$ is equal to one if $c_i = c_j$, and zero otherwise. 
Let $X^{(1)}, \dots, X^{(H)}$ be a list of perturbed datasets obtained by resampling subsets of items and/or covariates from the original dataset $X$.  If $I^{(h)}$ is the subset of the indices of the observations $I = \{1,2,\dots, N\}$ present in $X^{(h)}$, then the co-clustering matrix has $ij$-th element equal to one if $i,j \in I^{(h)}$ and $c_i = c_j$, zero otherwise.
We denote by $C^{(h)}$ the co-clustering matrix corresponding to dataset $X^{(h)}$ where the items have been assigned to $K$ classes using a clustering algorithm. 

The \emph{consensus matrix} $\Delta^K$ is an $N \times N$ matrix with elements
	\begin{equation}
	\Delta^K_{ij} = \frac{\sum_{h = 1}^H C^{(h)}_{ij}}{\min\left\lbrace 1, \sum_{h=1}^{H}\mathbb{I}^{(h)}_{ij}\right\rbrace}
	\end{equation}
	where $\mathbb{I}^{(h)}_{ij} = 1$ if both items $i$ and $j$ are present in dataset $X^{(h)}$.

Thus, CC performs multiple runs of a (stochastic) clustering algorithm (e.g. $k$-means, hierarchical clustering, etc.) to assess the stability of the discovered clusters, with the consensus matrix providing a convenient summary of the CC analysis. If all the elements of the consensus matrix are close to either one or zero, this means that every pair of items is either almost always assigned to the same cluster, or almost always assigned to different clusters. Therefore, consensus matrices with all the elements close to either zero or one indicate stable clusters.  
In the framework of consensus clustering, these matrices can also be used to determine the number of clusters, by computing and comparing the consensus matrices $\Delta^K$ for a range of numbers of clusters $\mathcal{K} = \{K_{\min},\dots, K_{\max}\}$ of interest and then pick the value of $K$ that gives the consensus matrix with the greater proportion of elements close to either zero or one  \citep{Monti2003}.

\subsubsection{COCA}
In contrast to consensus clustering (which we emphasise is concerned with assessing clustering stability when analysing a single dataset), the main goal of COCA is to summarise the clusterings found in {\em different} 'omic datasets, by identifying a ``global'' clustering across the datasets that is intended to summarise the clustering structures identified in each of the individual datasets.
In the first step, a clustering $\boldsymbol{c}^m$ is produced independently for each dataset $X_m$, $m = 1, \dots, M$, each with a different number of clusters $K_m$. We define $\bar{K} = \sum_{m=1}^{M} K_m$. Then, the clusters are summarised into a Matrix Of Clusters (MOC) of size $\bar{K} \times N$, with elements
\begin{equation}
\text{MOC}_{n,m_k} = \left\lbrace
\begin{array}{ll}
1 & \text{if } c^m_n = m_k,\\
0 & \textit{otherwise}.
\end{array}
\right.
\end{equation}
where by $m_k$ we denote the $k$-th cluster in dataset $m$, $k=1, \dots, K_m$ and $m = 1, \dots, M$.
The MOC matrix is then used as input to CC together with a fixed global number of clusters $K$.
The resulting consensus matrix is then used as the similarity matrix for a hierarchical clustering method (or any other distance-based clustering algorithm).

The global number of clusters $K$ is not always known. In \cite{TCGA2012}, where COCA was introduced, the global number of clusters was chosen as in \cite{Monti2003}, as explained above: CC was performed with different values of $K$ and then the one that gave the ``best'' consensus matrices were considered. Instead, \cite{Aure2017} suggest to choose the value of $K$ that maximises the average silhouette \citep{Rousseeuw1987} of the final clustering, since this was found to give more sensible results.

Since the construction of the MOC matrix just requires the cluster allocations, COCA has the advantage of allowing clusterings derived from different sources to be combined, even if the original datasets are unavailable or unwieldy. However, this method is unweighted, since all the clusters found in the first step have the same influence on the final clustering.
Moreover, the objective function that is optimised by the algorithm is unclear.

In what follows, we describe an alternative way of performing integrative clustering, that takes into account not only the clusterings in each dataset, but also the information about the similarities between items that are extracted from different types of data. Additionally, the new method allows weights to be given to each source of information, according to how useful it is for defining the final clustering.

\subsection{Kernel learning integrative clustering}
\label{sec:kernelmethods-kic}
Before introducing the new methodology, we recall the main principles behind the methods that we use to combine similarity matrices.
\subsubsection{Kernel methods}
\label{sec:kernel-methods}
Using kernel methods, it is possible to model non-linear relationships between the data points with a low computational complexity, thanks to the so-called \emph{kernel trick.}
For this reason, these have been widely used to extend many traditional algorithms to the non-linear framework, such as PCA \citep{Scholkopf1998}, linear discriminant analysis \citep{Mika1999, Roth2000, Baudat2000} and ridge regression \citep{Friedman2001, Shawe2004}.

A \emph{positive definite kernel} or, more simply, a \emph{kernel} $\delta$ is a symmetric map $\delta : \mathcal{X} \times \mathcal{X} \to \mathbb{R}$ for which for all $x_1, x_2, \dots, x_N \in \mathcal{X}$, the matrix $\Delta$ with entries $\Delta_{ij} = \delta(x_i, x_j)$ is positive semi-definite. The matrix $\Delta$ is called the \emph{kernel matrix} or \emph{Gram matrix}.
Kernel methods proceed by embedding the observations into a higher-dimensional feature space $\mathcal{H}$ endowed with an inner product $\langle \cdot, \cdot \rangle_{\mathcal{H}}$ and induced norm $\lVert \cdot\rVert_{\mathcal{H}}$, making use of a map $\phi: \mathcal{X} \to \mathcal{H}$.
Using Mercer's theorem, it can be shown that for any positive semi-definite kernel function, $\delta$, there exists a corresponding feature map, $\phi: \mathcal{X} \to \mathcal{H}$ (see e.g. \citealp{Vapnik1998}). That is, for each kernel $\delta$, there exists a feature map $\phi$ taking value in some inner product space $\mathcal{H}$ such that $\delta(x,x') = \langle \phi(x), \phi(x') \rangle_{\mathcal{H}}$. In practice, it is therefore often sufficient to specify a positive semi-definite kernel matrix, $\Delta$, in order to allow us to apply kernel methods such as those presented in the following sections. For a more detailed discussion of kernel methods, see e.g. \cite{Shawe2004}.
\subsubsection{Localised multiple kernel $k$-means clustering}
\label{sec:multiple-kernel-kmeans}
Kernel $k$-means is a generalisation of the $k$-means algorithm of \citet{Steinhaus1956} to the kernel framework \citep{Girolami2002}. The kernel trick is used to reformulate the problem of minimising the sum of squared distances between each point and the corresponding cluster centre (in the feature space) as a trace maximisation problem that only requires knowing the Gram matrix corresponding to the kernel of interest. Optimal cluster allocations can then be efficiently determined using kernel PCA.
More details on kernel $k$-means can be found in the Supplementary Material.

The clustering algorithm used here is the extension of the kernel $k$-means approach to multiple kernel learning \citep{Gonen2011} with sample-specific weights \citep{Gonen2014} aimed at removing sample-specific noise. We consider multiple datasets $X_1, \dots, X_M$ each with a different mapping function $\phi_m: \mathbb{R}^P \to \mathcal{H}_m$ and corresponding kernel $\delta_m(\boldsymbol{x}_i, \boldsymbol{x}_j) = \langle \phi_m(\boldsymbol{x}_i), \phi_m(\boldsymbol{x}_j) \rangle_{\mathcal{H}_m} $ and kernel matrix $\Delta_m$. Then, if we define $
\phi_{\Theta} (\boldsymbol{x}_i) = [\theta_{i1}\phi_1(\boldsymbol{x}_i)', \theta_{i2}\phi_2(\boldsymbol{x}_i)', \dots, \theta_{iM}\phi_M(\boldsymbol{x}_i)']'$,
where $\Theta \in \mathbb{R}^{N \times M}_{+}$ is a vector of kernel weights with elements $\theta_{im}$ such that $\sum_m \theta_{im} = 1$ and $\theta_{im} \geq 0$ for $i = 1, \dots, N$, the kernel function of this multiple feature problem is a convex sum of the single kernels:
\begin{align}
\delta_{\Theta}(\boldsymbol{x}_i, \boldsymbol{x}_j) & = \langle \phi_{\Theta}(\boldsymbol{x}_i), \phi_{\Theta}(\boldsymbol{x}_j) \rangle_{\mathcal{H}_m} = \sum_{m=1}^{M}  \theta_{im} \theta_{jm} \delta_m(\boldsymbol{x}_i, \boldsymbol{x}_j).
\end{align}
We denote the corresponding kernel matrix by $\Delta_{\Theta}$. The idea of localised multiple kernel $k$-means is to replace the Gram matrix used in kernel $k$-means by this weighted matrix.
The optimisation strategy proposed by \cite{Gonen2014} is based on the idea that, for some fixed vector of weights $\Theta$, this is a standard kernel $k$-means problem. Therefore, they develop a two-step optimisation strategy: (1) given a fixed vector of weights $\Theta$, solve the optimisation problem as in the case of one kernel, with kernel matrix given by $\delta_{\Theta}$ and then (2) minimise the objective function with respect to the kernel weights, keeping the assignment variables fixed. This is a convex quadratic programming (QP) problem that can be solved with any standard QP solver up to a moderate number of kernels $M$. 

\subsubsection{Identifying consensus matrices as kernels}
\label{sec:indentifying-co-clustering-matrices-as-kernels}
We prove that the consensus matrices defined in Section \ref{sec:coca} are positive semidefinite, and hence that they can be used as input for any kernel-based clustering method, including the integrative clustering method presented in the next section.
Given any $N \times N$ co-clustering matrix $C$, we can reorder the rows and columns to obtain a block-diagonal matrix with blocks $J_1, J_2, \dots, J_K$ where $K$ is the total number of clusters and $J_k$ is an $n_k \times n_k$ matrix of ones, with $n_k$ being the number of items in cluster $k$.
It is straightforward to show that the eigenvalues of a block-diagonal matrix are simply the eigenvalues of its blocks. Since each block is a matrix of ones, the eigenvalues of each block are nonnegative, and so any co-clustering matrix $C$ is positive semidefinite.
Moreover, given any set of $\lambda_m$, $m = 1, \dots, M$ nonnegative, and co-clustering matrices $C_m$, $m = 1, \dots, M$ , then $\sum_{m=1}^{M} \lambda_m C_m$ is positive semidefinite, because if $\lambda$ is a nonnegative scalar, and $C$ is positive semidefinite, then $\lambda C$ is also positive semidefinite and the sum of positive semidefinite matrices is a positive semidefinite matrix. Since every consensus matrix is of the form $\sum_m \lambda_m C_m$, we can conclude that any consensus matrix is positive semidefinite.

\subsubsection{Kernel Learning Integrative Clustering}
\label{sec:kic}
We recall from Section \ref{sec:kernel-methods} that any positive semidefinite matrix defines a feature map $\phi: \mathbb{R}^P \to \mathcal{H}$ and is therefore a valid kernel matrix.
The integrative clustering method that we introduce here is based on the idea that we can identify the consensus matrices produced by CC as kernels. That is, one can perform consensus clustering on each dataset to produce a consensus matrix $\Delta_m$ for each $m \in \{1, \dots, M\}$. This is a kernel $\Delta_m$, where the $ij$-th element corresponds to the similarity between items $i$ and $j$. Therefore, these matrices $\Delta_m$ can be combined through the (localised) multiple kernel $k$-means algorithm described in Section \ref{sec:multiple-kernel-kmeans}. 
This allows a weight to be obtained for each kernel, as well as a global clustering $\boldsymbol{c}$ of the items.
We note that this algorithm could also be applied using more than one similarity matrix per dataset, and also using kernel matrices other than (or in addition to) consensus matrices.
\section{Examples}
\label{ref:simulation-studies-coca-limitations}

\subsection{Simulated data}
\label{sec:synthetic-datasets}
To assess the KLIC algorithm described in Section \ref{sec:kic} and to compare it to COCA, we perform a range of simulation studies.
We generate several synthetic datasets, each composed of data belonging to six different clusters of equal size. 
Each dataset has total number of observations equal to 300. Each observation $\boldsymbol{x}_{n}^{(k)}$ is generated from a bivariate normal with mean $ks$ for each variable, where $k$ denotes the cluster to which the observation belongs and $s$ the separation level of the dataset. Higher values of $s$ give clearer clustering structures. The variance covariance matrix is the identity matrix.
\begin{figure}[h]
	\centering
	\includegraphics[width=.96\textwidth]{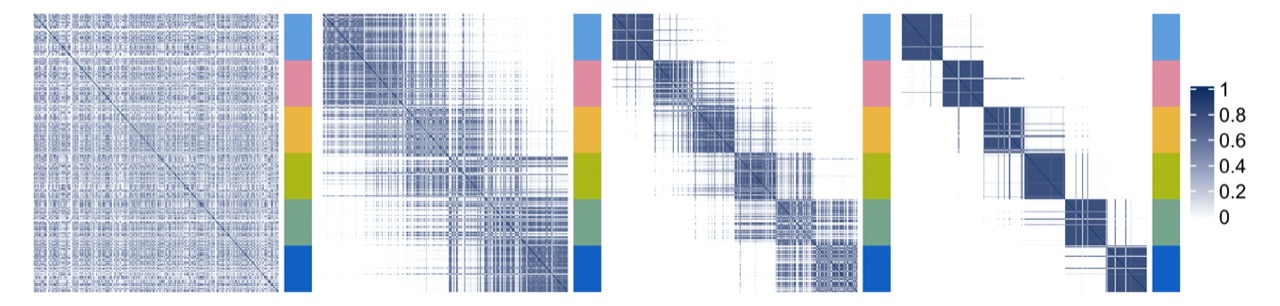}
	\caption{Consensus matrices of the synthetic data with different levels of noise, going from ``no cluster separability'' to ``high cluster separability''. Blue indicates high similarity. The colours of the bar to the right of each matrix indicate the cluster labels.}
	\label{fig:consensus-matrices-simulations}
\end{figure}

We consider the following settings:
\begin{enumerate}
	\item \emph{Similar datasets.} We generate four datasets that have the same clustering structure and cluster separability $s$. We denote the datasets by A, B, C, D. The goal of this experiment is to show that using localised kernel $k$-means on multiple consensus matrices leads to better results than those obtained using just one consensus matrix. To demonstrate how we may deal with irrelevant variables, we also repeat this experiment adding to each dataset 13 variables centred at zero that have no clustering structure, i.e.
\begin{equation}
	\mathbf{x}_1^{(k)}, \dots, \mathbf{x}_{50}^{(k)} \sim \mathcal{N}([ks, ks, \underbrace{0, \dots, 0}_{13}], I), \quad \forall k = 1, \dots, 6,
\end{equation}
where $I$ is the $15 \times 15$ identity matrix.
	\item \emph{Datasets with different levels of noise.} In this case we utilise four datasets that have the same clustering structure, but different levels of cluster separability $s$. We denote the datasets by 0 for ``no cluster separability'', 1 ``low cluster separability'', 2 ``medium cluster separability'', and 3 ``high cluster separability'' (Figure \ref{fig:consensus-matrices-simulations}). We use this example to show how the weights are allocated to each consensus matrix and why it is important to assign lower weights to datasets that are noisy or not relevant.
\end{enumerate}
We repeat each experiment 100 times.
For each synthetic dataset, we use consensus clustering to obtain the consensus matrices.
For simplicity, we set $K=6$. As for the clustering algorithm, we use $k$-means clustering with Euclidean distance, which we found to work well in practice. The Supplementary Material contains additional simulation settings. In particular, we consider a wide range of separability values for the setting with four similar datasets and the integration of datasets with nested clusters. Moreover, we perform a short sensitivity analysis of the choice or tuning options for the $k$-means algorithm.

\subsection{Multiplatform analysis of 12 cancer types}

\cite{Hoadley2014} performed a multiplatform integrative analysis of 3,527 tumour samples from 12 different tumour types, and used COCA to identify 11 integrated tumour subtypes. 
To do so, they applied different clustering algorithms to each data type separately: DNA copy number, DNA methylation, mRNA expression, microRNA expression, and protein expression. They then combined the five sets of clusters obtained in this way using COCA. The final clusters are highly correlated with the tissue-of-origin of each tumour sample, but some cancer types coalesce into the same clusters. The clusters obtained in this way were shown to be prognostic and to give independent information from the tissue-of-origin.

Here, we use the same data to try to replicate their analysis, and compare the clusters obtained with COCA to those obtained with KLIC.  To facilitate future analyses by other researchers, we have made available our scripts for processing and analysing these datasets using the freely available R statistical programming language \citep{R2020}, which include scripts that seek to replicate the original analysis of \cite{Hoadley2014}, {at \href{https://github.com/acabassi/klic-pancancer-analysis}{https://github.com/acabassi/klic-pancancer-analysis}}. 

\subsection{Transcriptional module discovery}

\emph{Transcriptional modules} are groups (i.e. clusters) of genes that share a common biological function and are co-regulated by a common set of transcription factors.
It has been recognised that integrative clustering methods can be useful for discovering transcriptional modules, by combining gene expression datasets with datasets that provide information about transcription factor binding \citep{Ihmels2002, Savage2010}. 

Here we consider transcriptional module discovery for yeast (\emph{Saccharomyces cerevisiae}).
We integrate the expression dataset of \cite{Granovskaia2010high} that contains measurements related to 551 genes whose expression profiles have been measured at 41 different time points of the cell cycle with
the ChIP-chip dataset of \citet{Harbison2004} which provides binding information for 117 transcriptional regulators for the same genes. The latter was discretised as in \citet{Savage2010} and \citet{Kirk2012bayesian}. 

\section{Results}

\subsection{Simulated data}
\label{sec:simulations}

In Section \ref{sec:simulations} we apply the developed methods to the synthetic datasets.
In Section \ref{sec:comparison-with-other-methods} we compare the performances of our method for integrative clustering to COCA and other integrative clustering algorithms.

\subsubsection{KLIC}

We apply KLIC to the synthetic datasets presented in Section~\ref{sec:synthetic-datasets}. 

\begin{figure}
	\centering
	\includegraphics[width=.35\textwidth]{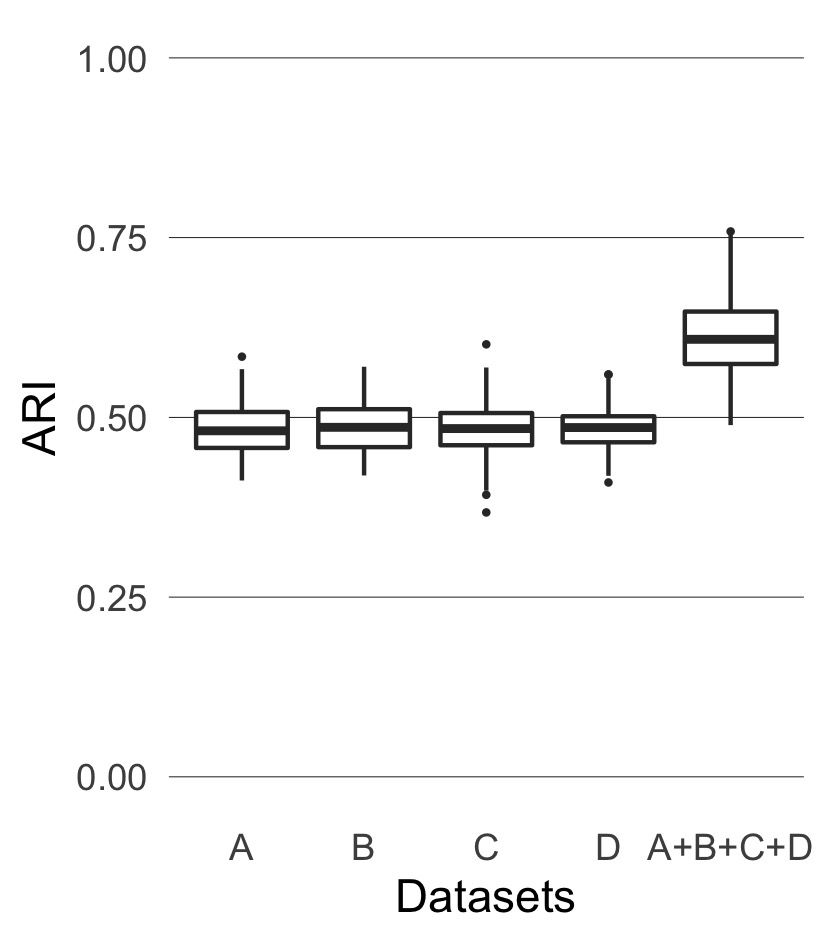}\quad 
	\includegraphics[width=.35\textwidth]{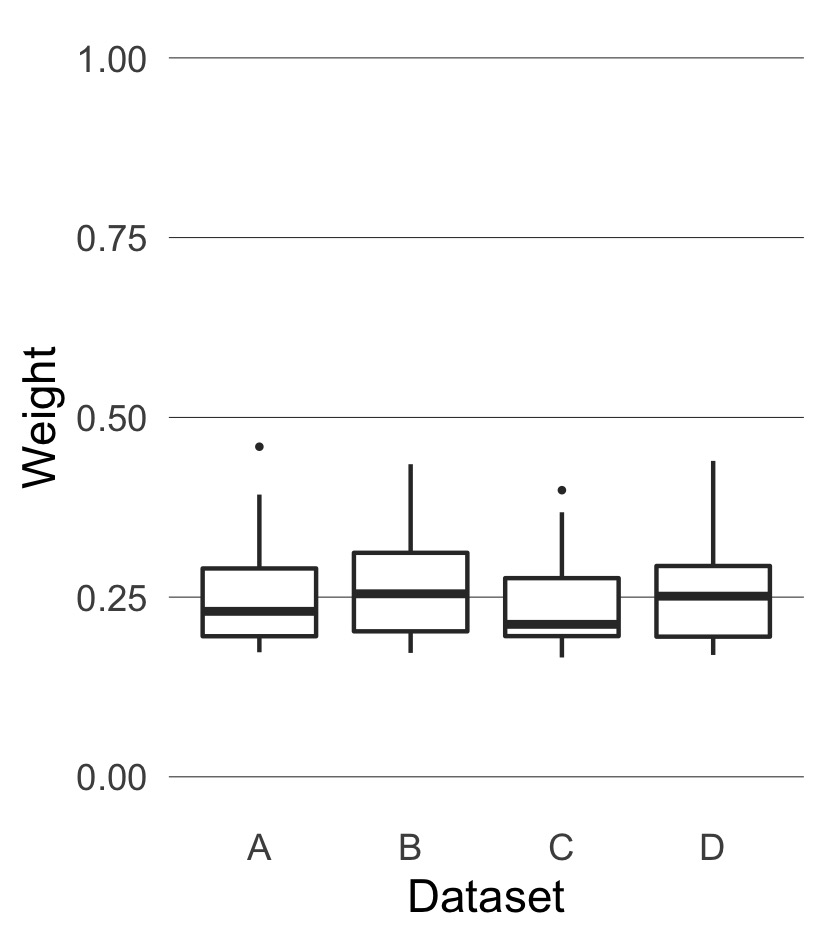}
	\caption{Results of applying KLIC to four similar datasets. Left: ARI of KLIC applied to each dataset separately (columns ``A'', ``B'', ``C'', and ``D'') and to all four datasets together (column ``A+B+C+D''). The ARI is higher in the last column because KLIC can combine information from all the datasets to find a global clustering. Right: kernel weights associated to each dataset, when applying KLIC to all four datasets together. The algorithm is able to recognise that each dataset contains the same amount of information regarding the global clustering, and therefore assigns on average the same weight to each dataset.}
	\label{fig:simulation-samerho-samecluster}
\end{figure}

\paragraph{Similar datasets.} First we run the kernel $k$-means algorithm on each of the consensus matrices that have the same clustering structure and noise level.
To assess the quality of the clustering, we compare the clustering found with the true one using the adjusted Rand index (ARI; \citealp{Rand1971}), which is equal to one if they are equal and is equal to zero if we observe as many similarities between the two partitions of the data as it is expected by chance.
Then we run KLIC on multiple datasets. In Figure \ref{fig:simulation-samerho-samecluster} are reported the box plots of the ARI obtained combining the four datasets together using KLIC (column ``A+B+C+D'') and the box plots of the average weights assigned by the KLIC algorithm to the observations in each dataset.
We observe that as expected, combining together more datasets helps recovering the clustering structure better than just taking the matrices one at a time. This is because localised kernel $k$-means allows to give different weights to each observation. Therefore, if data point $n$ is hard to classify in dataset $d_1$, but not in dataset $d_2$, we will have $\theta_{n d_1} < \theta_{n d_2}$.
However, on average the weights are divided equally between the datasets. This reflects the fact that all datasets have the same dispersion and, as a consequence, they contain on average the same amount of information about the clustering structure.

\begin{figure}
	\centering
	\includegraphics[width=.525\textwidth]{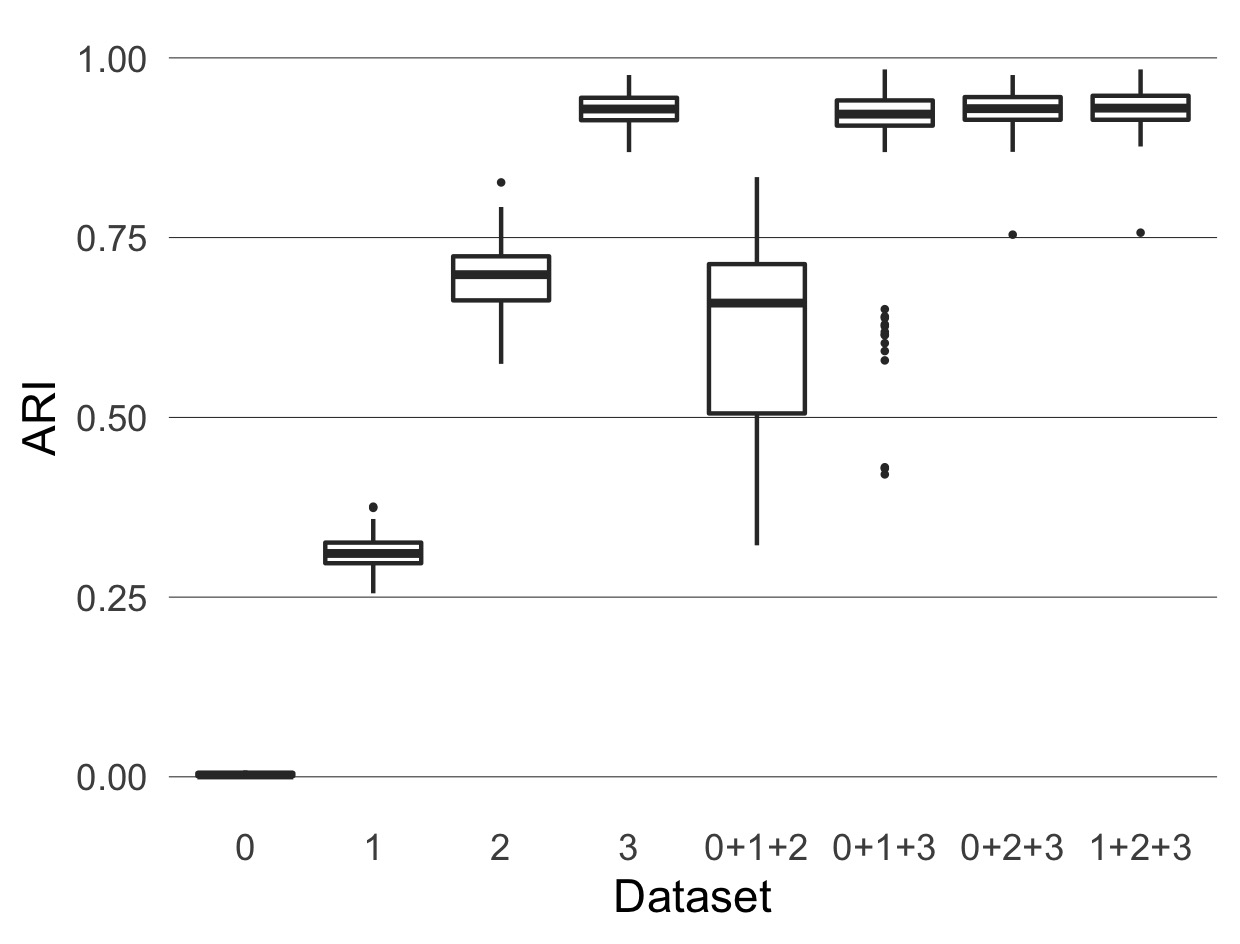}
	\includegraphics[width=.525\textwidth]{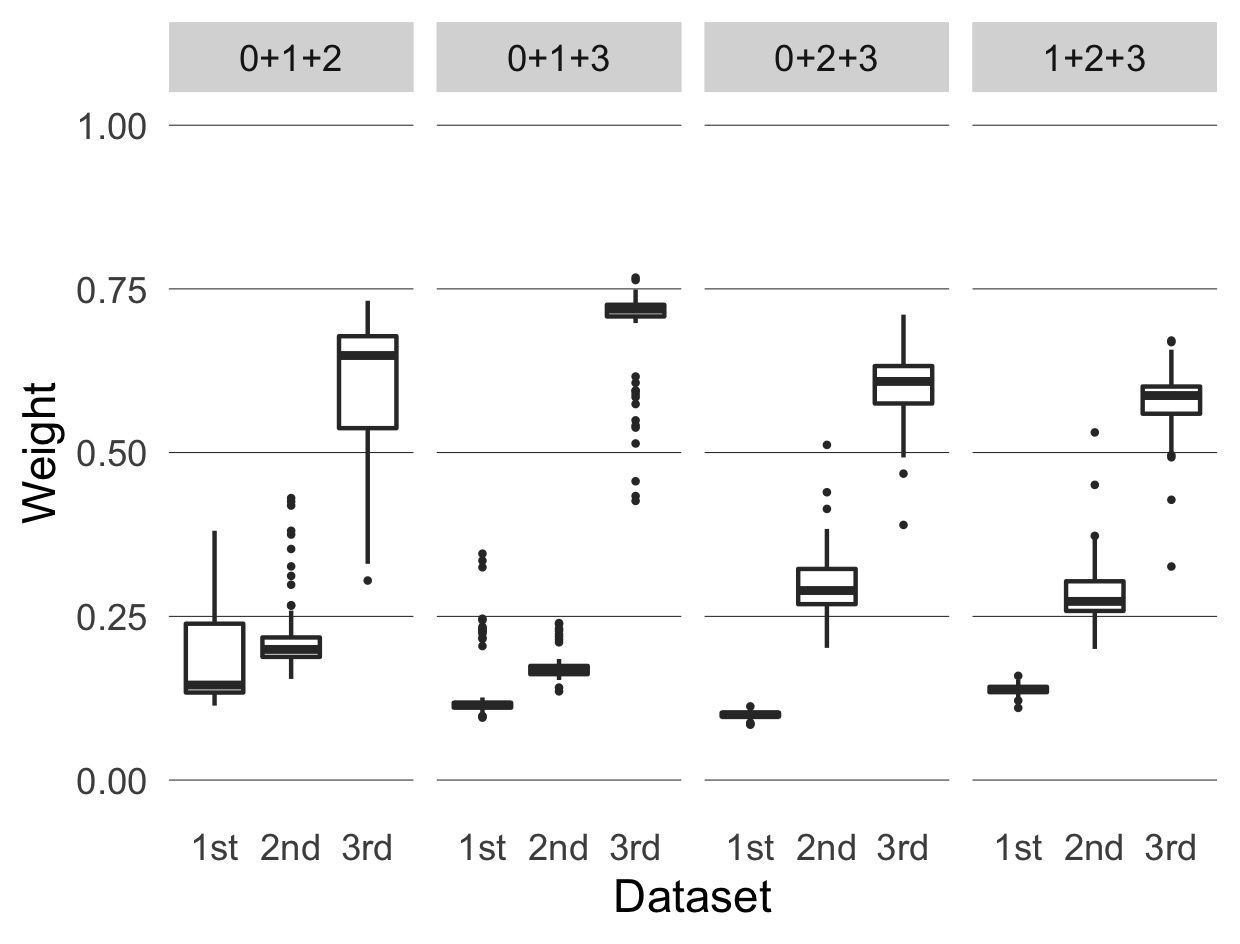}
	\caption{Results of applying KLIC to datasets with different levels of noise (``0'' indicates the dataset that has no cluster separability, ``1'' the dataset with low cluster separability, and so on). Top: ARI of KLIC applied to each dataset separately (columns ``0'', ``1'', ``2'', and ``3'') and to subsets of three of those datasets (columns ``0+1+2'', ``0+1+3'', ``0+2+3'', and ``1+2+3''). Bottom: kernel weights associated to each dataset in each of the experiments with multiple datasets, ordered by cluster separability. For example, the first subset is ``0+1+2'' so the weights marked as ``1st'' are those assigned to dataset ``0'', ``2nd'' are those assigned to ``1'' and so on. For each subset of datasets the weights of the noisier datasets (``1st'' and ``2nd'') are lower than those of the ``best'' dataset in the subset (``3rd''). This is reflected in an increased ARI in each subset, compared to applying KLIC to those datasets separately.}
	\label{fig:simulation-differentrho-samecluster}
\end{figure}

\paragraph{Datasets with different levels of noise.} Here we use the datasets shown in Figure \ref{fig:consensus-matrices-simulations}, that have the same clustering structure (six clusters of the same size each) but different levels of cluster separability.
We consider four different settings, each time combining three out of the four synthetic datasets.
Figure \ref{fig:simulation-differentrho-samecluster} shows the box plots of the 
ARI obtained using kernel $k$-means on the datasets taken one at a time (columns ``0'', ``1'', ``2'', ``3'') and the ARI obtained using KLIC on each subset of datasets (columns ``0+1+2'', ``0+1+3'', ``0+2+3'', ``1+2+3'').
As expected, the consensus matrices with clearer clustering structure give higher values of the ARI on average. Moreover, the ARI obtained combining three matrices with different levels of cluster separability is on average the same or higher as in the case when only the ``best'' matrix is considered.
This is because larger weights are assigned to the datasets that have clearer clustering structure. In the bottom part of  Figure \ref{fig:simulation-differentrho-samecluster} are reported the box plots of the average weights given by the localised multiple kernel $k$-means to the observations in each dataset. It is easy to see that each time the matrix with best cluster separability has higher weights than the other two.

\subsubsection{Comparison between KLIC, COCA and other methods}
\label{sec:comparison-with-other-methods}

We compare the performance of KLIC to the one obtained using COCA, as well as to two other comparable integrative clustering algorithms for which implementations are readily available; namely, iCluster and Clusternomics.
Additionally, we compare to localised multiple kernel $k$-means using standard radial basis function (RBF) kernels.
We use the same synthetic datasets as in the previous section.

For COCA, we use the $k$-means algorithm with Euclidean distance, fixing the number of clusters to be equal to the true one, to find the clustering labels of each dataset. Many other clustering algorithms can be used, but this is the one that gives the best results among the most common ones. To find the global clustering, we build the consensus matrices using 1000 resamplings of the data, each time with $80\%$ of the observations and all the features. The final clustering is done using hierarchical clustering with average linkage on the consensus matrix.
The iCluster model is fitted using the \verb|tune.iCluster2| function of the R package \textit{iCluster} \citep{shen2012icluster}, with number of clusters set to six.
For Clusternomics we use the \verb|contextCluster| function of the R package \textit{clusternomics} \citep{gabasova2017clusternomics}, providing the true number of clusters both for the partial and global clusterings. To assess the impact of RBF kernel parameter choice, we consider two ways to set the free parameter, $\sigma$, of the kernel. In one setting we fix $\sigma =1$, a common default value. In the second setting, $\sigma$ is tuned for each dataset to maximise the average ARI between the clustering obtained with kernel $k$-means using the RBF kernel and the true clusters (more information about this procedure can be found in the Supplementary Material). Although this procedure clearly could not be applied in practice (where the true clustering is unknown), it is used here to determine a putative upper bound on the performances of MKL with this kernel.

\begin{figure}
	\centering
	\includegraphics[width=0.4\textwidth]{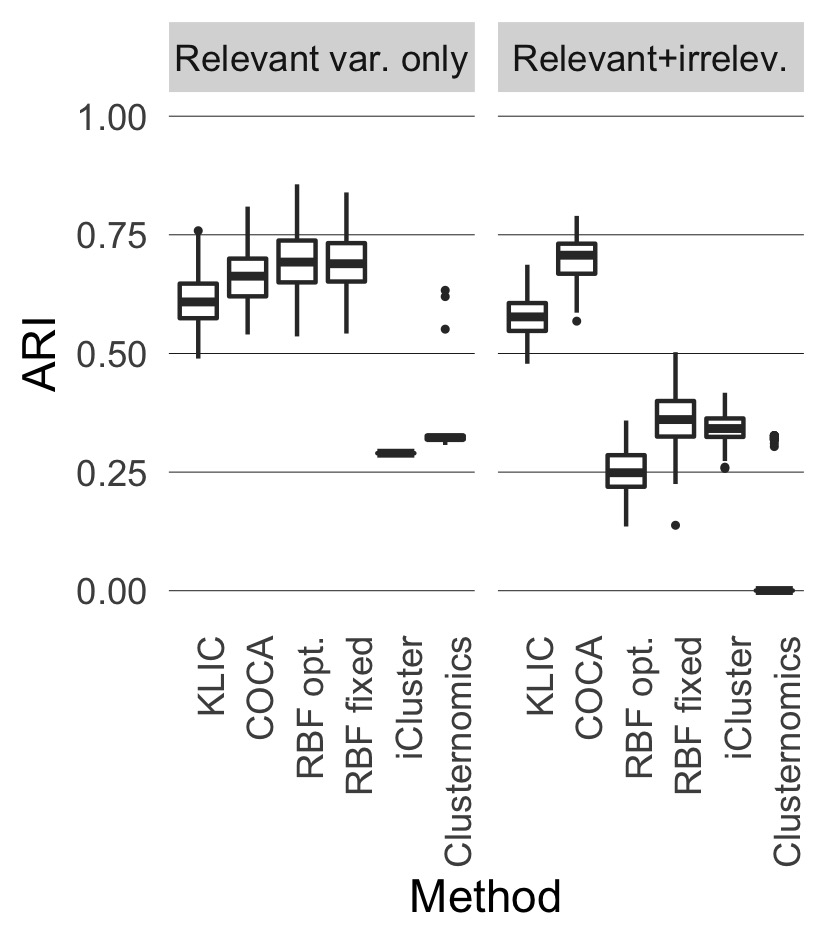}
	\includegraphics[width=0.8\textwidth]{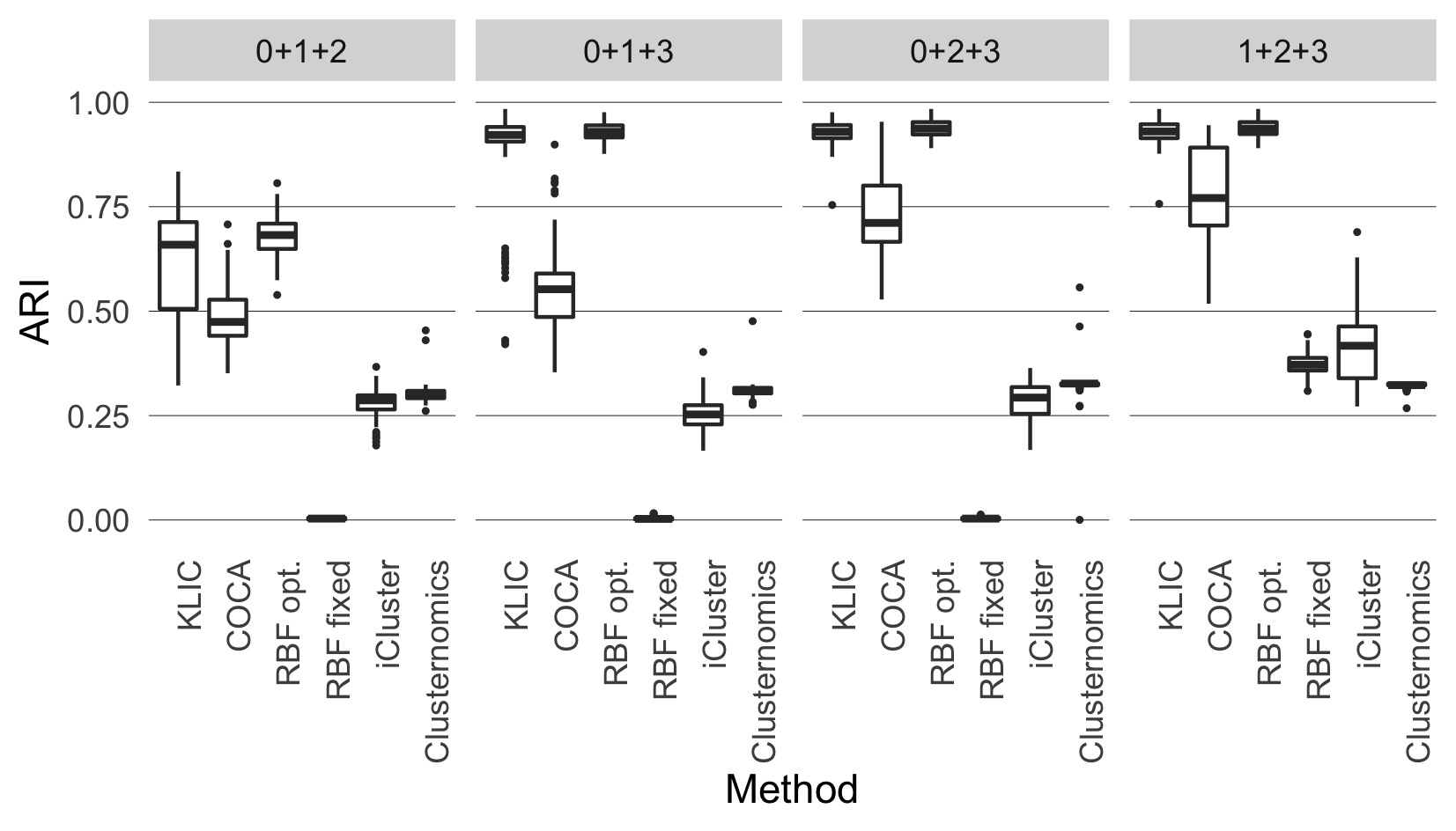}
	\caption{Comparison between KLIC, COCA, and other clustering algorithms.  The labels ``RBF opt.'' and ``RBF fixed'' refer to the MKL method using an RBF kernel with either $\sigma$ optimised or fixed at 1 (see text). Top: ARI obtained with each clustering algorithm using four datasets having the same clustering structure and cluster separability (as in Figure \ref{fig:simulation-samerho-samecluster}). Bottom: ARI obtained with COCA and KLIC for each of the subsets of heterogeneous datasets considered in Figure \ref{fig:simulation-differentrho-samecluster}. The high ARI obtained with KLIC in all settings shows the advantage of using this method, especially when some of the datasets are noisy.}
	\label{fig:comparisons}
\end{figure}

\paragraph{Similar datasets.} We combine four datasets that have the same clustering structure and cluster separability.
In Figure \ref{fig:comparisons} is shown the ARI of all considered methods applied to 100 sets of data of this type. 
In the first setting, where only variables relevant for the clustering are present, the localised multiple kernel $k$-means with RBF kernel has the highest median ARI, followed by COCA and KLIC.
To cluster the data that include noisy variables, we replace the $k$-means algorithm by the sparse $k$-means feature selection framework of \citet{witten2010framework} in COCA and KLIC, using the R package \textit{sparcl} \citep{witten2018sparcl}. Thanks to this, the performances of these two methods are not affected by the presence of irrelevant variables. COCA, in particular, has the highest median ARI, followed by KLIC. This shows that both methods work well in the case of multiple datasets that have the same clustering structure and level of noise and, in contrast to the four other methods considered here, can be straightforwardly modified to deal with the presence of irrelevant features.

\paragraph{Datasets with different levels of noise.} We also compare the behaviour of all methods in the presence of multiple datasets with the same clustering structure, but different levels of cluster separability. The ARI is shown in Figure \ref{fig:comparisons}.
We observe that, in each of the four simulation settings, KLIC and the optimised version of localised multiple kernel $k$-means with RBF kernel have the highest ARI scores. The reason for this is that COCA, iCluster, and Clusternomics are not weighted methods, so their ability to recover the true clustering structure is decreased by adding noisy datasets. Instead, we have shown in the previous section that KLIC allows to give lower weights to the noisiest datasets, achieving better performances. We emphasise that the optimal values of the RBF parameters have been determined making use of the true cluster labels, which will not be possible in real applications. The performance achieved when the RBF kernel parameter, $\sigma$, is fixed to 1 may therefore be more representative of what can be achieved in practice.

\quad 

Overall, these comparisons suggest that KLIC may be a good default choice, since it can be run in such a way that it is robust to both the inclusion of noisy variables (via the choice of an appropriate clustering algorithm) and of noisy datasets.

\subsection{Multiplatform analysis of 12 cancer types}
\begin{figure}[h!]
	\centering
	\begin{subfigure}[b]{.85\textwidth}
		\includegraphics[width=\textwidth]{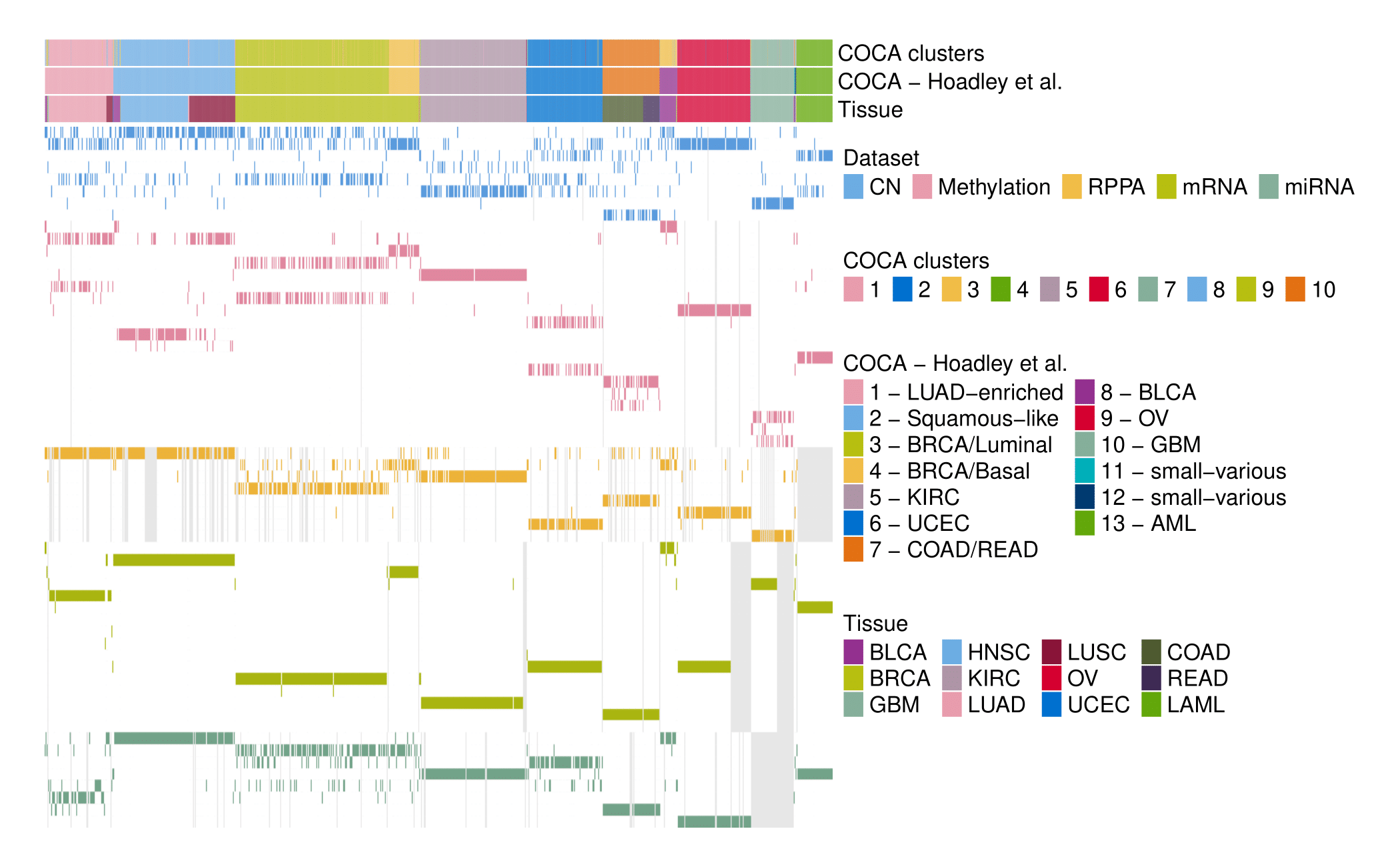}
		\caption{Matrix-of-clusters.}
		\label{fig:pancan12-moc}
	\end{subfigure}
	\begin{subfigure}[b]{.4375\textwidth}
		\includegraphics[width=\textwidth]{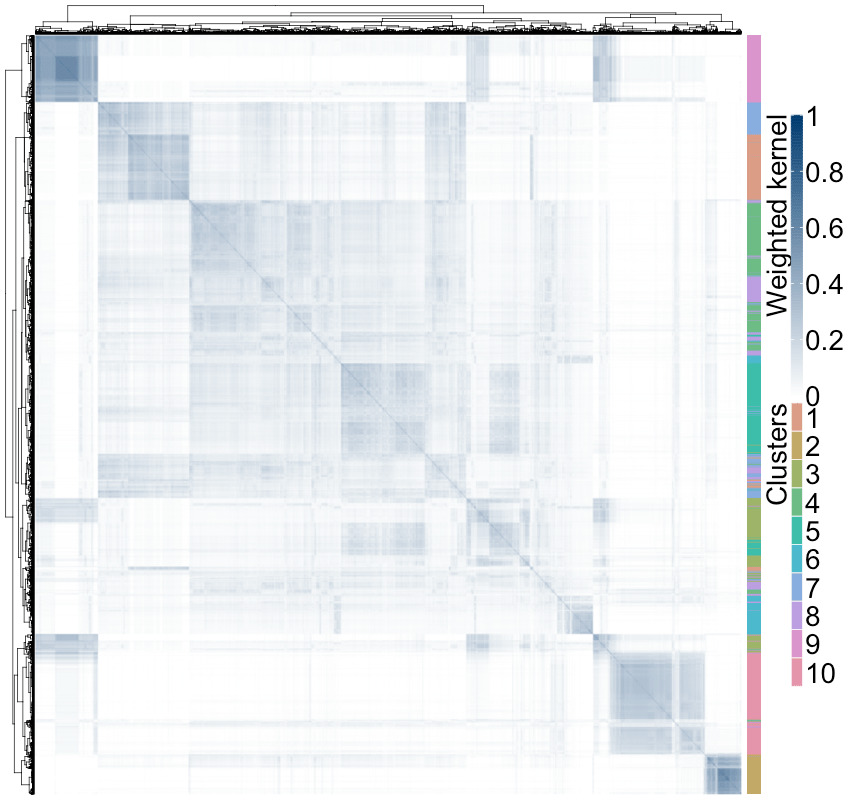}
		\caption{Weighted similarity matrix.}
		\label{fig:pancan12-weighted-similarity-matrix}
	\end{subfigure}
	\begin{subfigure}[b]{.455\textwidth}
		\includegraphics[width=\textwidth]{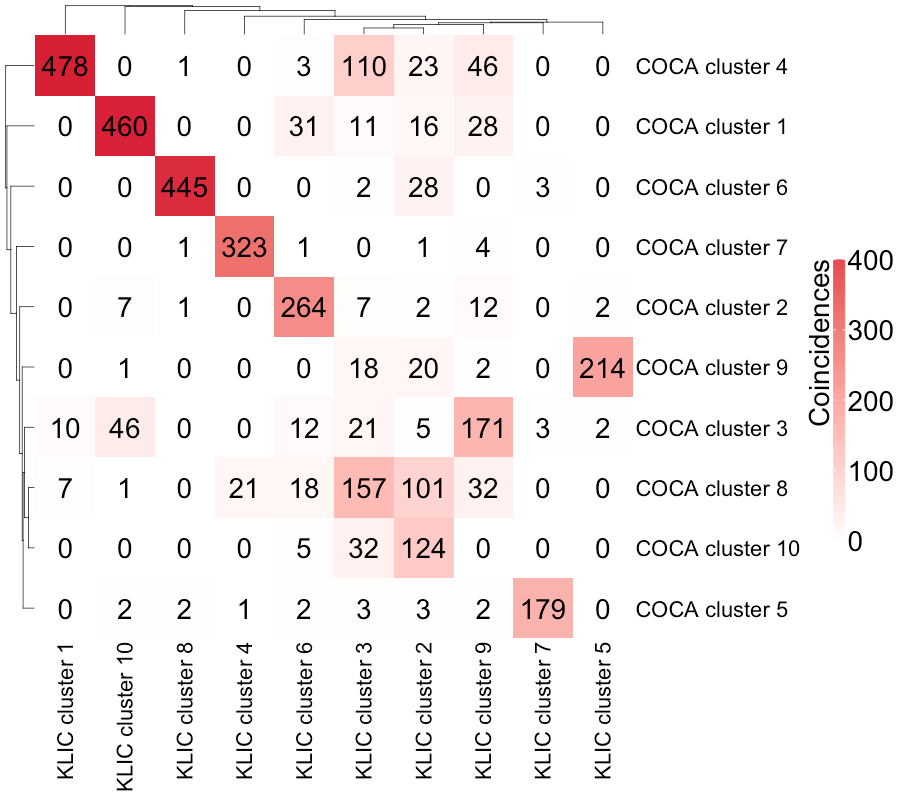}
		\caption{Coincidence matrix.}
		\label{fig:pancan12-coincidence-matrix}
	\end{subfigure}
	\caption{Multiplatform analysis of 12 cancer types. \textbf{(a)} Matrix-of-clusters of the pan-cancer data, each row corresponds to a cluster in one of the dataset, and each column corresponds to a tumour sample. Coloured cells show which tumours belong to each cluster. Gray cells indicate missing observations. \textbf{(b)} Weighted similarity matrix. \textbf{(c)} Coincidence matrix comparing the clusters given by COCA and KLIC.}
	\label{fig:pancan12}
\end{figure}

The first step of the data analysis is dedicated to replicating the analysis performed by \cite{Hoadley2014}. 
The DNA copy number, DNA methylation, mRNA expression, microRNA expression, and  protein expression data were preprocessed in the same way as \cite{Hoadley2014} did. 
We then clustered the tumour samples independently for each dataset, using the same clustering algorithm as in the original paper.  We compared the clusters we obtained to those reported by \cite{Hoadley2014} for different number of clusters, and we found that the best correspondence was given by choosing the same number of clusters as in the original paper, except for the microRNA expression data, for which we found the best number of clusters to be seven (instead of 15). Figure \ref{fig:pancan12-moc} shows the MOC matrix formed by these clusters and the resulting COCA clusters. As can be seen from the Figure, each dataset has some missing observations. The corresponding entries in the MOC matrix were set to zero. We chose the number of clusters that maximises the silhouette, as suggested by \cite{Aure2017}, which is ten.

We then applied KLIC to the preprocessed data, building one consensus matrix for each dataset, using the same clustering algorithm and number of clusters as for COCA. We assigned weight zero to every missing observation (more details on how to use KLIC with incomplete data can be found in the Supplementary Material). The weighted consensus matrix is shown in Figure \ref{fig:pancan12-weighted-similarity-matrix}. The weights assigned on average to the observations in each dataset are as follows: copy number 31.4\%, methylation 19.2\%, miRNA 17.8\%, mRNA 16.4\%, protein 15.2\%.

Similarly to what was observed by \cite{Hoadley2014}, both the clusters obtained using COCA and KLIC correspond well with the tissue-of-origin classification of the tumours. However, there are a few differences between the two: the coincidence matrix is shown in Figure \ref{fig:pancan12-coincidence-matrix}. Further details on how we tried to replicate the data analysis of  \cite{Hoadley2014} and how we applied KLIC to these data can be found in the Supplementary Material.

\subsection{Transcriptional module discovery}

\begin{figure}[h!]
	\centering	
	\includegraphics[width=0.32\textwidth]{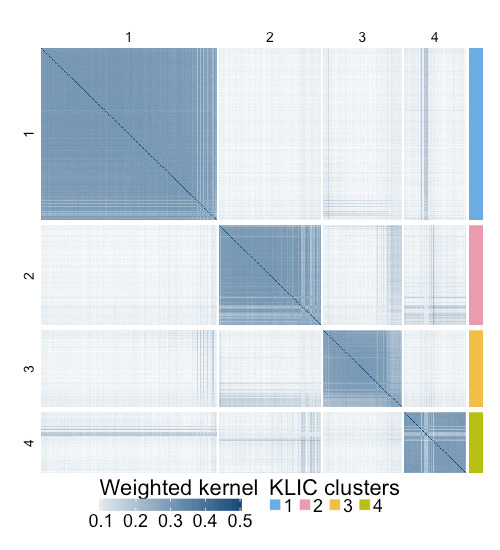}
	\includegraphics[width=0.32\textwidth]{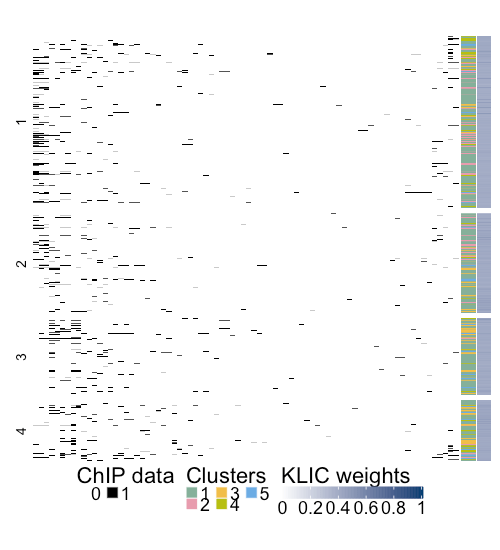}
	\includegraphics[width=0.32\textwidth]{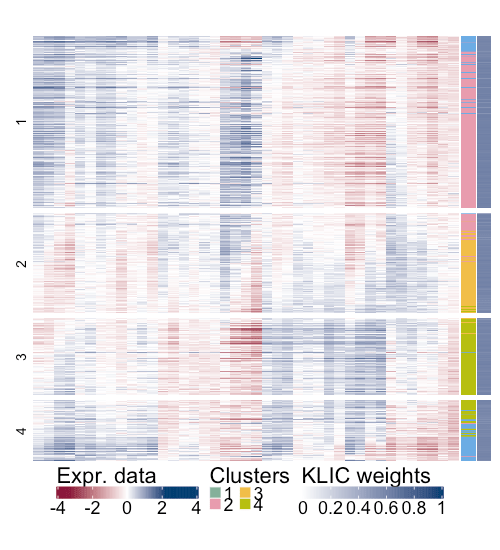}
	\caption{Transcriptional module discovery, KLIC output.
	Left: weighted kernel matrix obtained with KLIC, where each row and column corresponds to a gene, and final clusters.
	Centre: transcription factor data, where each row represents a gene and each column a transcription factor, black dots correspond to transcription factors that are believed to be able to bind to the promoter region of the corresponding gene with high confidence; clusters obtained using BHC on the transcription factor data and weight assigned by KLIC to each data point.
	Right: gene expression data, where each row is a gene and each column a time point, clusters obtained using PAM on the gene expression data, and weights assigned by KLIC to each data point.}
	\label{fig:transcriptional-module-discovery}
\end{figure}

We clustered the 551 genes based on the gene expression and transcription factor data using KLIC. For each dataset, the consensus matrices were obtained as explained in Section \ref{sec:coca}. The clustering algorithms used in this step were partitioning around medoids (PAM; \citealp{Kaufman2009finding}) with the correlations between data points as distances for the gene expression data and Bayesian hierarchical clustering (BHC) for the transcription factor data \citep{Heller2005bayesian, Cooke2011bayesian}. The consensus matrices obtained in this way were then used as input to KLIC. The algorithm was run with number of clusters ranging from two to 20. We found that the silhouette is maximised by setting the number of clusters to four.  Figure \ref{fig:transcriptional-module-discovery} shows the weighted kernel matrix given by KLIC where the rows and columns are sorted by final cluster. Next to it are reported the data, where the observations are in the same order as in the kernel matrix. The clusters obtained independently on each dataset are also shown on the right of each plot. The kernel matrices of each dataset can be found in the Supplementary Material.

We also applied COCA to this dataset, with the initial clusters for each dataset obtained with the same clustering algorithms as those used for the consensus matrices. The metrics used to choose the number of clusters for the initial clustering of the expression data are reported in the Supplementary  Material. BHC does not require the number of clusters to be set by the user. \textcolor{black}{For the final clustering the number of clusters was chosen in order to maximise the silhouette, considering all values between two and ten. This resulted in choosing the 10-cluster solution.}

In order to assess the quality of the clusters, we make use of the Gene Ontology Term Overlap (GOTO) scores of \cite{Mistry2008}. Each score is an indication of the number of annotations that, on average, are shared by genes belonging to the same clusters. These are available for three different ontologies: biological process, molecular function and cellular component. More details on these scores and how they are calculated can be found in the Supplementary Material of \cite{Kirk2012bayesian}. We report in Table \ref{table:goto-scores} the GOTO scores of both KLIC and COCA clusters, for both number of clusters selected by KLIC (four) and COCA (ten). We also show the scores obtained clustering each dataset separately. We observe that, while in the case of four clusters no information is lost by combining the datasets, by dividing data into ten clusters one obtains more biologically meaningful clusters. Moreover, KLIC does a better job at combining the datasets, by better exploiting the information contained in the data and down-weighting the kernel of the ChIP dataset, which contains less information. More details about the kernel matrices and weights can be found in the Supplementary Material.

\begin{table}[h]
\centering
\begin{tabular}{ l l l c c c }
Clusters & Dataset(s) & Algorithm & GOTO BP & GOTO MF & GOTO CC \\
\hline
8 & ChIP & \textcolor{black}{BHC} & 6.09 & 0.90 & 8.33 \\
4 & Expression & \textcolor{black}{PAM} & 6.12 & 0.91 & 8.41 \\
4 & ChIP+Expr. & COCA & 6.12 & 0.91 & 8.41 \\
4 & ChIP+Expr. & KLIC  & 6.12 & 0.91 & 8.41 \\
10 & ChIP+Expr. & COCA & 6.28 & 0.93 & 8.51 \\
10 & ChIP+Expr. & KLIC & 6.32 & 0.95 & 8.53 \\
\hline
\end{tabular}
\caption{Gene Ontology Term Overlap scores for different sets of data, clustering algorithms and numbers of clusters. ``BP'' stands for Biological Process ontology, ``MF'' for Molecular Function, and ``CC'' for Cellular Component.}
\label{table:goto-scores}
\end{table} 

\section{Discussion}
In the first part of this work we have given the algorithm for COCA, a widely used method in integrative clustering of genomic data, highlighting the main issues of using this method.
We have also presented KLIC, a novel approach to integrative clustering, that allows multiple datasets to be combined to find a global clustering of the data and is well-suited for the analysis of large datasets, such as those often encountered in genomics applications. A defining difference between KLIC and COCA is that, while COCA performs a combination of the clusters found in each dataset, KLIC uses the similarities between data points observed in each dataset to perform the integrative step. 
Moreover, KLIC weights each dataset individually, which allows more informative datasets to be upweighted relative to less informative ones, as demonstrated in our simulation study.
Finally, we have used KLIC to integrate multiple 'omic datasets, in two different real world applications, finding biologically meaningful clusters. The results compare favourably to those obtained with COCA.


\section*{Funding}
This work was supported by the MRC [MC\_UU\_00002/10 and MC\_UU\_00002/13], and the National Institute for Health Research [Cambridge Biomedical Research Centre at the Cambridge University Hospitals NHS Foundation Trust]. The views expressed are those of the authors and not necessarily those of the NHS, the NIHR or the Department of Health and Social Care.

\bibliographystyle{apalike}
\addcontentsline{toc}{section}{Bibliography}
\bibliography{main}

\end{document}


\begin{center}
{\LARGE\bf Multiple kernel learning for integrative consensus clustering of 'omic datasets}
\end{center}
\medskip
\begin{center}
{\large Alessandra Cabassi$^{1}$ and Paul D. W. Kirk$^{1,2}$ \\[15pt]
\emph{$^{1}$MRC Biostatistics Unit}\\
\emph{$^{2}$Cambridge Institute of Therapeutic Immunology \& Infectious Disease (CITIID)\\
University of Cambridge, U.K.}\\
}

\end{center}

\bigskip

\bigskip\bigskip

\tableofcontents
\clearpage


\section{Methods}
\subsection{Kernel $k$-means clustering}
\label{sec:kernel-kmeans}
Before moving on to the kernel $k$-means, we first describe the original $k$-means clustering algorithm \citep{Steinhaus1956}. Let $\boldsymbol{x}_1, \dots, \boldsymbol{x}_N$ indicate the observed dataset, with $\boldsymbol{x}_n \in \mathbb{R}^P$ and $z_{nk}$ be the corresponding cluster labels, where $\sum_k z_{nk} = 1$ and $z_{nk} = 1$ if $x_n$ belongs to cluster $k$, zero otherwise.
We denote by $Z$ the $N \times K$ matrix with $ij$-th element equal to $z_{ij}$.
The goal of the $k$-means algorithm is to minimise the sum of all squared distances between the data points $\boldsymbol{x}_n$ and the corresponding cluster centroid $\boldsymbol{m}_k$. The optimisation problem is 
\begin{mini!}[2]
	{Z}{\sum_{n} \sum_{k} z_{nk} \lVert \boldsymbol{x}_n - \boldsymbol{m}_k \rVert^2_2 \label{eq:of_kmeans}}
	{\label{opt:kmeans}}
	{}
	\addConstraint{\sum_k z_{nk}}{=1, \; \forall n}
	\addConstraint{ N_k = }{\sum_n z_{nk}, \; \forall k}
	\addConstraint{\boldsymbol{m}_k }{= \frac{1}{N_k} \sum_n z_{nk} \boldsymbol{x}_n, \; \forall k.}
\end{mini!}

Now we can show how the kernel trick works in the case of the $k$-means clustering algorithm \citep{Girolami2002}.
Redefining the objective function of Equation \eqref{eq:of_kmeans} based on the distances between observations and cluster centres in the feature space $\mathcal{H}$, the optimisation problem becomes:
\begin{mini!}[2]
	{Z}{\sum_n \sum_k z_{nk} \lVert \phi(\boldsymbol{x}_n) - \tilde{\boldsymbol{m}}_k \rVert^2_{\mathcal{H}} \label{eq:of_kernelkmeans}}
	{\label{opt:kernelkmeans}}
	{}
	\addConstraint{\sum_k z_{nk}}{=1, \; \forall n}
	\addConstraint{ N_k = }{\sum_n z_{nk}, \; \forall k}
	\addConstraint{\tilde{\boldsymbol{m}}_k }{= \frac{1}{N_k} \sum_n z_{nk} \phi(\boldsymbol{x}_n), \; \forall k.}
\end{mini!}
where we indicated by $\tilde{\boldsymbol{m}}_k$ the cluster centroids in the feature space $\mathcal{H}$.  
Using this kernel, each term of the sum in Equation \eqref{eq:of_kernelkmeans} can be written as a function of $ \delta(\boldsymbol{x}_i, \boldsymbol{x}_j).$
Therefore, there is no need to evaluate the map $\phi$ at every point $\boldsymbol{x}_i$ to compute the objective function of Equation \eqref{eq:of_kernelkmeans}. Instead, one just needs to know the values of the kernel evaluated at each pair of data points $\delta(\boldsymbol{x}_i,\boldsymbol{x}_j)$, $i,j = 1, \dots, N$. This is what is commonly referred to as the kernel trick.

Defining $L$ as the $K \times K$ diagonal matrix with $k$th diagonal element equal to $N_k^{-1}$ and $\Delta$ the $N \times N$ matrix with $ij$th entry equal to $\delta(\boldsymbol{x}_i, \boldsymbol{x}_j)$, the optimisation problem \eqref{opt:kernelkmeans} can be rewritten as a trace maximisation problem \citep{Gonen2014}:
\begin{maxi!}[2]
	{Z}{\text{tr}(L^{\frac{1}{2}}Z'\Delta ZL^{\frac{1}{2}})}
	{\label{opt:kernelkmeans_matrix}}
	{}
	\addConstraint{Z 1_k}{=1_n}
	\addConstraint{z_{nk}}{\in \{0,1\}, \quad \forall n,k. \label{eq:integrality}}
\end{maxi!}
The integrality constraints make this problem difficult to solve.
However, the corresponding linear problem obtained by relaxing the integer constraints of Equation~\eqref{eq:integrality} to  $0 \leq z_{nk} \leq 1$ for all $n,k$ can be solved by performing kernel PCA on the kernel matrix $\Delta$ and setting the matrix $H = ZL^{\frac{1}{2}}$ to the $K$ eigenvectors that correspond to $K$ largest eigenvalues \citep{Scholkopf1998}. The clustering solution can be found by first normalising all rows of $H$ to be on the unit sphere and then performing $k$-means clustering on the normalised matrix. Other possible approaches to derive a final clustering from $H$ are listed in \citet{Shawe2004}. 

\subsection{How to use KLIC with incomplete data}
\label{sec:incomplete-data}

This section is dedicated to giving further details about how missing data can be handled by using KLIC. The strategy explained in this section was used in the application of KLIC to the multiplatform analysis of 12 cancer types in Section 4.2 of the main paper.

The optimisation problem that is solved to find the optimal clustering and weights in localised multiple kernel $k$-means is:
\begin{maxi!}[2]
	{H, \Theta}{\text{tr}(H'\Delta_{\Theta} H) - \text{tr}(\Delta_{\Theta})\label{eq:of_localisedmultiplekernelkmeans}}
	{\label{opt:localisedmultiplekernelkmeans}}
	{}
	\addConstraint{H'H}{=1_k}
	\addConstraint{\Theta'1_M}{ = 1}
	\addConstraint{\Delta_{\Theta}}{= \sum_m (\boldsymbol{\theta}_m \boldsymbol{\theta}_m') \circ \Delta_m,} 
\end{maxi!}
where $\circ$ is the Hadamard product.
As stated in Section 2.2.2 of the main paper, one can optimise the objective function of Equation \eqref{eq:of_localisedmultiplekernelkmeans} with a two-step procedure, that iteratively (1) solves a standard kernel $k$-means problem with kernel $\delta_{\Theta}$, keeping the weight matrix $\Theta$ fixed and then (2) optimises the objective function with respect to $\Theta$. Again, the first step reduces to solving one optimisation problem with a single kernel (Equations \ref{opt:kernelkmeans_matrix}) and in the second step one just needs to solve a quadratic programming (QP) problem. In particular, the QP problem in step (2) is:
\begin{mini!}[2]
	{\Theta}{\sum_{m=1}^{M} \boldsymbol{\theta}_m^T \big((I_n-HH^T) \circ \Delta_m  \big) \boldsymbol{\theta}_m  \label{eq:of_localisedmultiplekernelkmeans-qp}}
	{\label{opt:localisedmultiplekernelkmeans-qp}}
	{}
	\addConstraint{\Theta}{\in \mathbb{R}_{+}^{N \times M}}
	\addConstraint{\Theta'1_M}{ = 1_N.}
\end{mini!}

Now, if some of the observations are missing in some of the datasets, we can define by $I_m \subset \{1, \dots, N\}$ the
set of the missing values in each dataset $m = 1, \dots, M$ and make sure that the corresponding kernel $\Delta_m$ is such that 
\begin{align*}
\Delta^m_{ij} = 0 & \quad \forall i \in I_m, j \not = i,\\
\Delta^m_{ii} = 1 & \quad \forall i \in I_m.
\end{align*}
The resulting matrix $\Delta_m$ is a weighted sum of co-clustering matrices with structure
\begin{equation*}
\Delta_m = 
\begin{bmatrix*}[c]
\Delta'_m & 0 & 0 & 0 \\
0 & 1 & 0 & 0\\
0 & 0  & \ddots & 0\\
0 & 0 & 0  & 1 \\
\end{bmatrix*},
\end{equation*}
where $\Delta'_m$ is the $m$-th kernel matrix for the available data and the observations are ordered such that the missing ones are at the bottom of the matrix for presentational purposes. Therefore, it is a valid kernel matrix. 

Moreover, it is possible to cancel the influence the missing observations on the final solutions by setting their weight to zero in optimisation problem \eqref{opt:localisedmultiplekernelkmeans-qp}:
\begin{mini!}[2]
	{\Theta}{\sum_{m=1}^{M} \boldsymbol{\theta}_m^T \big((I_n-HH^T) \circ \Delta_m  \big) \boldsymbol{\theta}_m  \label{eq:of_localisedmultiplekernelkmeans-qp-missingdata}}
	{\label{opt:localisedmultiplekernelkmeans-qp-missingdata}}
	{}
	\addConstraint{\Theta}{\in \mathbb{R}_{+}^{N \times M}}
	\addConstraint{\Theta'1_M}{ = 1_N}
	\addConstraint{\theta_{mi}}{ = 0}{\forall i \in I_m,\ m = 1, \dots, M. \label{eq:additional-constraints}}
\end{mini!}
This corresponds to adding $|I_1|  + \dots + |I_M|$ equality constraints, each one on a different variable, or, equivalently, to removing a number $|I_1|  + \dots + |I_M|$ of variables from the optimisation problem. Therefore, \eqref{opt:localisedmultiplekernelkmeans-qp-missingdata} is a QP problem.
The objective function \eqref{opt:localisedmultiplekernelkmeans} can then be minimised by iterating between steps (1) and (2) as in the previous case, with the additional constraints \eqref{eq:additional-constraints} in step (2).

\subsection{Algorithms}

\begin{algorithm}
    	
 \SetKwInOut{Input}{Input}
 \SetKwInOut{Output}{Output}
 \SetKwInOut{Initialise}{Initialise}
 
 \Input{Dataset $X$, number of clusters $K$.}
 \Initialise{Consensus matrix $\Delta^K = 0_{N \times N}$.\\
 Matrix of resampling counts $D_{ij} = 0_{N \times N}$.}
 \For{$h \in \{1, \dots, H\}$}{
    	$X^{(h)} = $ resample from the rows and/or columns of $X$\\
    	$\mathbf{c}^{(h)} =$ divide the items of $X^{(h)}$ into $K$ clusters\\
    	$C^{(h)} =$ build the co-clustering matrix corresponding to $\mathbf{c}^{(h)}$\\ 
    	\For{$i,j \in \{1,\dots, n\}$}{
    		$\Delta^K_{ij} = \Delta^K_{ij} + C^{(h)}_{ij}$\\
    		$D_{ij} = D_{ij} + \mathds{1}_{ij}^{(h)}$
    	}
 }
 \For{$i,j \in \{1,\dots, n\}$}{
 	$\Delta^K_{ij} = \Delta^K_{ij} / \min\left\lbrace D_{ij}, 1 \right\rbrace$
 }
 \Output{Consensus matrix $\Delta^K$.}

 \caption{Consensus cluster (CC).}
 \label{algo:consensus-cluster}
\end{algorithm}

\begin{algorithm}
	
	\SetKwInOut{Initialise}{Initialise}	
	\SetKwInOut{Input}{Input}
	\SetKwInOut{Output}{Output}

	\Input{$M$ datasets $X_m$\\
	Number of clusters $K_m$ in each dataset\\
	Global number of clusters $K$.}
	\Initialise{MOC = $0_{\bar{K} \times N}$.}
	\For{$m \in \{1, \dots, M\}$}{
	$\mathbf{c}^m$ = cluster the items in dataset $X_m$ into $K_m$ clusters\\
	\For{$n \in \{1, \dots, N\}$, $k \in \{1, \dots, K_m\}$}{
			Set $MOC_{n, m_k} = 1$ if $\mathbf{c}^m_i = k$ 
		}
	}
	\For{$h \in \{1, \dots, H\}$}{
		$MOC^{(h)} = $ resample from the rows and/or columns of MOC\\
		$\mathbf{c}^{(h)} =$ divide the items of $X^{(h)}$ into $K$ clusters\\
    		$C^{(h)} =$ build the co-clustering matrix corresponding to $\mathbf{c}^{(h)}$\\ 
			\For{$i,j \in \{1,\dots, n\}$}{
				$\Delta_{ij} = \Delta_{ij} + C^{(h)}_{ij}$\\
    				$D_{ij} = D_{ij} + \mathbb{I}_{ij}^{(h)}$
			}
	}
	\For{$i,j \in \{1,\dots, n\}$}{
 		$\Delta_{ij} = \Delta_{ij} / \min\left\lbrace D_{ij}, 1 \right\rbrace$
 	}
	Find final clustering $\boldsymbol{c}^{K}$ using hierarchical clustering on $\Delta^{K}$.\\
	\Output{Cluster labels $\boldsymbol{c}^{K}$.}
	\caption{Cluster of clusters analysis (COCA)}
	\label{algo:coca}
\end{algorithm} 

\begin{algorithm}
	
	\SetKwInOut{Input}{Input}
	\SetKwInOut{Output}{Output}
	\SetKwInOut{Initialise}{Initialise}
	\SetKwRepeat{Do}{do}{while}
	
	\Input{$M$ datasets $X_m$\\
	Maximum number of clusters $K$.}
	\For{$m \in \{1, \dots, M\}$}{
		$\Delta_m$ = compute kernel for $X_m$
	}
	\For{$k \in \{1, \dots, K\}$}{
		$[\boldsymbol{w}_k, \boldsymbol{c}_k]$ = apply multiple kernel k-means to $\Delta_1, \dots, \Delta_M$ \\
		$\boldsymbol{s}_k$ = calculate average silhouette of $\boldsymbol{c}_k$
	}
	Choose $k$ such that $s_k \geq s_j, \forall j \not = k$.\\
	\Return $k$, $\boldsymbol{w}_k$, $\boldsymbol{c}_k$.\\
	\Output{Best number of clusters $k$\\
	Set of kernel weights $\boldsymbol{w} = [w_1, \dots, w_M]$\\
	Cluster labels $\boldsymbol{c} = [c_1, \dots, c_N] $}
	\caption{KLIC: Kernel Learning Integrative Clustering}
	\label{algo:ukic}
\end{algorithm} 

\clearpage

\section{Simulation study}
\subsection{RBF kernel}
\label{sec:rbf-kernel}

The RBF kernel is defined as
\begin{equation}
\delta (\mathbf{x}, \mathbf{x}') = \exp \left\lbrace - \frac{\| \mathbf{x} - \mathbf{x}'\|^2}{2 \sigma^2} \right\rbrace,
\end{equation}
where $\boldsymbol{x}, \boldsymbol{x}' \in \mathbb{R}^P$, $\| \cdot \|$ is the Euclidean distance and the parameter $\sigma$ is the so-called \emph{characteristic length scale}. In order to find the best possible value of $\sigma$ for each synthetic dataset, we generate 100 dataset for each value $s$ (the parameter that indicates the separation between cluster means) considered in our simulation setting, which are as follows:
\begin{itemize}
	\item $ s = 1.5$ in setting 1 (similar datasets);
	\item $ s = 0, 1, 2, 3$ in setting 2 (datasets with different levels of noise);
	\item $ s= 0, 0.5, 1, 1.5, 2, 2.5, 3, 3.5, 4$ in the additional simulation settings presented below (Section \ref{sec:additional-simulation-settings}).
\end{itemize}
For each dataset, we build one kernel for each of the following values of $\sigma$: 0.001, 0.005, 0.01, 0.05, 0.1, 0.5, 1, 5, 10, 50. We then use kernel $k$-means to cluster the data and compute the ARI between the clustering obtained in this way and the true cluster labels (Figure \ref{fig:rbf-ari}). We then choose the value of $\sigma$ maximising the average ARI for each value of $s$.

\begin{figure}[H]
	\centering
	\includegraphics[width=.8\textwidth]{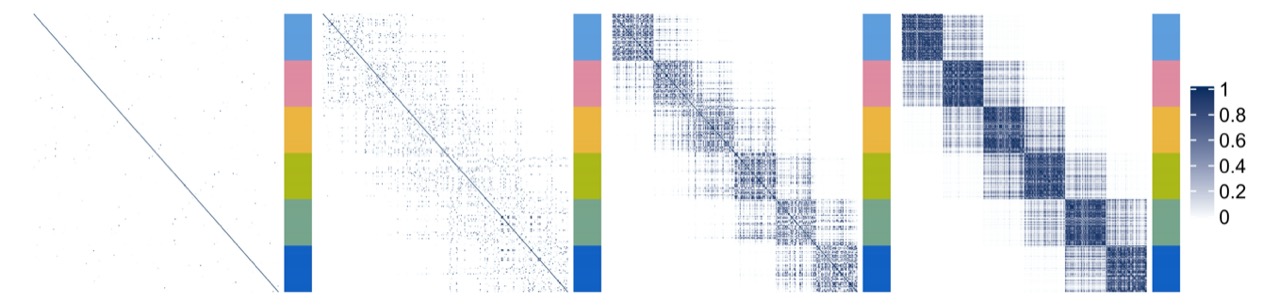}
	\caption{Kernels obtained for the same datasets as those used for Figure 1 (first row) in the main paper, using RBF kernels.}
	\label{fig:rbf-gram-matrices}
\end{figure}

\begin{figure}
	\centering
	\includegraphics[width=.8\textwidth]{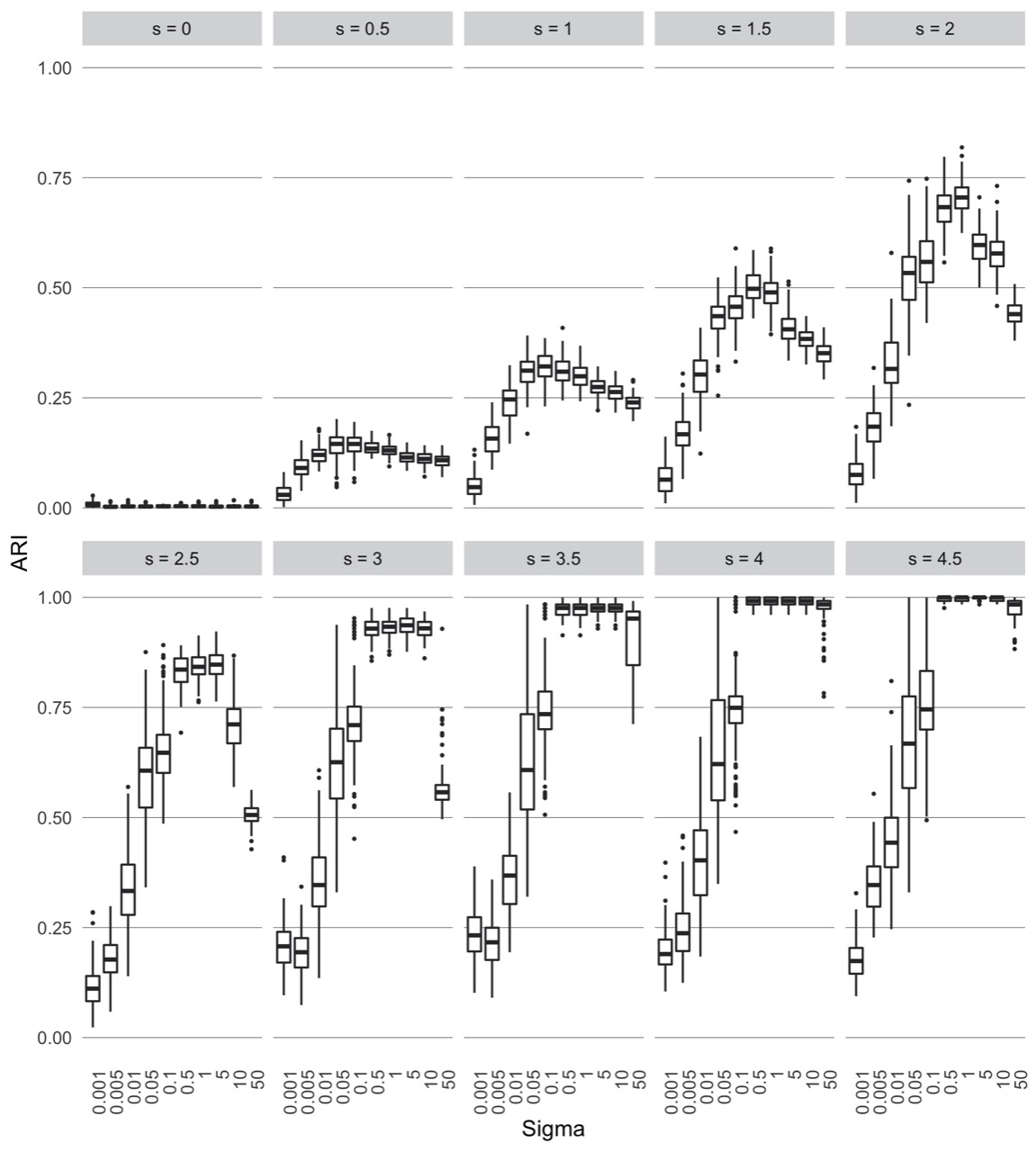}
	\caption{ARI between the clusters obtained with an kernel $k$-means on RBF kernels for different values of the characteristic length scale parameter and separation between clusters.}
	\label{fig:rbf-ari}
\end{figure}

\clearpage
\subsection{Additional simulation settings}
\label{sec:additional-simulation-settings}

We present here some additional simulation settings that were omitted from the main paper for the sake of brevity.

\subsubsection{Datasets with nested clusters}

\begin{figure}[h]
	\centering
	\includegraphics[width=.48\textwidth]{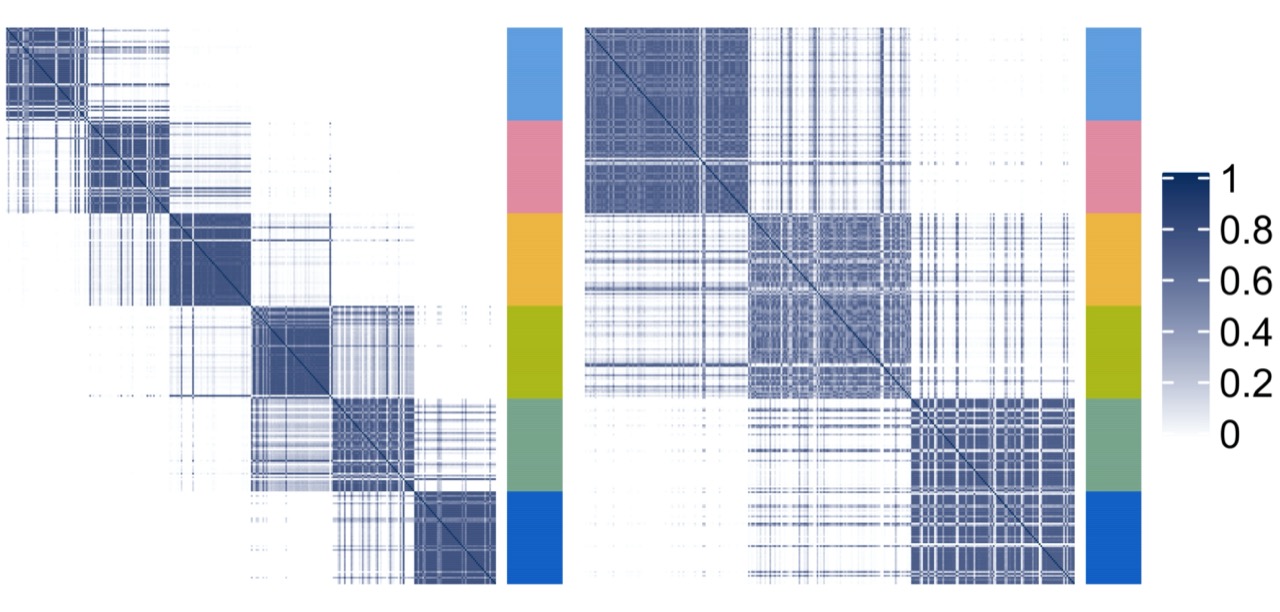}
	\caption{Consensus matrices of the synthetic data. Blue indicates high similarity. The colours of the bar \textcolor{black}{to the right} of each matrix indicate the cluster labels. Consensus matrices of two datasets with nested clusters: the one on the left has six clusters, whereas the one on the right has three clusters formed by merging two of the clusters of the dataset with six clusters.}
	\label{fig:consensus-matrices-simulations}
\end{figure}

We investigate how the algorithm copes with the ambiguous situation of nested clusters. To this end, we generate two datasets with the same value of the parameter $s$ setting the distance between cluster centres. The first one has six clusters, while the second one only has three clusters, each of them containing two of the clusters of the other dataset (Figure \ref{fig:consensus-matrices-simulations}).  

Since the algorithm works only with a fixed number of clusters, we try both with $K = 3$ and $K = 6$.
The ARI and the average weights assigned to each matrix are reported in Figure \ref{fig:simulation-nestedclusters}.
For $K=6$, the weights assigned to each matrix are not as we expected: the matrix with three clusters is weighted slightly more highly than the other one. To investigate this phenomenon, we introduce an additional way to score how strong the signal is in each dataset. We use the \emph{cophenetic correlation coefficient}, a measure of how faithfully hierarchical clustering would preserve the pairwise distances between the original data points \textcolor{black}{\citep{Sokal1962comparison, Brunet2004metagenes}}.
Given a dataset $X= [\boldsymbol{x}_1, \boldsymbol{x}_2, \dots, \boldsymbol{x}_N]$ and a similarity matrix $\Delta \in \mathbb{R}^{N\times N}$, we define the \emph{dendrogrammatic distance} between $\boldsymbol{x}_i$ and $\boldsymbol{x}_j$ as the height of dendrogram at which these two points are first joined together by hierarchical clustering and we denote it by  $\eta_{ij}$. The cophenetic correlation coefficient $\rho$ is calculated as 
\begin{equation}
\rho = \frac{\Sigma_{i<j} (\Delta_{ij} - \bar{\Delta})(\eta_{ij} - \bar{\eta})}{\sqrt{\sum_{i < j} (\Delta_{ij} - \bar{\Delta}) \Sigma_{i<j} (\eta_{ij} - \bar{\eta})}},
\end{equation}
where $\bar{\Delta}$ and $\bar{\eta}$ are the average values of $\Delta_{ij}$ and $\eta_{ij}$ respectively.
The cophenetic correlation coefficient of a consensus matrix can be interpreted as an indication of the level of its dispersion or, equivalently, of the stability of the  clustering used in CC. If the clusters are invariant under subsampling of the data features/observations, then the consensus matrix has all entries equal to either one or zero, and cophenetic correlation coefficient equal to one. On the other hand, if clusters vary at each iteration of consensus clustering, the entries of the consensus matrix are scattered between zero and one, and the corresponding cophenetic correlation coefficient is negative.  The consensus matrices shown in Figure 1 of the main paper, for instance, have increasing cophenetic correlation going from left (lower cluster separability) to right (higher cluster separability).
We find that in this case the consensus matrices with $K=3$ have slightly higher cophenetic correlation than the ones with $K=6$ with the same level of cluster separability $s$. This explains why higher weights are assigned to the former. This suggests that, in ambiguous cases, localised kernel k-means assigns higher weights (on average) to the kernels with highest cophenetic correlation. Intuitively, the sum of within-cluster distances in the feature space is zero when each pair of data points has similarity one if both data points are in the same cluster, and zero otherwise. Minimising that sum thus corresponds to finding the weights and cluster allocations that lead to a weighted kernel that is as close as possible to a kernel with cophenetic correlation one. 

\begin{figure}[h!]
	\centering
	
	\begin{subfigure}[b]{\textwidth}
	\centering
	\includegraphics[width=.3\textwidth]{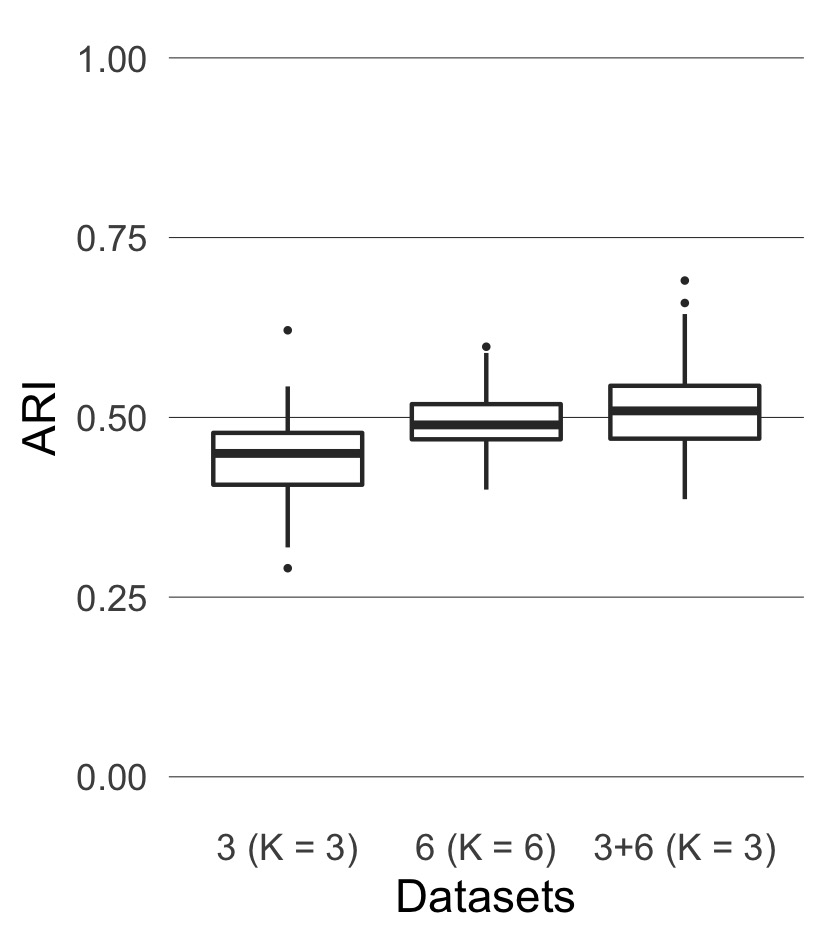}
	\includegraphics[width=.15\textwidth]{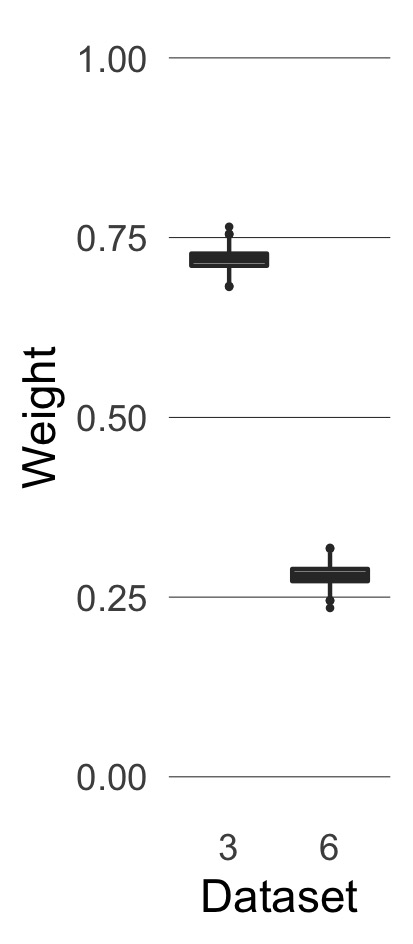}
	\includegraphics[width=.15\textwidth]{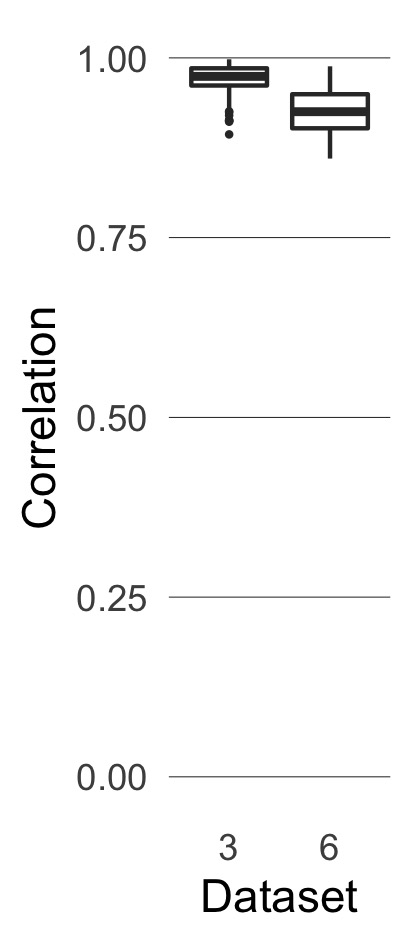}
	\caption{True number of clusters for CC, $K = 3$ for global clustering.}
	\end{subfigure}
	
	\begin{subfigure}[b]{\textwidth}
	\centering
	\includegraphics[width=.3\textwidth]{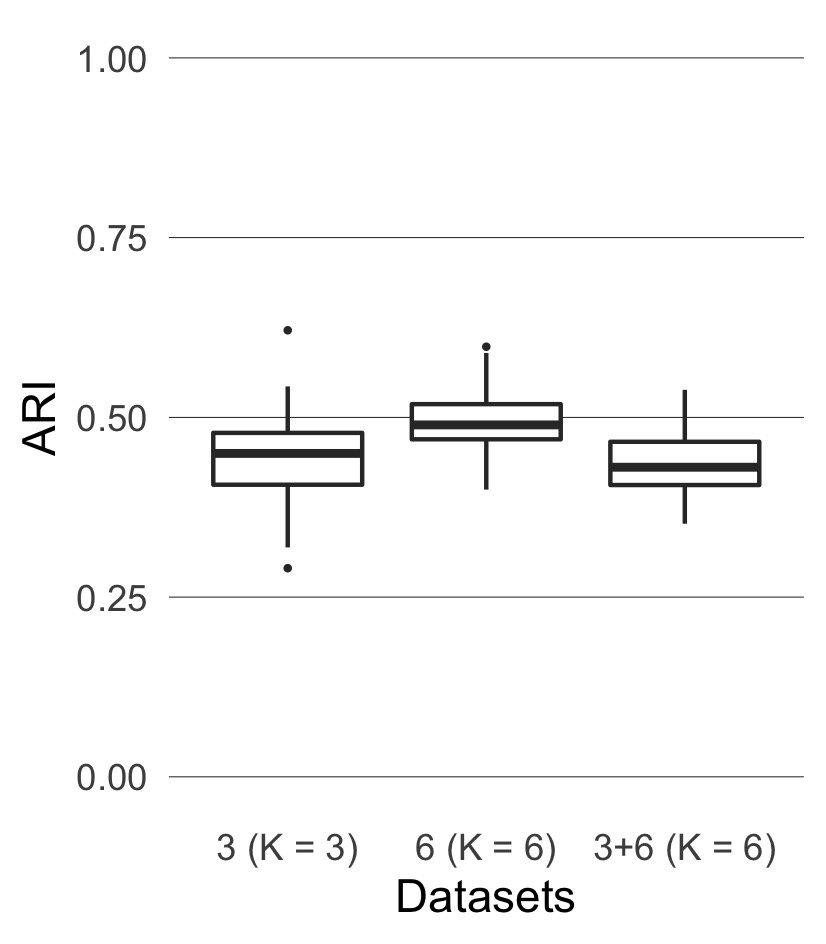}
	\includegraphics[width=.15\textwidth]{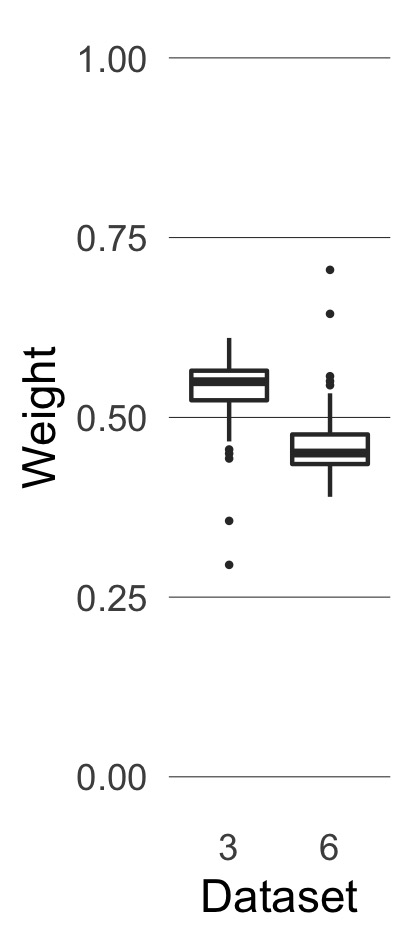}
	\includegraphics[width=.15\textwidth]{s4c-cophene-new-kmeans-bis}
	\caption{True number of clusters for CC, $K = 6$ for global clustering.}
	\end{subfigure}
	
	\caption{Results of applying KLIC to datasets that have nested clusters. Left: ARI of KLIC applied to the datasets with three and six clusters separately (columns ``3'' and ``6'' respectively) and to those two datasets combined (column ``3+6''). Centre: the weights assigned to each dataset. Right: cophenetic correlation coefficients of the consensus matrices built with $K=3$ (for the dataset with three clusters) and $K=6$ (for the dataset with six clusters). Higher weights are given to the kernels with higher cophenetic correlation, irrespectively of their number of clusters.}
	\label{fig:simulation-nestedclusters-bis}
\end{figure}

\clearpage
We also report the results obtained setting either $K=3$ or $K=6$ at each step of KLIC, i.e. consensus clustering of each dataset and MKL.

\begin{figure}[h!]
	\centering
	\begin{subfigure}[b]{\textwidth}
	\centering
	\includegraphics[width=.3\textwidth]{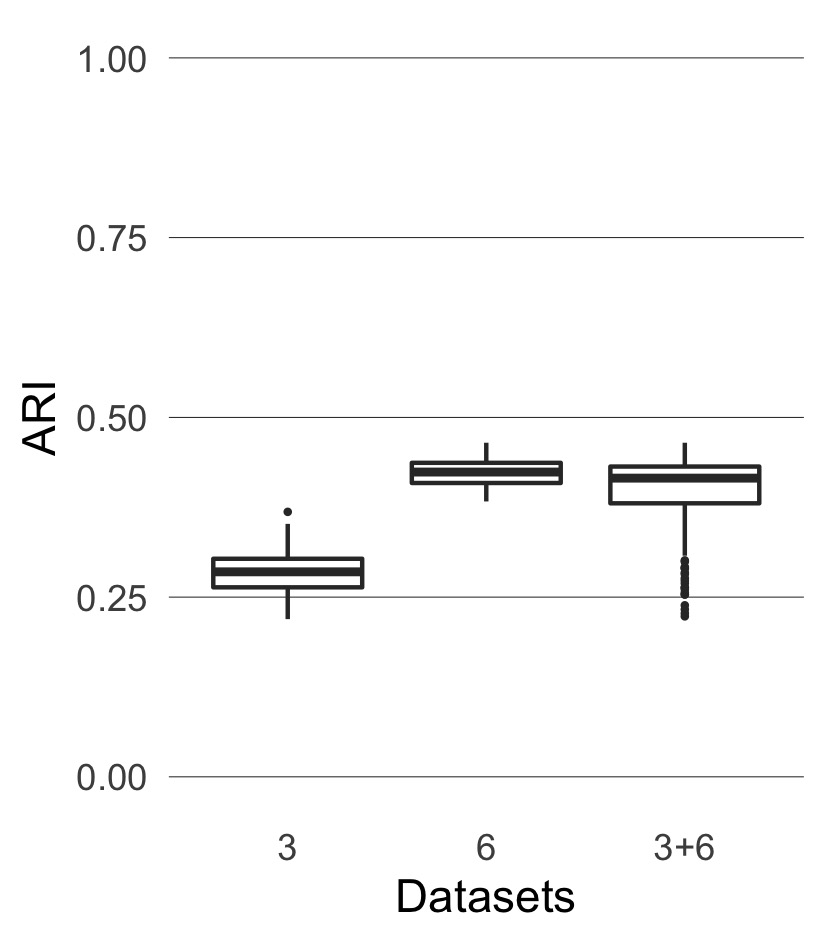}
	\includegraphics[width=.15\textwidth]{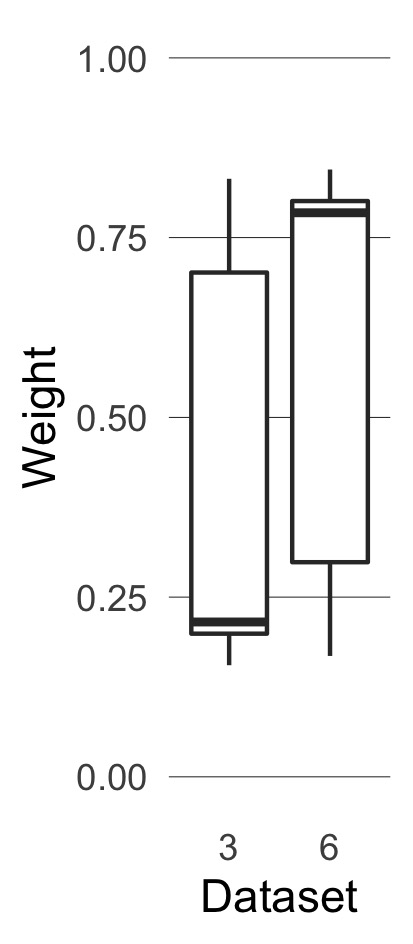}
	\includegraphics[width=.15\textwidth]{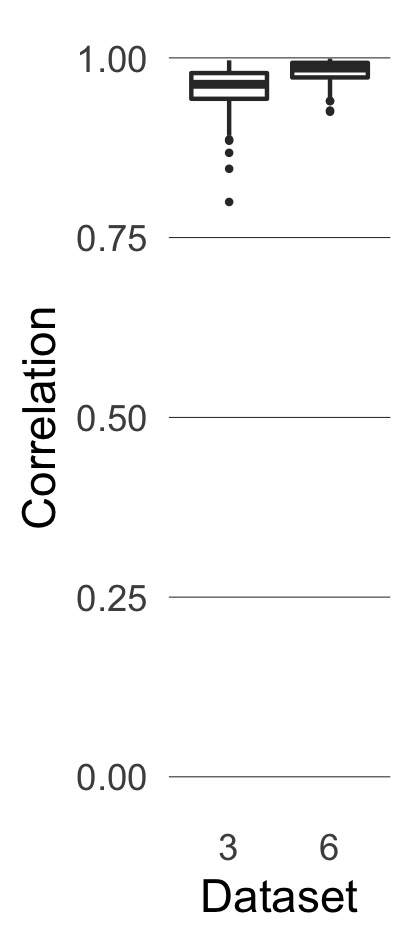}
	\caption{$K = 3$ at each step.}
	\end{subfigure}
	
	\begin{subfigure}[b]{\textwidth}
	\centering
	\includegraphics[width=.3\textwidth]{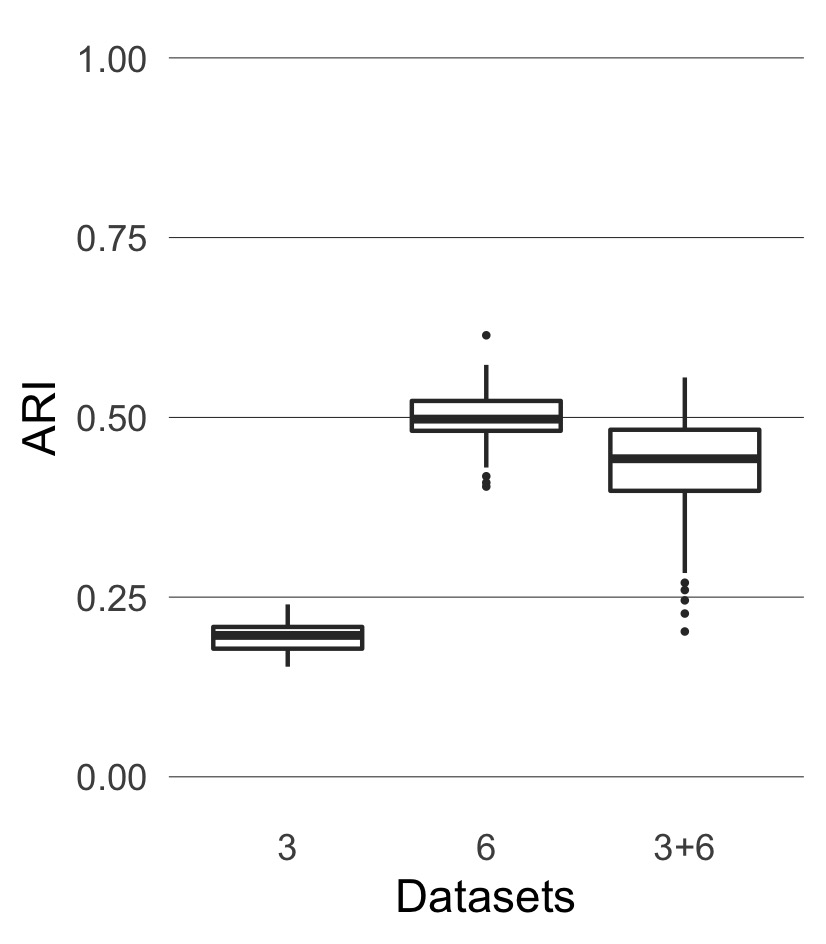}
	\includegraphics[width=.15\textwidth]{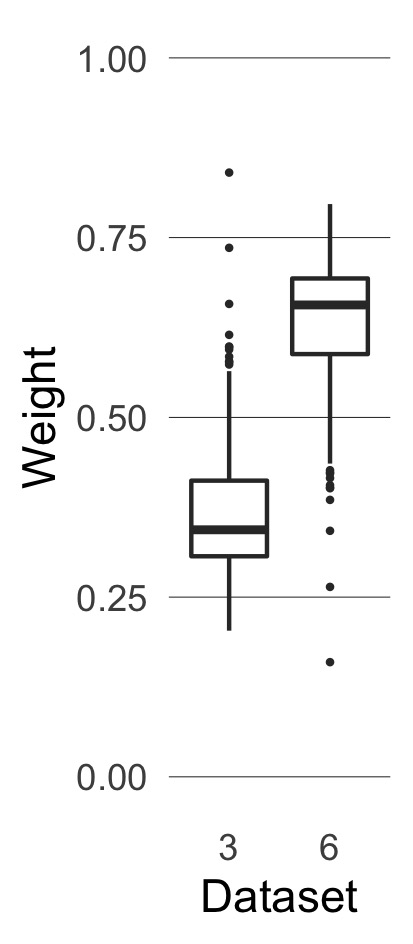}
	\includegraphics[width=.15\textwidth]{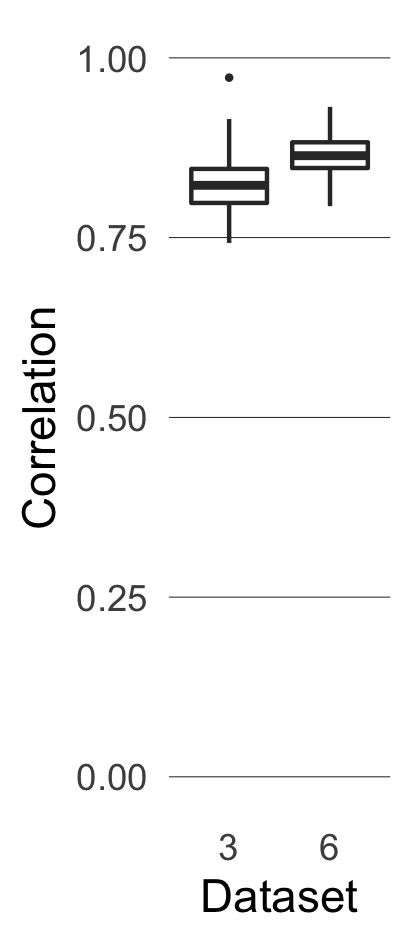}
	\caption{$K=6$ at each step.}
	\end{subfigure}
	
	\caption{Results of applying KLIC to datasets that have nested clusters. Left: ARI of KLIC applied to the datasets with three and six clusters separately (columns ``3'' and ``6'' respectively) and to those two datasets combined (column ``3+6''). Centre: the weights assigned to each dataset. Right: cophenetic correlation coefficients of the consensus matrices built with $K=3$ (top) and $K=6$ (bottom).  Higher weights are given to the kernels with higher cophenetic correlation, irrespectively of their number of clusters.}
	\label{fig:simulation-nestedclusters}
\end{figure}

\subsubsection{Comparison between KLIC, COCA, and other methods}
\begin{figure}
	\centering
	\begin{subfigure}[b]{.31\textwidth}
		\includegraphics[width=\textwidth]{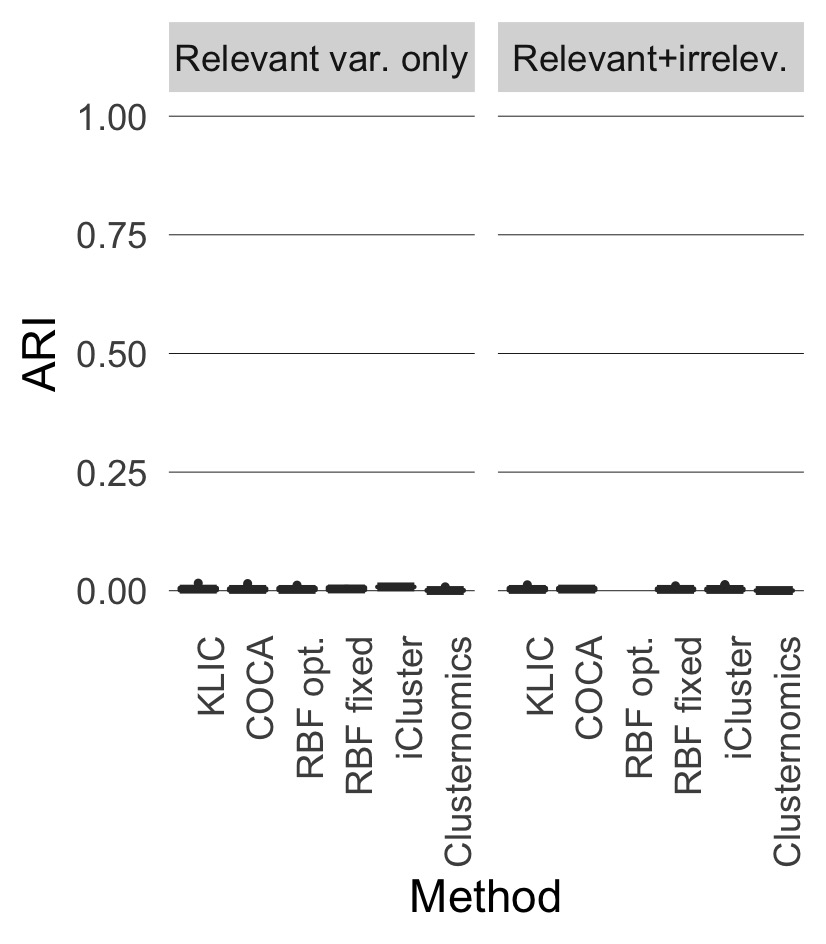}
		\caption{$s=0$.}
	\end{subfigure}
	\begin{subfigure}[b]{.31\textwidth}
		\includegraphics[width=\textwidth]{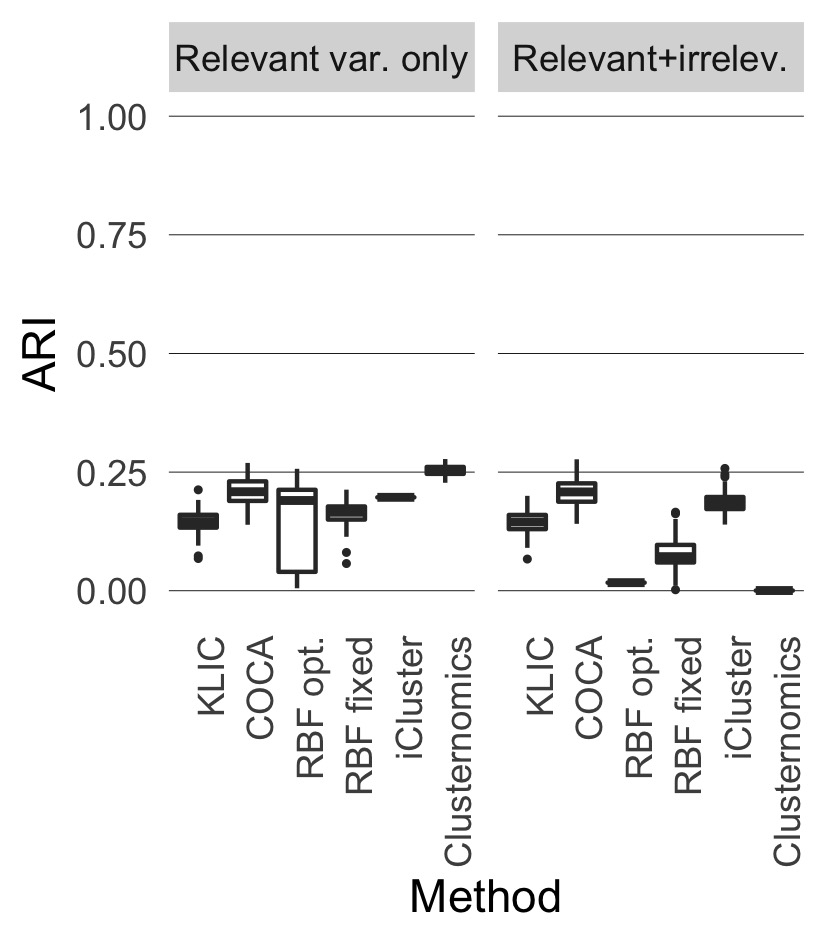}
		\caption{$s=0.5$.}
	\end{subfigure}
		\begin{subfigure}[b]{.31\textwidth}
		\includegraphics[width=\textwidth]{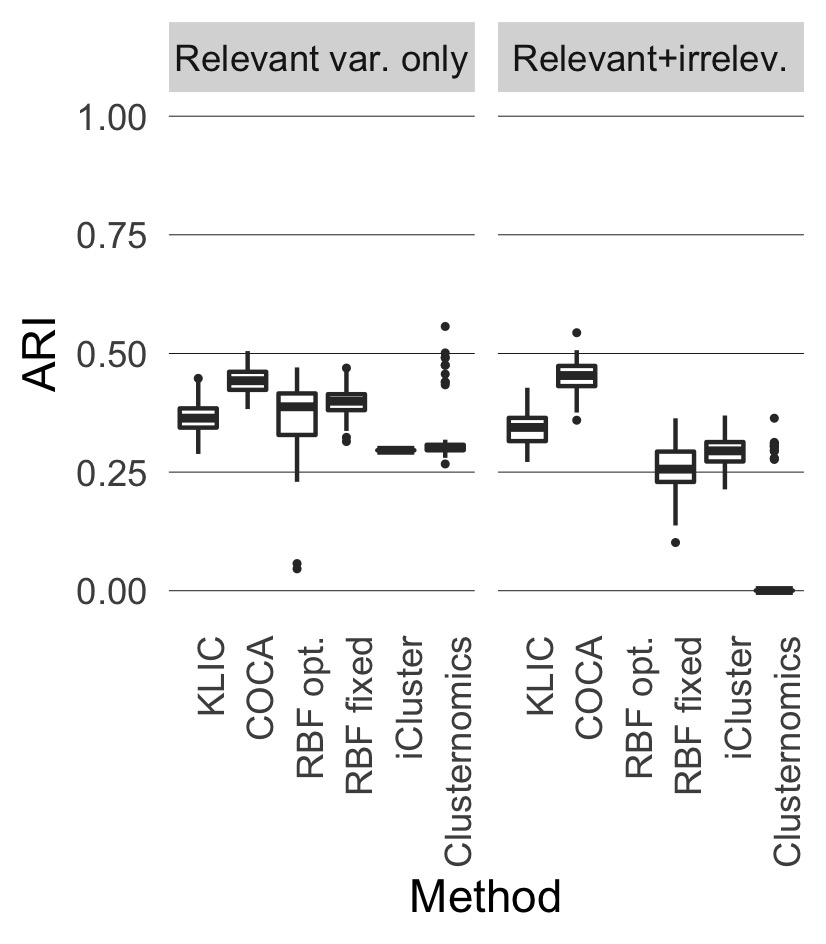}
		\caption{$s=1$.}
	\end{subfigure}
	\begin{subfigure}[b]{.31\textwidth}
		\includegraphics[width=\textwidth]{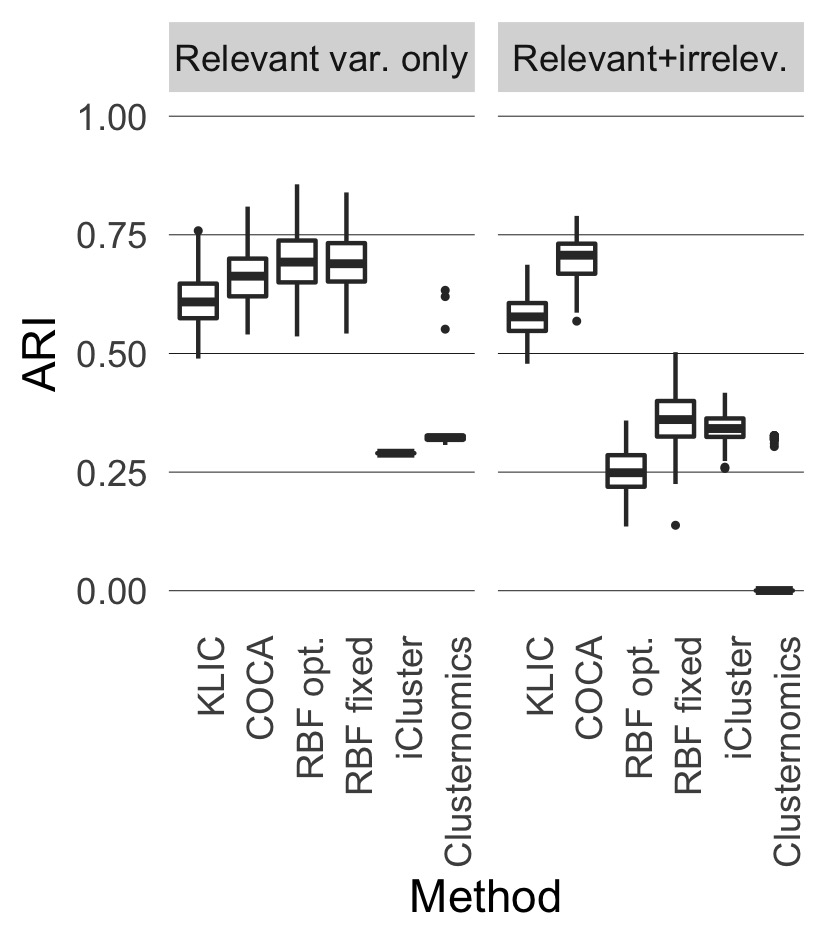}
		\caption{$s=1.5$.}
	\end{subfigure}
		\begin{subfigure}[b]{.31\textwidth}
		\includegraphics[width=\textwidth]{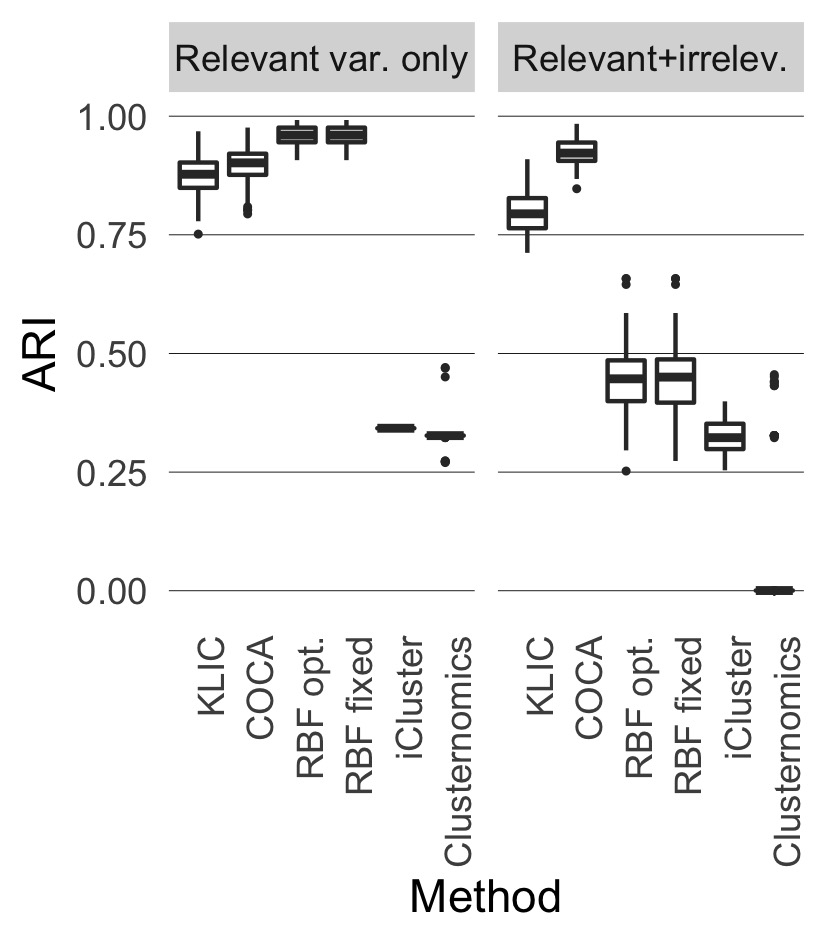}
		\caption{$s=2$.}
	\end{subfigure}
	\begin{subfigure}[b]{.31\textwidth}
		\includegraphics[width=\textwidth]{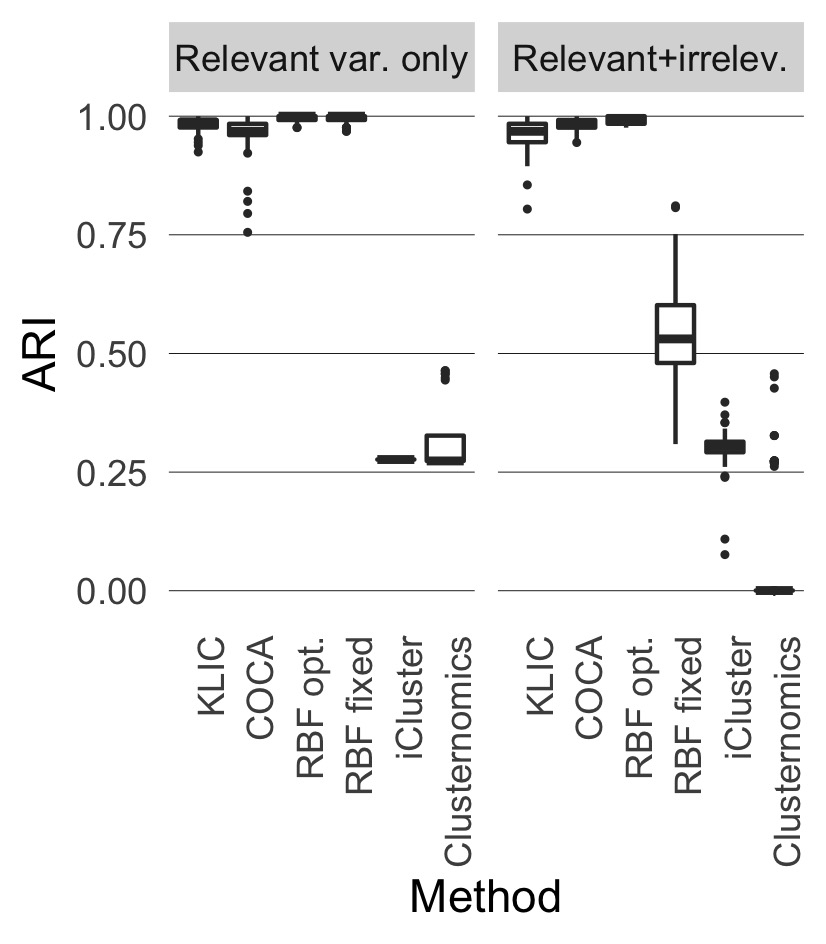}
		\caption{$s=2.5$.}
	\end{subfigure}
	\begin{subfigure}[b]{.31\textwidth}
		\includegraphics[width=\textwidth]{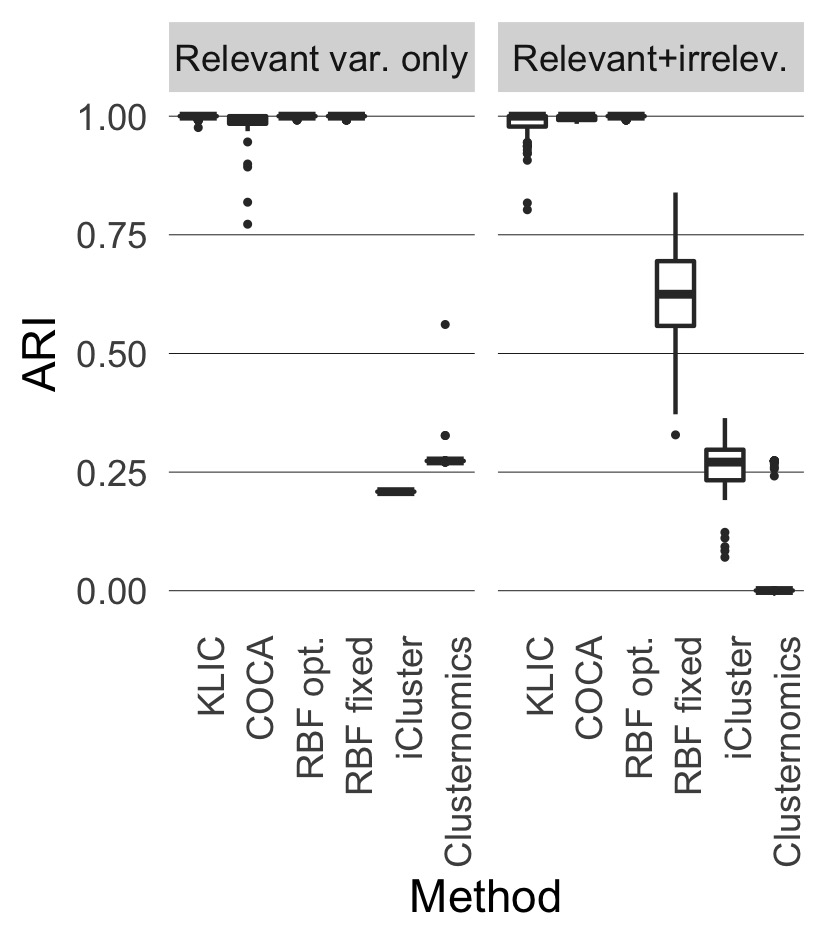}
		\caption{$s=3$.}
	\end{subfigure}
	\begin{subfigure}[b]{.31\textwidth}
		\includegraphics[width=\textwidth]{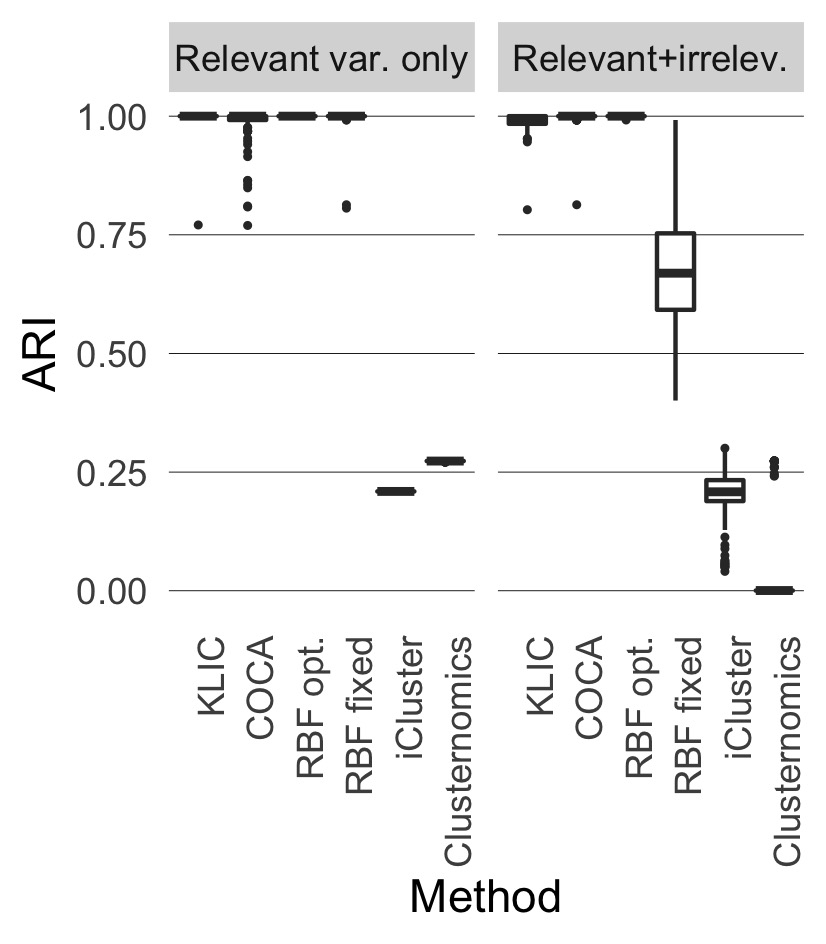}
		\caption{$s=3.5$.}
	\end{subfigure}
		\begin{subfigure}[b]{.31\textwidth}
		\includegraphics[width=\textwidth]{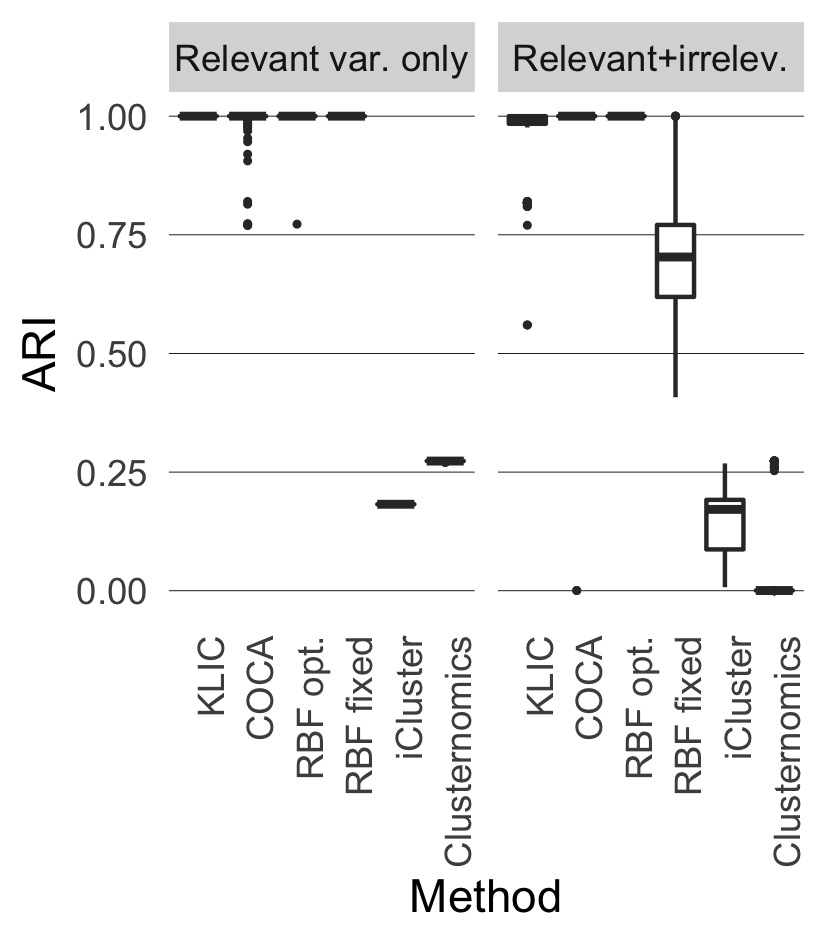}
		\caption{$s=4$.}
	\end{subfigure}
 \caption{Comparison between KLIC, COCA, and other clustering algorithms. ARI obtained using four datasets having the same clustering structure and cluster separability (as in Figure 2).}
 \label{fig:comparison-different-separation-levels}
\end{figure}

For simulation setting 1 (four datasets with the same level of cluster separability) only the results obtained with $s = 1.5$ are reported in the main paper. For completeness, we show here the corresponding figures for a range of other values of $s$ in Figure \ref{fig:comparison-different-separation-levels}.

\clearpage
\subsubsection{Sensitivity analysis}

\begin{figure}[h!]
	\centering
	\begin{subfigure}[b]{.31\textwidth}
		\centering
		\includegraphics[width=\textwidth]{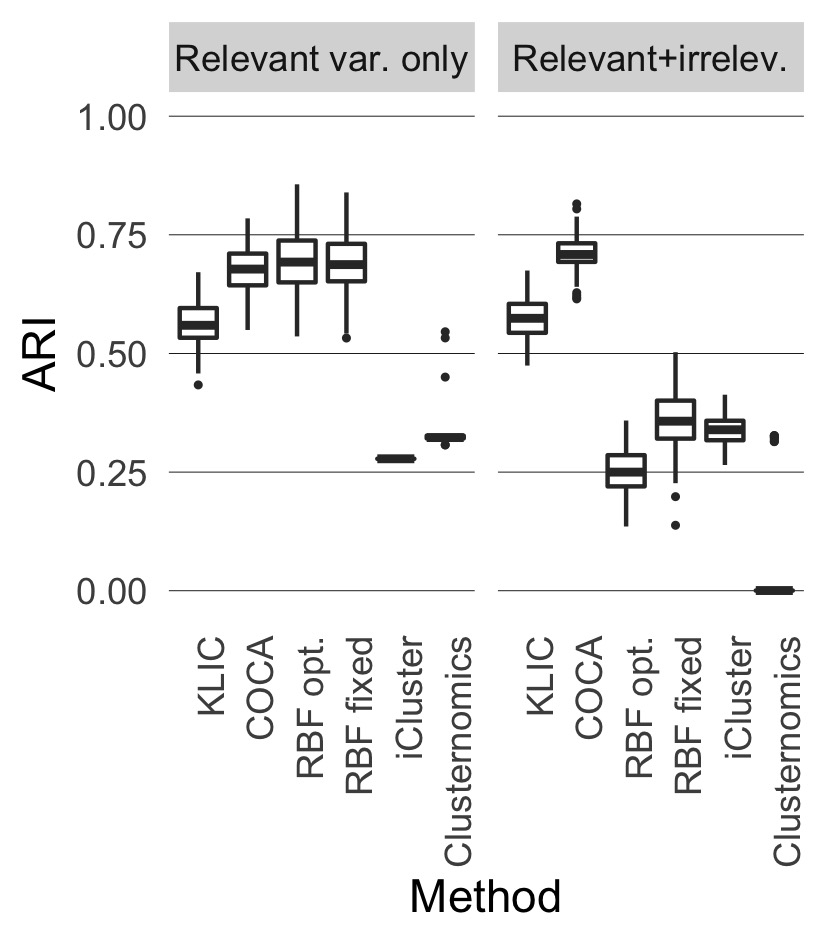}
	\end{subfigure}
	
	\begin{subfigure}[b]{.62\textwidth}
		\centering
		\includegraphics[width=\textwidth]{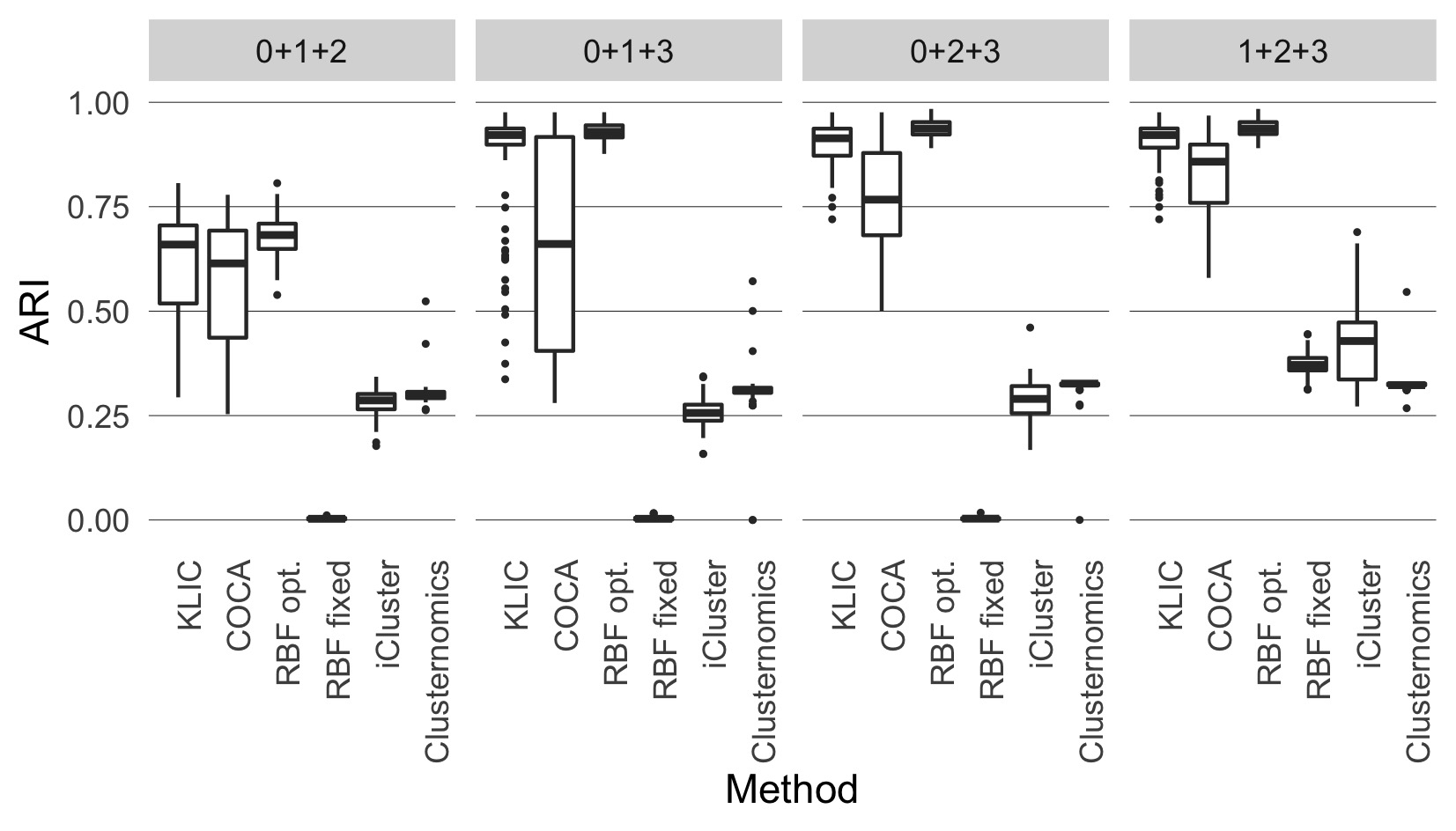}
	\end{subfigure}
	\caption{Comparison between KLIC, COCA, and other clustering algorithms.  The labels `RBF opt.' and `RBF fixed' refer to the MKL method using an RBF kernel with either $\sigma$ optimised or fixed at 1. Top: ARI obtained with each clustering algorithm using four datasets having the same clustering structure and cluster separability (as in Figure 2 in the main paper). Bottom: ARI obtained with COCA and KLIC for each of the subsets of heterogeneous datasets considered in Figure 3 in the main paper. The high ARI obtained with KLIC in all settings shows the advantage of using this method, especially when some of the datasets are noisy.}
	\label{fig:sensitivity-analysis}
\end{figure}

The results presented in the main paper were obtained with the parameter \verb|nstart| of the \verb|kmeans| function (which determines the number of random initialisations of the algorithm) set to one both for KLIC and COCA. Figure \ref{fig:sensitivity-analysis} shows the ARI obtained for the same simulation settings as in Figure 4 in the main paper, with the \verb|nstart| parameter set to 20. The figure shows that COCA is quite sensitive to the choice of this parameter, while KLIC is not.  This explains the difference observed in Figure 4 of the main paper between the ARI of COCA obtained when using $k$-means and sparse $k$-means, since those two methods have different default values of \verb|nstart|.

\clearpage

\section{Multiplatform analysis of 12 cancer types}
\label{sec:pancan12}

\textcolor{black}{In Section \ref{sec:pancan12-hoadley} we explain the steps we took to try to replicate the data preprocessing and cluster analysis of \cite{Hoadley2014multiplatform}. In Section \ref{sec:pancan12-klic} we give more details on the input and output of KLIC for this particular application.}

\subsection{Replicating the analysis of \cite{Hoadley2014multiplatform}}
\label{sec:pancan12-hoadley}

For each type of data we followed as closely as possible the procedures presented in the supplementary material of \cite{Hoadley2014multiplatform}. We present here the steps that we followed. The malignancies and corresponding acronyms considered in this study are: glioblastoma multiforme (GBM), serous ovarian carcinoma (OV), colon (COAD) and rectal (READ) adenocarcinomas, lung squamous cell carcinoma (LUSC), breast cancer (BRCA), acute myelogenous leukemia (AML), endometrial cancer (UCEC), renal cell carcinoma (KIRC), and bladder urothelial adenocarcinoma (BLCA).  The agreement between the clustering analysis presented here and the clustering presented in the original Hoadley {\em et al.} paper ranged from excellent (for the protein and mRNA datasets) to quite poor (for the miRNA dataset).

\begin{figure}[!ht]
	 \makebox[\textwidth][c]{\includegraphics[width=1.3\textwidth]{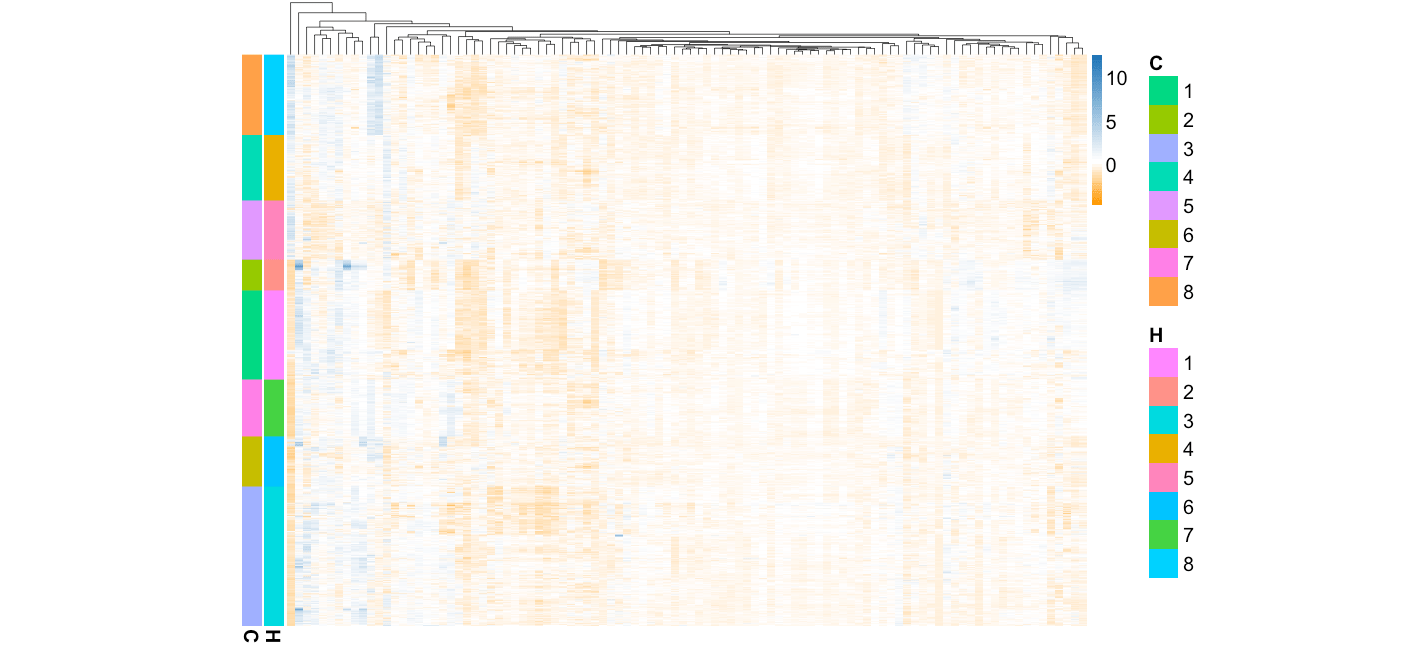}}%
	\caption{Protein expression clusters.  High values are indicated in blue and low values in orange. C: Clusters found in this analysis.  H: Clusters found in the original analysis of Hoadley {\em et al.} Adjusted Rand index between C and H:  1.}
	\label{fig:pancan12-protein-expression-clusters}
\end{figure}

\paragraph{Protein expression}
We used hierarchical clustering with Ward's agglomeration method and  Pearson's correlation as the distance. Our clusters match exactly those of Hoadley {\em et al.} (i.e. the ARI is equal to one, see Figure \ref{fig:pancan12-protein-expression-clusters}).

\begin{figure}[!ht]
	\centering
     \includegraphics[width=\textwidth]{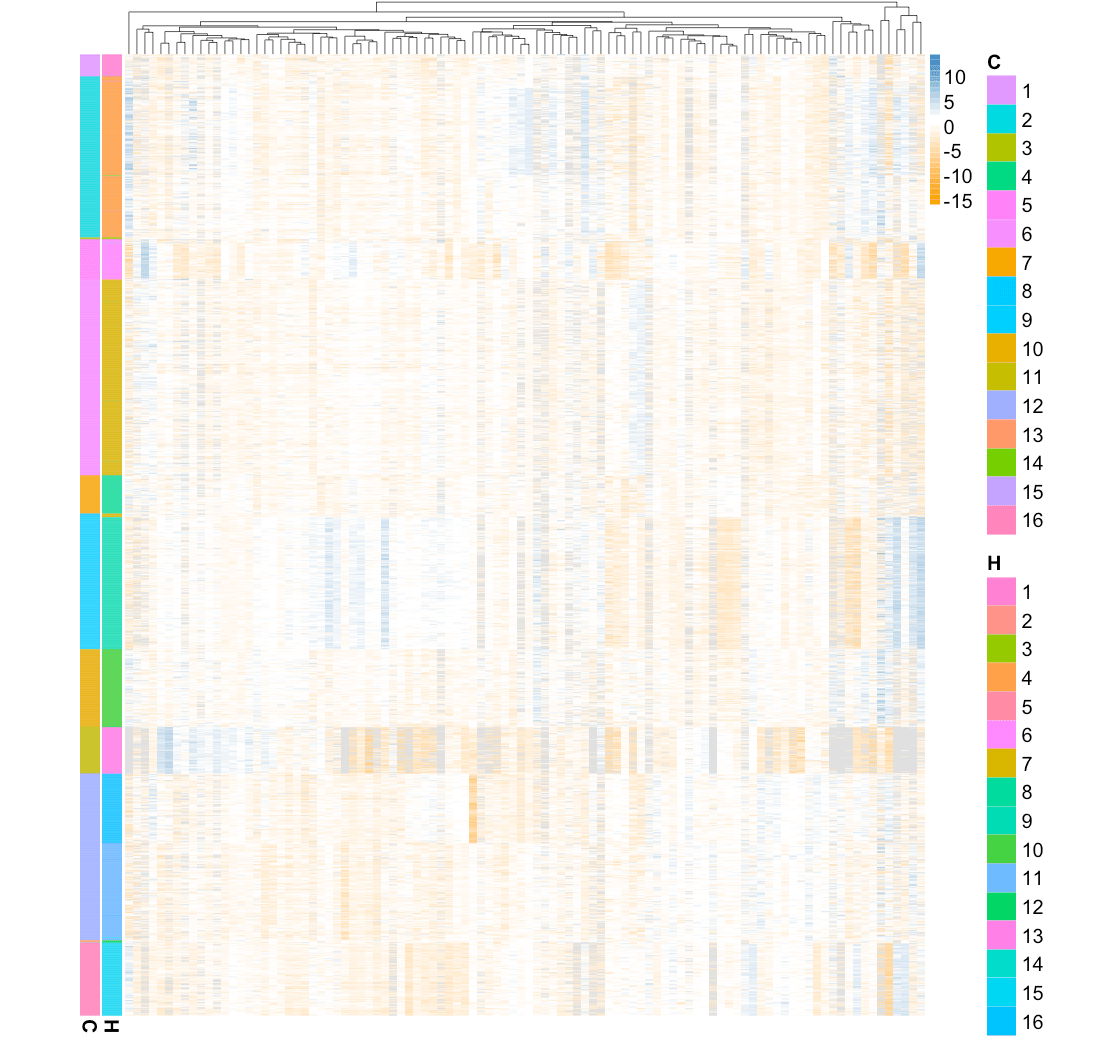}
	\caption{mRNA expression clusters. High values are indicated in blue and low values in orange. The dataset contains 600 genes but here we show only 100 of them. C: Clusters found in this analysis.  H: Clusters found in the original analysis of Hoadley {\em et al.} Adjusted Rand index between C and H:  0.917.}
	\label{fig:pancan12-mrna-expression-clusters}
\end{figure}

\newpage
\paragraph{mRNA expression}
For mRNA expression, we proceeded as indicated by \cite{Hoadley2014multiplatform}. We chose the genes present in 70\% of samples and then selected the 6,000 most variable genes. Then we used the ConsensusClusterPlus R package with settings \verb|maxK=20|, \verb|innerLinkage| \verb|="average"| \verb|finalLinkage="average"|, \verb|distance="pearson"|, \verb|corUse="pairwise.complete.obs"|. The ARI is 0.917 (see Figure \ref{fig:pancan12-mrna-expression-clusters}).

\begin{figure}[!ht]
	\centering
	\includegraphics[width=0.9\textwidth]{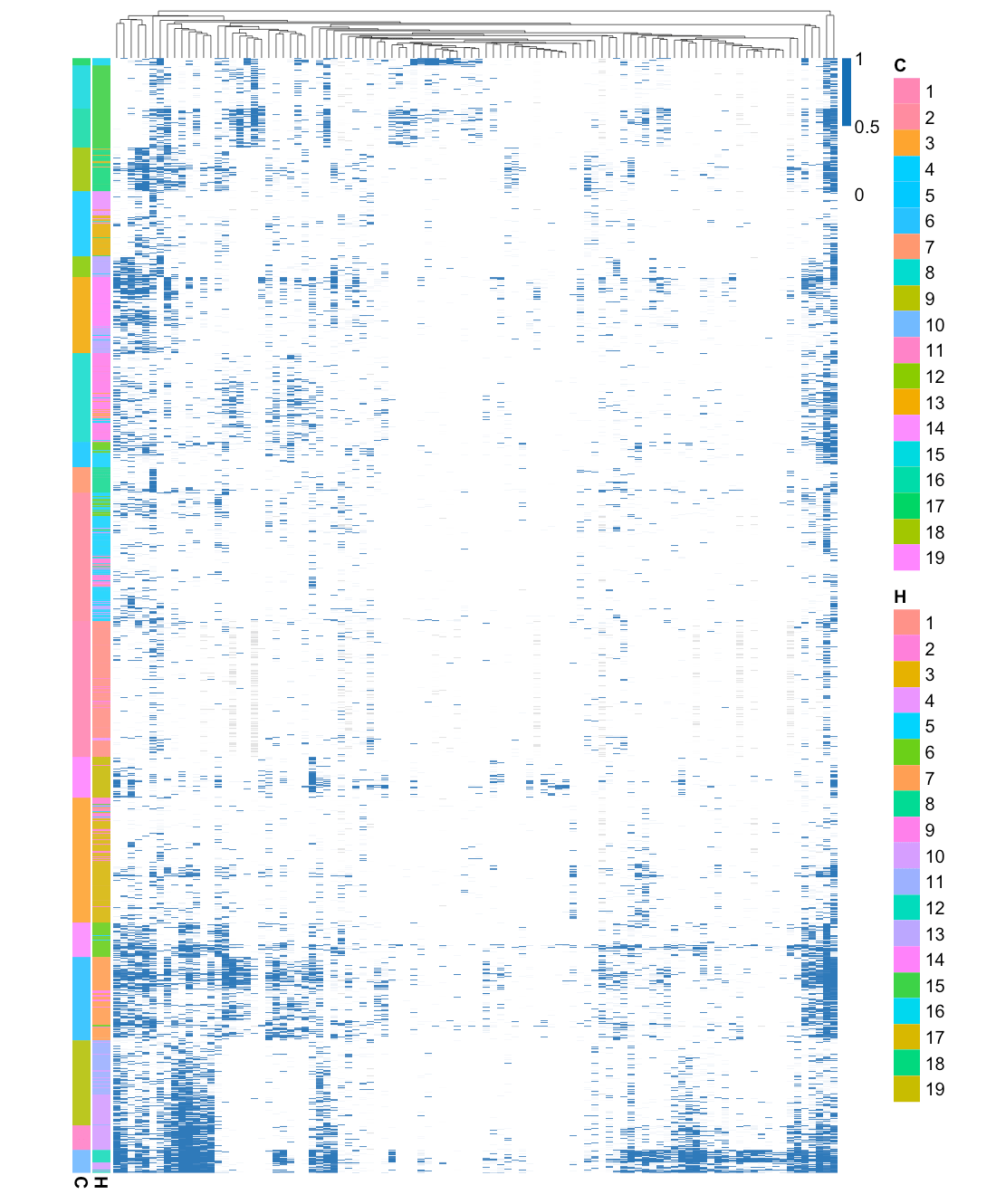}
	\caption{DNA methylation clusters. Blue cells correspond to methylated loci. Missing values are indicated in grey colour. Only 100 CpG loci are shown here, but the full dataset contains 2,043.  C: Clusters found in this analysis.  H: Clusters found in the original analysis of Hoadley {\em et al.} Adjusted Rand index between C and H:  0.680.}
	\label{fig:pancan12-dna-methylation-clusters}
\end{figure}

\newpage
\paragraph{DNA methylation}
\textcolor{black} {We used hierarchical clustering with Jaccard's distance and Ward's agglomeration method. \cite{Hoadley2014multiplatform} chose to divide the data into 19 clusters, so we did the same. Comparing our clusters to those of \cite{Hoadley2014multiplatform}, we obtained an ARI of 0.680 (see Figure \ref{fig:pancan12-dna-methylation-clusters}).}

\begin{figure}[!ht]
	\centering
	 \includegraphics[width=.84\textwidth]{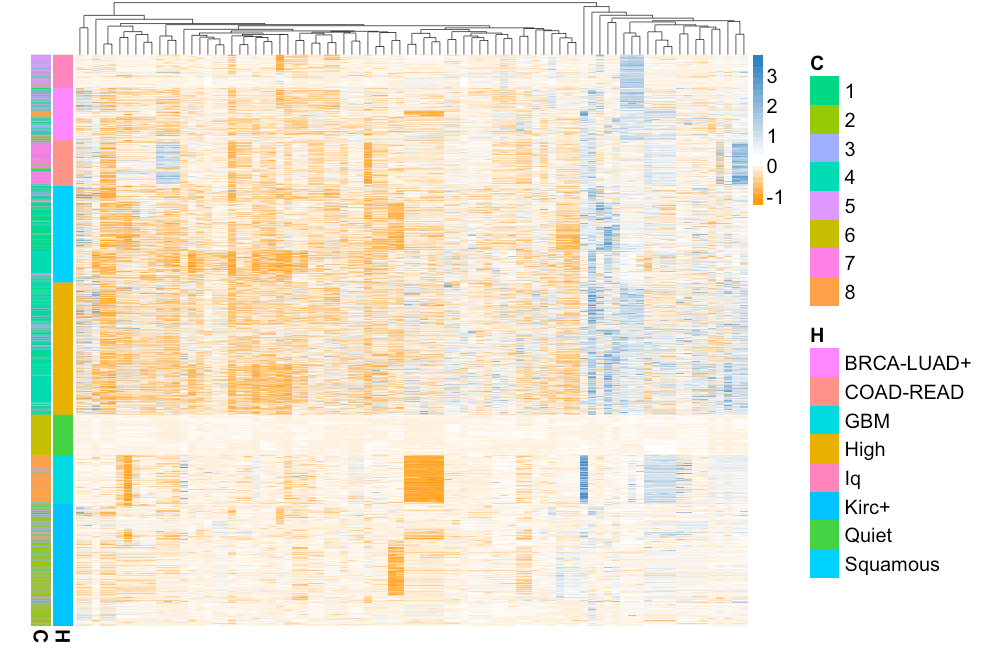}
	\caption{Somatic copy number clusters. High values are indicated in blue and low values in orange. C: Clusters found in this analysis.  H: Clusters found in the original analysis of Hoadley {\em et al.} Adjusted Rand index between C and H: 0.333.}
	\label{fig:pancan12-somatic-copy-number-clusters}
\end{figure}

\newpage
\paragraph{DNA copy number}
The clusters for the somatic copy number dataset were found using hierarchical clustering with Euclidean distance and Ward's method. The number of clusters was set to eight in the original manuscript based on the cophenetic distances and therefore we did the same here. The adjusted Rand index (ARI) comparing the clustering found in the present analysis with the clustering found in the original analysis of Hoadley {\em et al.} is 0.333 (see Figure \ref{fig:pancan12-somatic-copy-number-clusters}).

\begin{figure}[!ht]
	\centering
	\begin{subfigure}[b]{.5\textwidth}
	\includegraphics[width=\textwidth]{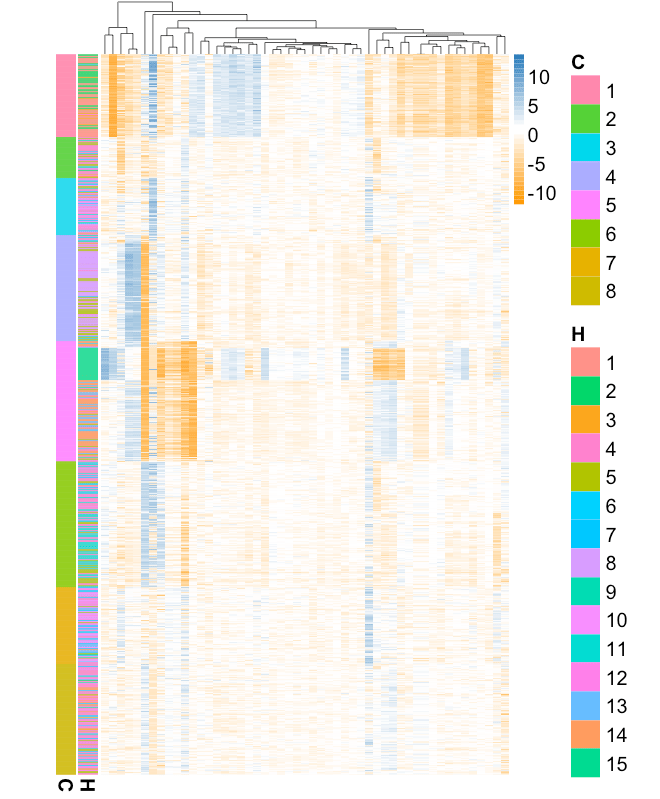}
	\caption{Clusters. High values are indicated in blue, low values in orange.}
	\end{subfigure}
	
	\begin{subfigure}[b]{.3\textwidth}
	\includegraphics[width=\textwidth]{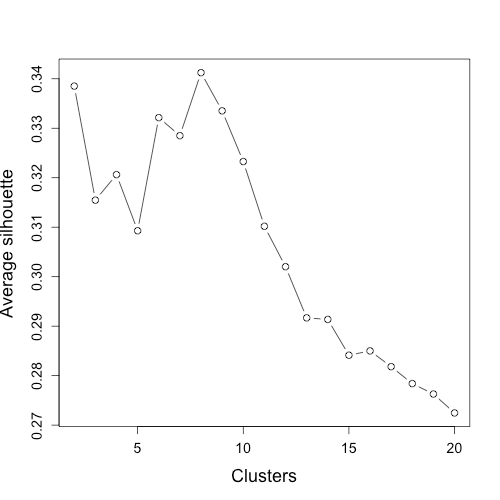}
	\caption{\textcolor{black}{Silhouette.}}
	\end{subfigure}
	\caption{microRNA expression. C: Clusters found in this analysis.  H: Clusters found in the original analysis of Hoadley {\em et al.} Adjusted Rand index between C and H: 0.255.}
	\label{fig:pancan12-microrna-expression-clusters}
\end{figure}

\newpage
\paragraph{microRNA expression}
In the original manuscript the clusters of the microRNA-seq data were determined using a software program called \textit{Cluster 3} \citep{deHoon2004open}. The same software was used to scale the data. Since it is was not possible to retrieve the clusters presented in the paper using this software, we used R to scale the data as was done by Cluster 3, namely applying a logarithmic transformation to the data and then median-centring. We found the final clusters using agglomerative hierarchical clustering in R (\verb|agnes| command. We selected the number of clusters that maximises the silhouette, which is eight. The ARI is 0.255 (see Figure \ref{fig:pancan12-microrna-expression-clusters}).

\subsection{Output of KLIC}
\label{sec:pancan12-klic}

\begin{figure}[!ht]
	\centering
	\captionsetup[subfigure]{justification=centering}
	\begin{subfigure}[b]{.30\textwidth}
		\centering
		\includegraphics[width=\textwidth]{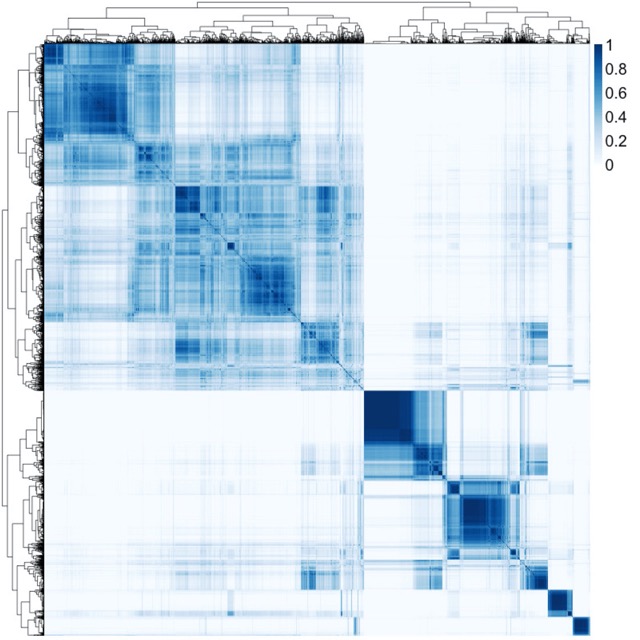}
		\caption{DNA copy number.\\
		Cophenetic correlation coefficient: 0.736.}
	\end{subfigure}
	\begin{subfigure}[b]{.30\textwidth}
		\centering
		\includegraphics[width=\textwidth]{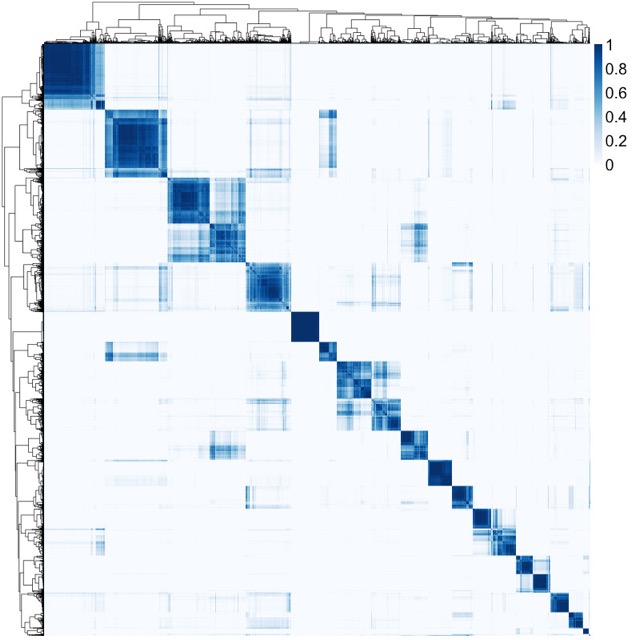}
		\caption{DNA methylation.\\
		Cophenetic correlation coefficient: 0.859.}
	\end{subfigure}
	\begin{subfigure}[b]{.30\textwidth}
		\centering
		\includegraphics[width=\textwidth]{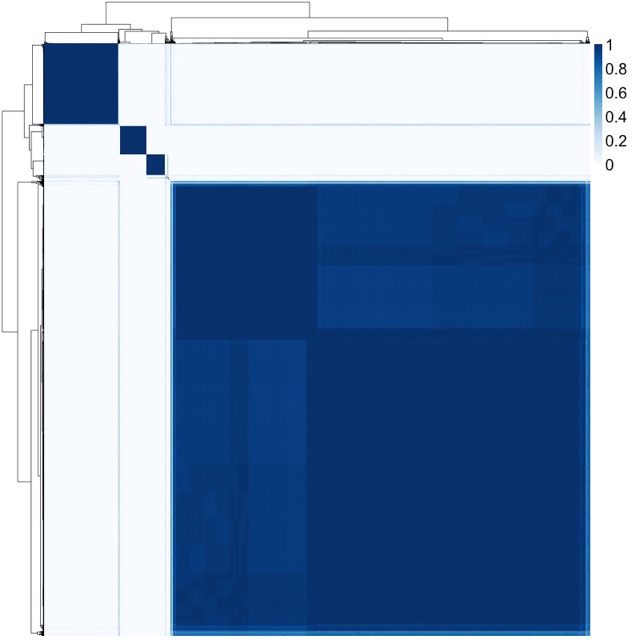}
		\caption{mRNA expression.\\
		Cophenetic correlation coefficient: 0.974.}
	\end{subfigure}
	\begin{subfigure}[b]{.30\textwidth}
		\centering
		\includegraphics[width=\textwidth]{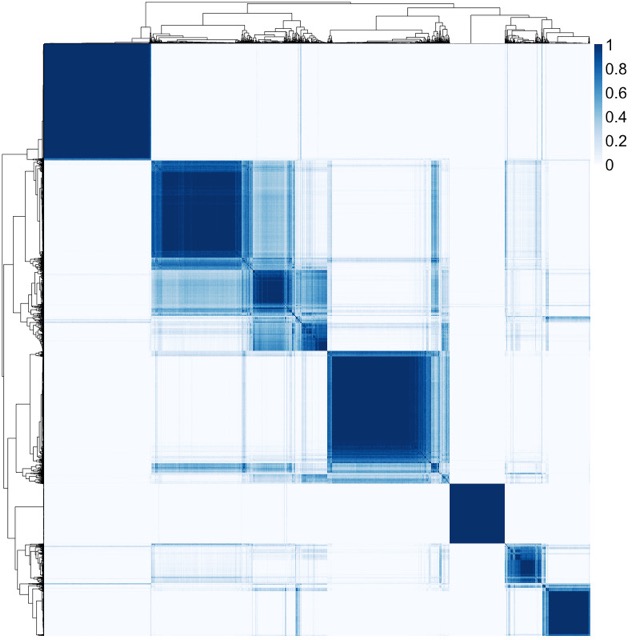}
		\caption{miRNA expression.\\
		Cophenetic correlation coefficient: 0.923.}
	\end{subfigure}
	\begin{subfigure}[b]{.30\textwidth}
		\centering
		\includegraphics[width=\textwidth]{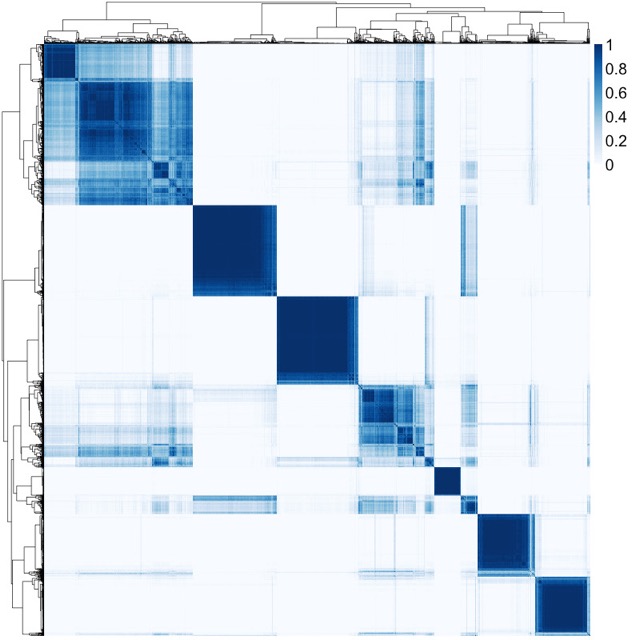}
		\caption{Protein expression.\\
		Cophenetic correlation coefficient: 0.888.}
	\end{subfigure}
	\caption{Kernel matrices.}
	\label{fig:pancan12-kernels}
\end{figure}

\begin{figure}[!ht]
\centering
	\captionsetup[subfigure]{justification=centering}
	
	\begin{subfigure}[b]{\textwidth}
		\centering
		\includegraphics[width=.75\textwidth]{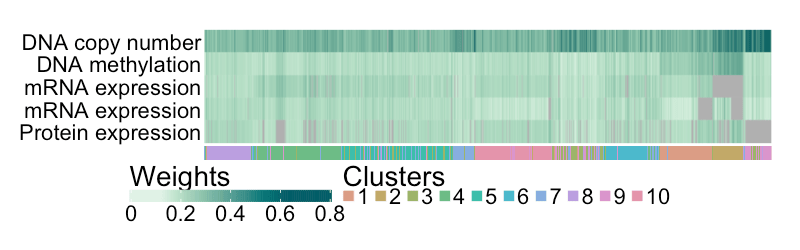}
		\caption{Weights.}
		\label{fig:pancan12-klic-weights}
	\end{subfigure}
	
	\begin{subfigure}[b]{.48\textwidth}
		\centering
		\includegraphics[width=\textwidth]{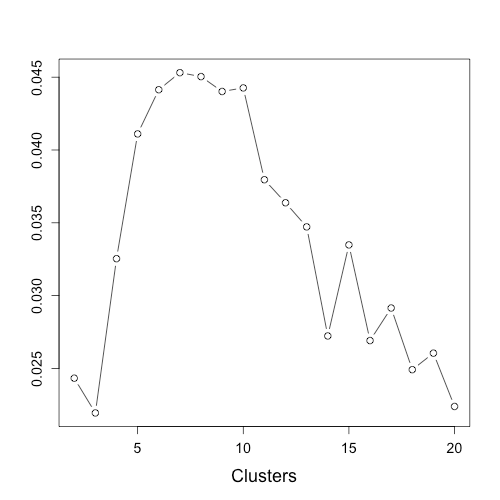}
		\caption{Average silhouette.}
		\label{fig:pancan12-klic-silhouette}
	\end{subfigure}
	\begin{subfigure}[b]{.48\textwidth}
		\centering
		\includegraphics[width=\textwidth]{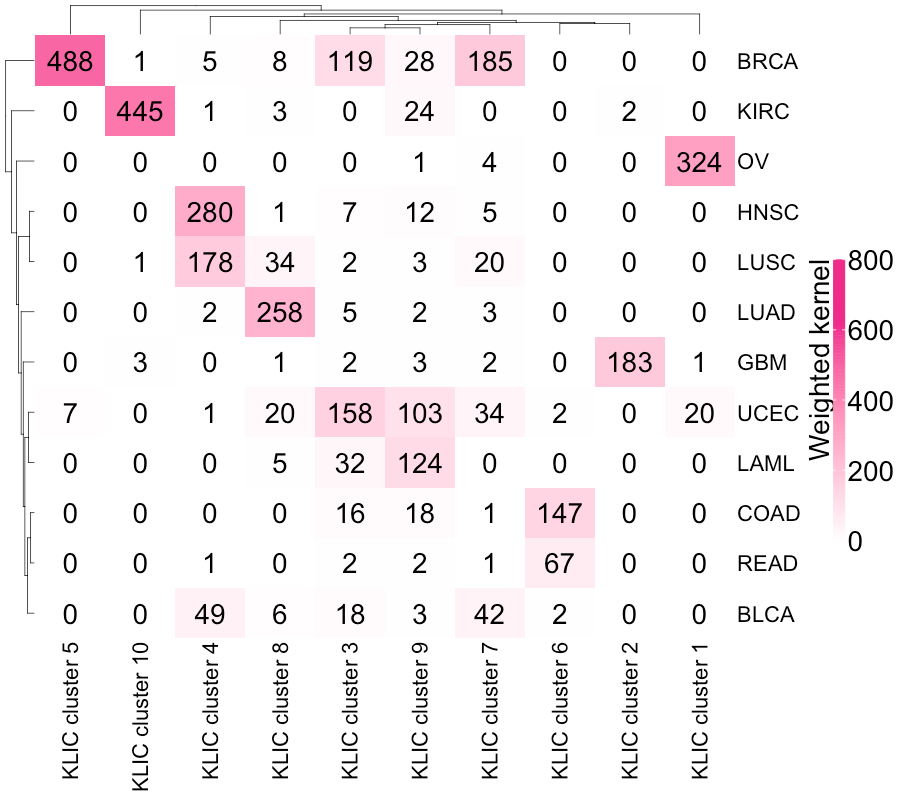}
		\caption{Matrix of coincidences.}
		\label{fig:pancan12-klic-coincidences}
	\end{subfigure}
	\caption{Output of KLIC. (a) Weights. Low weights are indicated in white and higher weights in green. Grey cells correspond to missing values, which have zero weight. (b) Average silhouette. The maximum is obtained for seven clusters. All numbers of clusters comprised between six and ten have similar values. (c) Matrix showing the correspondences between the clusters obtained by using KLIC and the tumour tissues.}
	\label{fig:pancan12-klic}
\end{figure}

The kernels corresponding to each dataset  are shown in Figure \ref{fig:pancan12-kernels}, for each of them we also report the cophenetic correlation coefficient. Figure \ref{fig:pancan12-klic-weights} shows the weights associated to each observation in each dataset. Figure \ref{fig:pancan12-klic-silhouette} shows the average silhouette for all the number of clusters considered: the optimal values are between six and ten. Finally, Figure \ref{fig:pancan12-klic-coincidences} shows the correspondences between the clusters obtained using KLIC and the tumour tissues. Most clusters correspond quite well with one or two tissue types (e.g. cluster 10 contains almost exclusively samples of renal cell carcinoma and cluster 6 contains colon and rectal adenocarcinomas), but not all.

\clearpage

\section{Transcriptional module discovery}
\label{sec:transcriptional-module-discovery}

This section is structured as follows. First, we give further details regarding the application of KLIC and COCA to transcriptional module discovery using Bayesian Hierarchical Clustering as the clustering algorithm for the ChIP data. Then, we consider other algorithms that could have been applied to this dataset and compare the new results with those reported in the main paper. Finally, we give more details about the choice of the number of clusters for PAM.

\subsection{Clustering algorithms for the ChIP data}
The ChIP dataset is quite sparse. The data were discretised so that only transcription factors that are believed with high confidence to be able to bind to a gene's promoter region are marked as ``ones''; all the others are ``zeros''. For this reason, in addition to BHC, we considered two clustering algorithms that are able to take into account this feature of the data. However, we show in Sections \ref{sec:chip-data-pam} and \ref{sec:chip-data-gbnp} that these methods often cluster genes with few transcription factors (i.e. observations for which most variables are zero) together, while the other genes end up in separate small clusters that are less stable under subsampling of the data. This leads to consensus matrices that have high cophenetic correlation coefficients but carry little information. We show that combining the corresponding kernels to that of the expression data does not always give more meaningful clustering solutions than those obtained on each data type separately. This highlights the importance of the kernel matrices as an intermediate diagnostic tool for KLIC, which can help choosing the right clustering algorithms.

\subsubsection{Bayesian Hierarchical Clustering}

Bayesian Hierarchical Clustering (BHC; \citealp{Heller2005bayesian}) is a method for agglomerative hierarchical clustering. The idea is that, similarly to classical agglomerative clustering algorithms, at the start each data point is considered as a different cluster; then, at each step, two clusters are merged. The main difference between classical hierarchical clustering and BHC is that in BHC merging is done based on Bayesian hypothesis testing, where the alternative hypotheses are ``all data in clusters $c_i$ and $c_j$ were generated from the same probabilistic model'' and ``the data in $c_i$ and $c_j$ has two or more clusters in it''. The pair of clusters that is selected for merging is the one with highest probability of the merged hypothesis.

\begin{figure}
	\centering
	\captionsetup[subfigure]{justification=centering}
	\begin{subfigure}[b]{.49\textwidth}
		\centering
		\includegraphics[width=\textwidth]{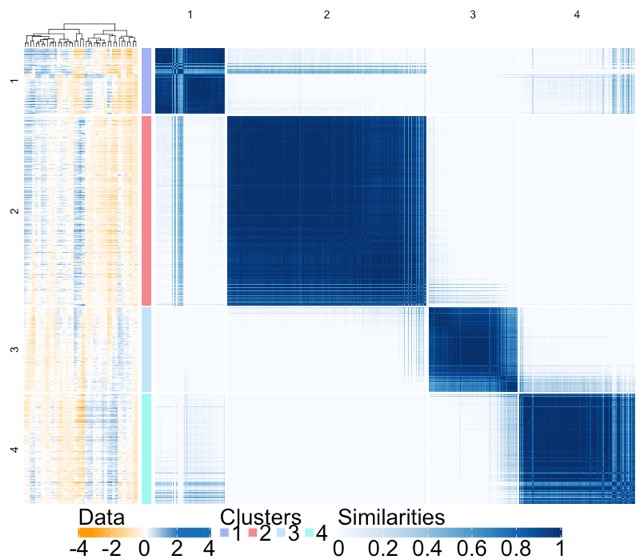}
		\caption{Expression data, PAM.\\
		Cophenetic correlation coefficient: 0.971.}
		\label{fig:consensus-matrix-expression}
	\end{subfigure}
   \begin{subfigure}[b]{.49\textwidth}
		\centering
		\includegraphics[width=\textwidth]{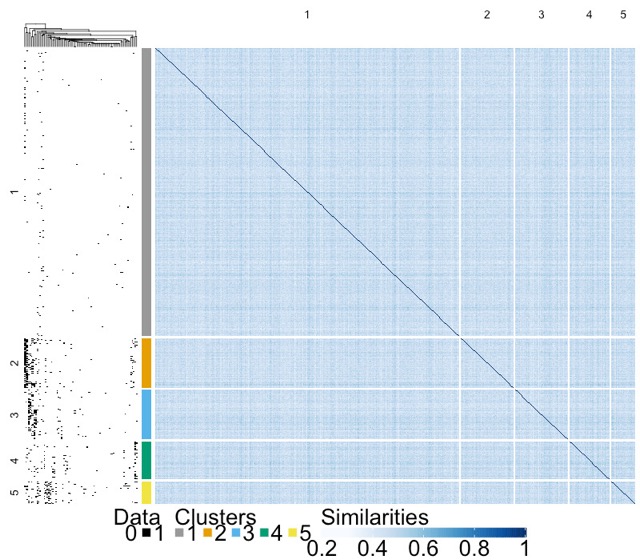}
		\caption{ChIP data, BHC.\\
		Cophenetic correlation coefficient: 0.103.}
		\label{fig:consensus-matrix-chip-bhc}
	\end{subfigure}
	\begin{subfigure}[b]{.49\textwidth}
		\centering
		\includegraphics[width=\textwidth]{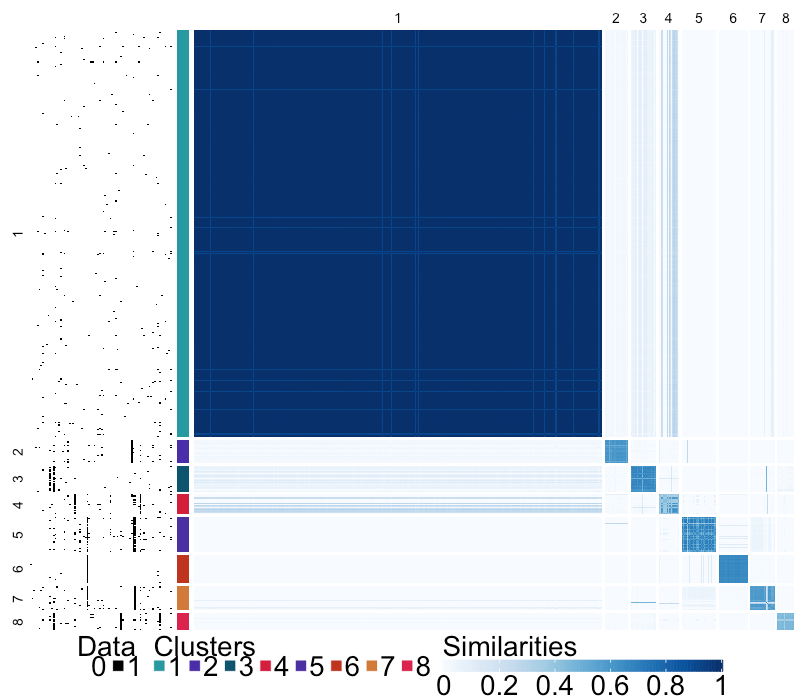}
		\caption{ChIP data, PAM.\\
		Cophenetic correlation coefficient: 0.996.}
		\label{fig:consensus-matrix-chip-pam}
	\end{subfigure}
	\begin{subfigure}[b]{.49\textwidth}
		\centering
		\includegraphics[width=\textwidth]{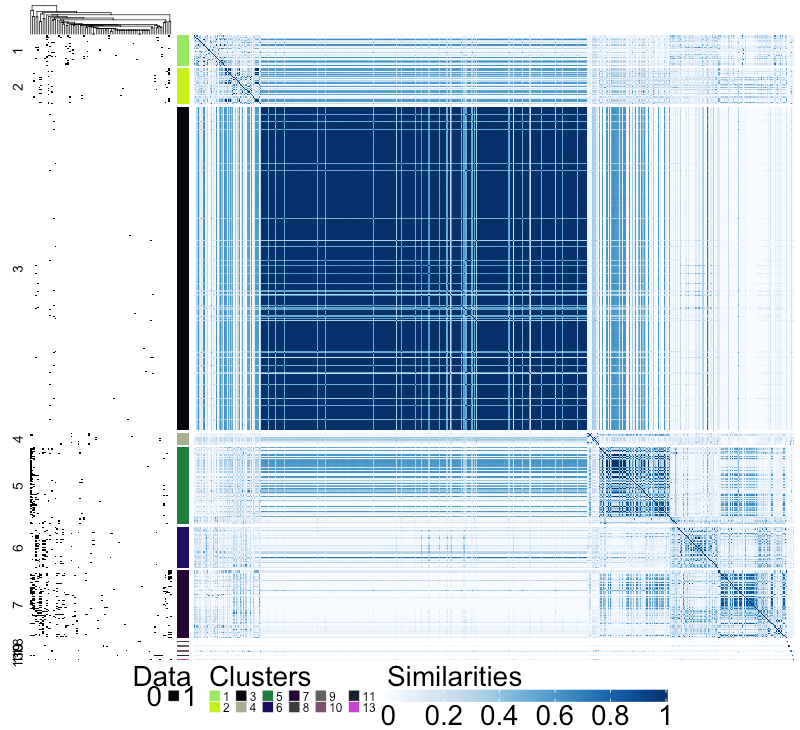}
		\caption{ChIP data, GBNP.\\
		Cophenetic correlation coefficient: 0.931.}
		\label{fig:consensus-matrix-chip-gbnp}
	\end{subfigure}
	\caption{Consensus matrices.}
	\label{fig:transcriptional-module-discovery-consensus-matrices}
\end{figure}

Figure \ref{fig:consensus-matrix-chip-bhc} shows the clusters found on all the data (on the left) as well as the consensus matrix obtained by applying BHC to 200 random subsamples of 95\% of the data. This shows that, while the clustering algorithm works well on the full dataset, different clustering structures are found in the data subsamples, giving a fuzzy similarity matrix. This is due to the fact that most clusters are very small, and are hard to identify when only a subset of the data is available. \textcolor{black}{The output of COCA obtained with this clustering algorithm is shown in Figure \ref{fig:transcriptonal-module-discovery-coca}, the output KLIC is shown in the main paper.} Higher weights are assigned on average to the expression data, with an average of 0.58.

\subsubsection{PAM with Gower's distance}
\label{sec:chip-data-pam}

Another clustering algorithm that could have been applied to this dataset is PAM with Gower's distance \citep{Gower1971general}. In this case, all variables are binary and therefore Gower's distance is equivalent to Jaccard's distance. For two multivariate binary observations $x_i$ and $x_j$, this is defined as one minus the Jaccard index:
\begin{equation}
J = \frac{M_{11}}{M_{01} + M_{01} + M_{11}},
\end{equation}
where $M_{11}$ is the number of variables where $x_i$ and $x_j$ both have value of $1$, $M_{01}$ is the number of variables where $x_i$ is 0 and $x_j$ is 1 and viceversa for $M_{01}$. This distance is particularly suited for this dataset because here the ones correspond to transcription factors that are believed with high confidence to be able to bind to the promoter region of the corresponding gene, whereas zeros are transcription factors for which we are not able to reject the hypothesis that they do not bind to that promoter region. Thus, in a sense, ones carry more information than zeros. 

The consensus matrix obtained by subsampling 200 times 95\% of the data is shown in Figure \ref{fig:consensus-matrix-chip-pam}, \textcolor{black}{the output of COCA and KLIC in Figures \ref{fig:transcriptonal-module-discovery-coca} and \ref{fig:transcriptional-module-discovery-klic} respectively}. 
Details on how the number of clusters was chosen are given in Section \ref{sec:number-of-clusters}. As usual, the number of clusters for KLIC and COCA was chosen in order to maximise the silhouette. KLIC selected $K=3$ and COCA $K=10$.
GOTO scores  for the clustering found with PAM algorithm and Gower's distance, as well as those given by KLIC and COCA for three and ten clusters are reported in Table \ref{table:goto-scores}.  Higher weights are assigned to the ChIP data, with an average of 0.78.

\subsubsection{Greedy Bayesian non-parametric clustering algorithm}
\label{sec:chip-data-gbnp}

The last clustering algorithm that we considered is a greedy approximation to the Gibbs sampling algorithm for Dirichlet process mixture models of \cite{Neal2000}. In the greedy version of the algorithm used here at each iteration cluster allocations are made in a deterministic fashion, assigning each observation to the cluster with highest probability, instead of sampling the cluster labels according to their conditional probabilities.

\begin{figure}
	\centering
	\captionsetup[subfigure]{justification=centering}
	\begin{subfigure}{.6\textwidth}
		\centering
		\includegraphics[width=\textwidth]{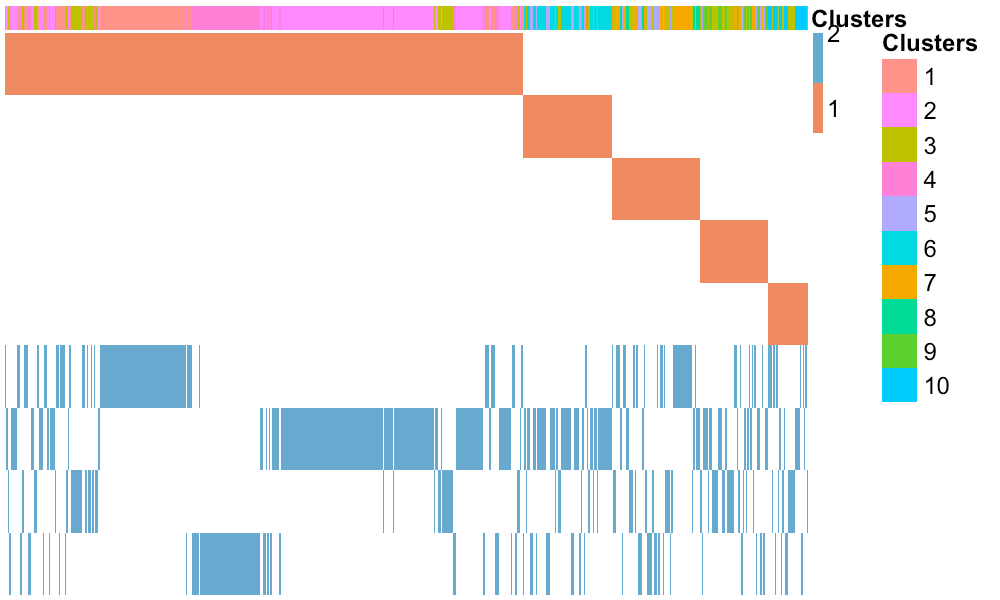}
		\caption{BHC}
		\label{fig:coca-bhc}
	\end{subfigure}
	\vspace{.5cm}
	
	\begin{subfigure}{.6\textwidth}
		\centering
		\includegraphics[width=\textwidth]{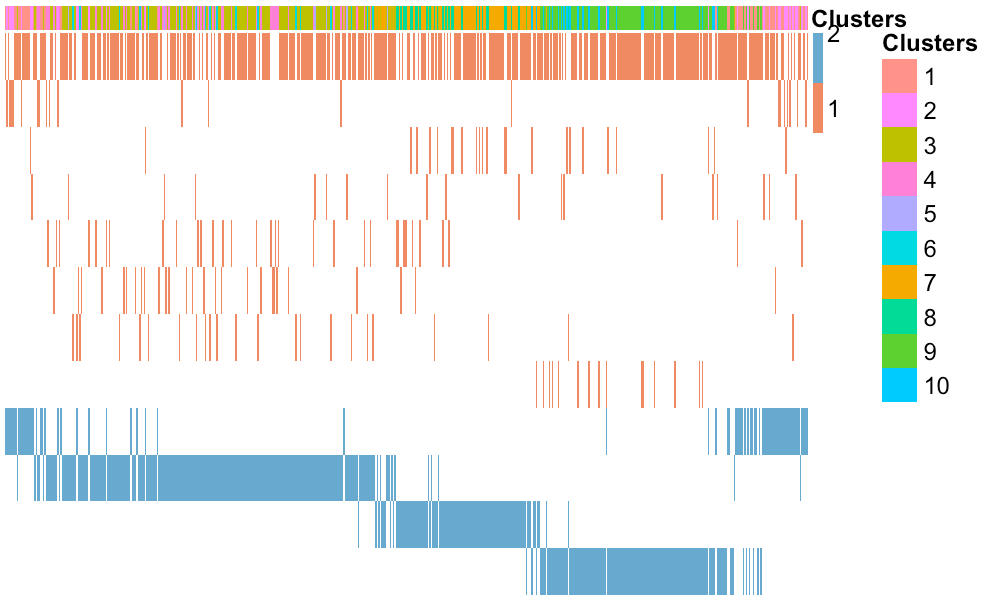}
		\caption{PAM with Gower's distance.}
		\label{fig:coca-pam}
	\end{subfigure}	
	\vspace{.5cm}
	
	\begin{subfigure}{.6\textwidth}
		\centering
		\includegraphics[width=\textwidth]{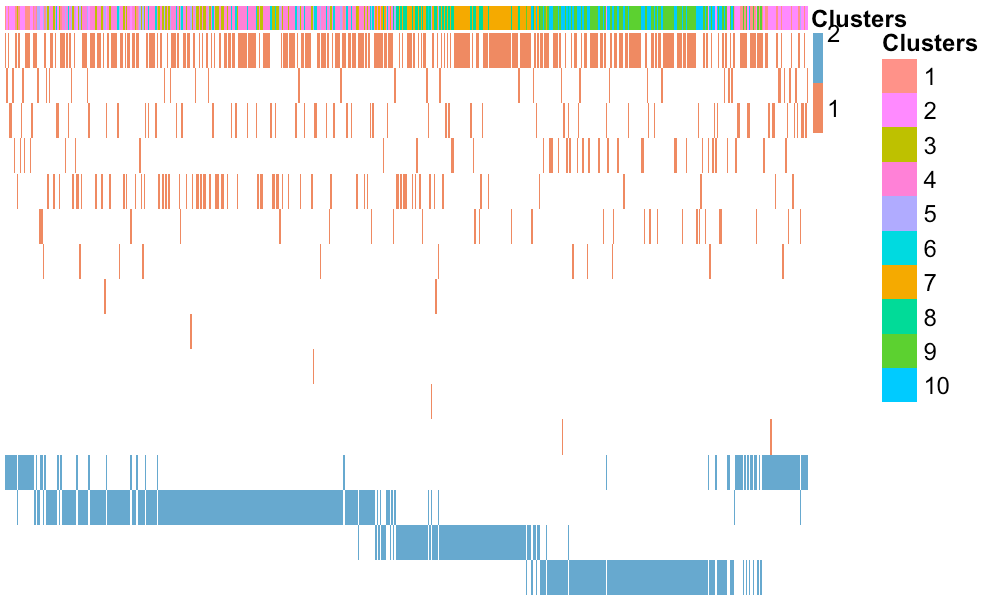}
		\caption{GBNP.}
		\label{fig:coca-gbnp}
	\end{subfigure}	
	
	\caption{Transcriptional module discovery. Output of COCA.}
	\label{fig:transcriptonal-module-discovery-coca}
\end{figure}

\begin{figure}
	\centering
	\captionsetup[subfigure]{justification=centering}
	\begin{subfigure}[b]{\textwidth}
		\centering
		\makebox[\textwidth][c]{\includegraphics[width=1.2\textwidth]{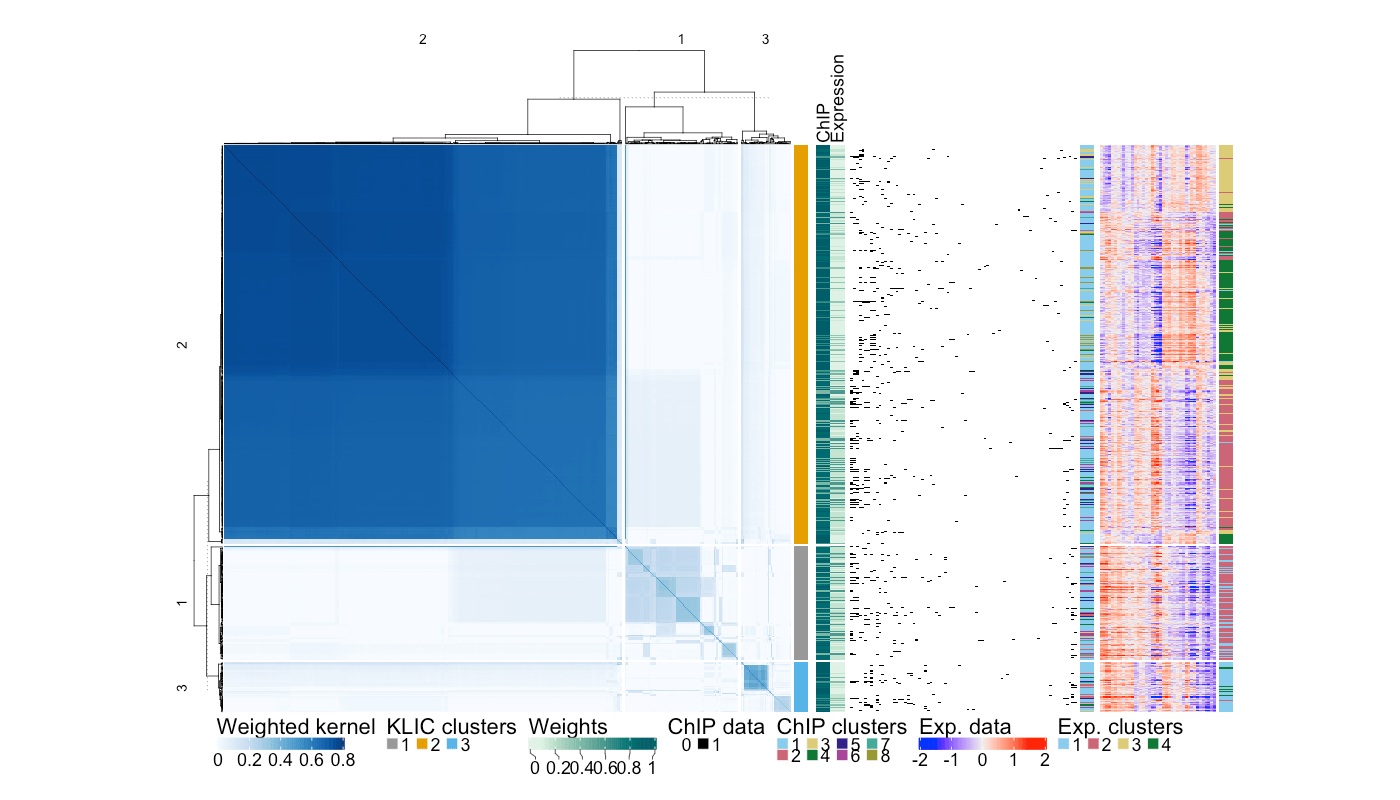}}
	\end{subfigure}
	\begin{subfigure}[b]{\textwidth}
		\centering
		\makebox[\textwidth][c]{\includegraphics[width=1.2\textwidth]{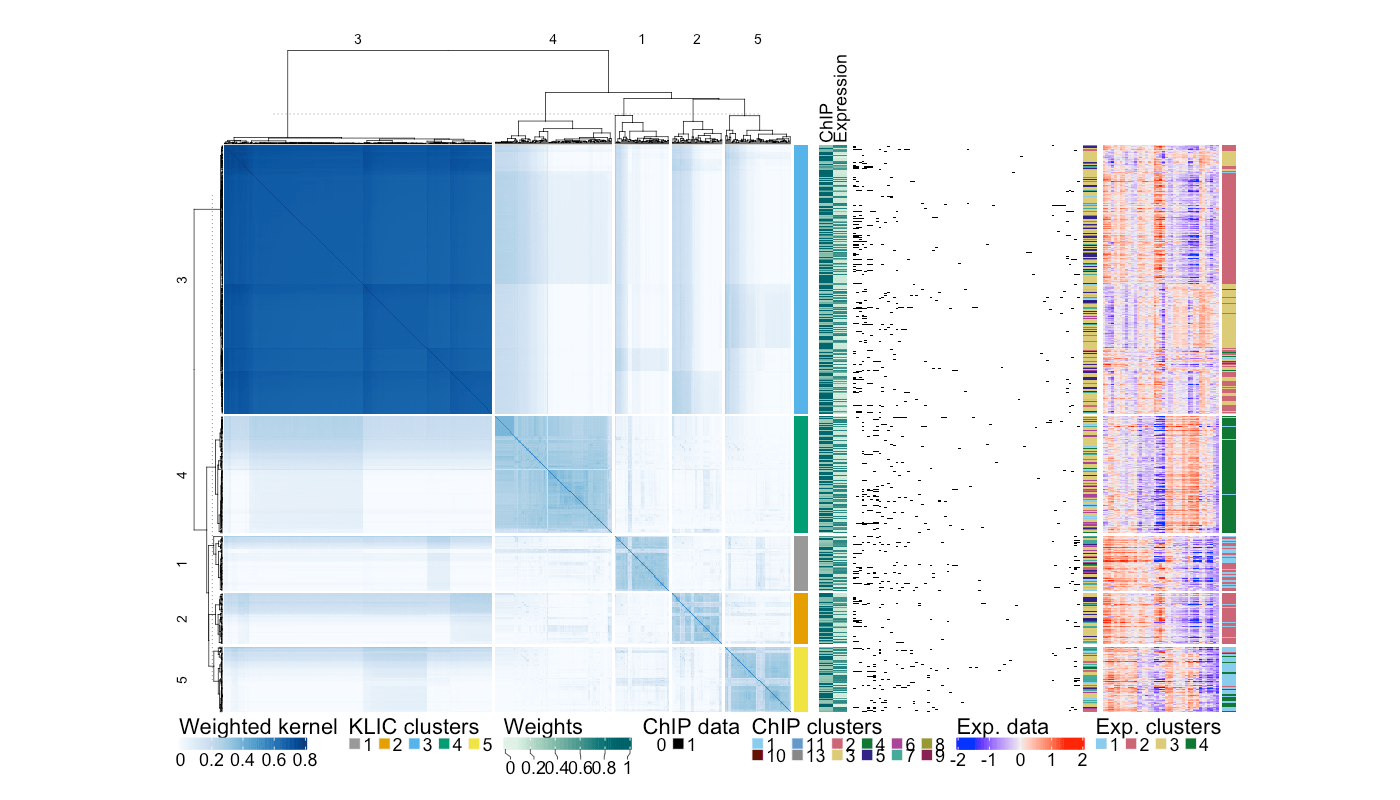}}
	\end{subfigure}	
	\caption{Transcriptional module discovery. Output of KLIC. PAM with Gower's distance (above) and GBNP (below).}
	\label{fig:transcriptional-module-discovery-klic}
\end{figure}

Figure \ref{fig:consensus-matrix-chip-gbnp}  shows the consensus matrix, \textcolor{black}{Figures \ref{fig:transcriptonal-module-discovery-coca} and \ref{fig:transcriptional-module-discovery-klic} show the output of COCA and KLIC respectively}. (Note that, for brevity, we refer to this method as ``GBNP'', which stands for Greedy Bayesian NonParametric algorithm.) Higher weights are assigned to the ChIP data points, with an average of 0.59.

\begin{table}
\centering
\begin{tabular}{ l l l c c c }
Clusters  & Dataset(s) & Algorithm & GOTO BP & GOTO MF & GOTO CC \\ 
\hline
4 & Expression  & PAM correlation & 6.1194 & 0.9075 & 8.4139 \\
8 & ChIP & PAM Gower's & 6.0872 & 0.8959 & 8.3261 \\
5 & ChIP  & BHC & 6.0020 & 0.9192 & 8.2886 \\
12 & ChIP & GBNP & 6.0192 & 0.9176 & 8.3664 \\
4 & ChIP+Expression & COCA (PAM + BHC) & 6.1194 & 0.9075 & 8.4139 \\
4 & ChIP+Expression & KLIC (PAM + BHC) & 6.1221 & 0.9074 & 8.4103 \\
10 & ChIP+Expression & COCA (PAM + BHC) & 6.2767 & 0.9347 & 8.5137 \\
10 & ChIP+Expression & KLIC (PAM + BHC) & 6.3240 & 0.9473 & 8.5310 \\
3 & ChIP+Expression & COCA (PAM + PAM) & 5.9609 & 0.8991 & 8.2780 \\
3 & ChIP+Expression & KLIC (PAM + PAM) & 5.9188 & 0.8915 & 8.1766 \\
10 & ChIP+Expression & COCA (PAM + PAM) &6.3429 & 0.9211 & 8.5126 \\
10 & ChIP+Expression & KLIC (PAM + PAM) & 6.3724 &  0.9094  & 8.4868 \\
5 & ChIP+Expression & COCA (PAM + GBNP) & 6.1298 & 0.9078 & 8.4218 \\
5 & ChIP+Expression & KLIC (PAM + GBNP) & 5.9629 &0.9108 & 8.3246 \\
10 & ChIP+Expression & COCA (PAM + GBNP) & 6.1605 & 0.9118 & 8.4796 \\
10 & ChIP+Expression & KLIC (PAM + GBNP) & 6.2277 &  0.9262 &  8.4814\\
\hline
\end{tabular}
\caption{Gene Ontology Term Overlap scores for different sets of data, clustering algorithms and numbers of clusters. ``BP'' stands for ``biological process'' ontology, ``MF'' for ``molecular function'', and ``CC'' for ``cellular component''.}
\label{table:goto-scores}
\end{table}

\clearpage

\subsection{Choice of the number of clusters}
\label{sec:number-of-clusters}

\begin{figure}[h]
	\centering
	\begin{subfigure}[b]{.48\textwidth}
		\centering
		\includegraphics[width=\textwidth]{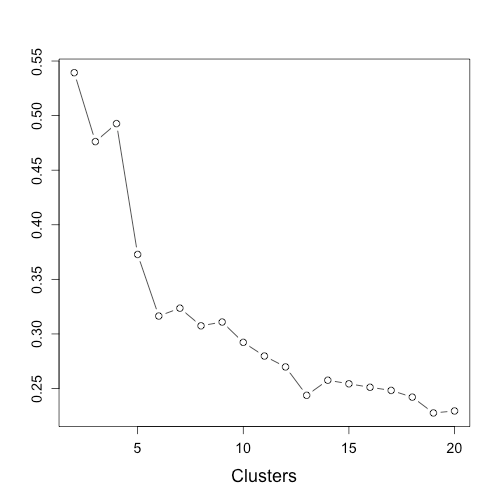}
		\caption{Average silhouette.}
	\end{subfigure}
	\begin{subfigure}[b]{.48\textwidth}
		\centering
		\includegraphics[width=\textwidth]{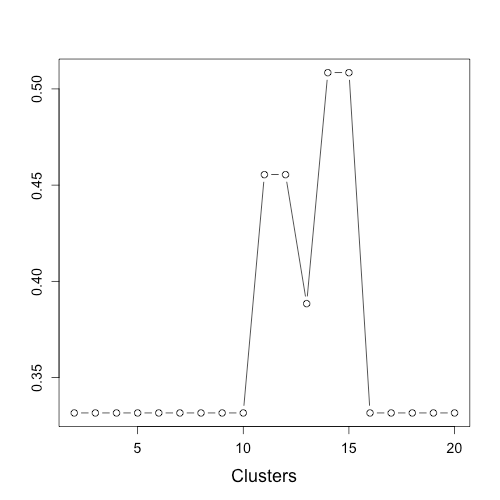}
		\caption{Widest gap.}
	\end{subfigure}
	\begin{subfigure}[b]{.48\textwidth}
		\centering
		\includegraphics[width=\textwidth]{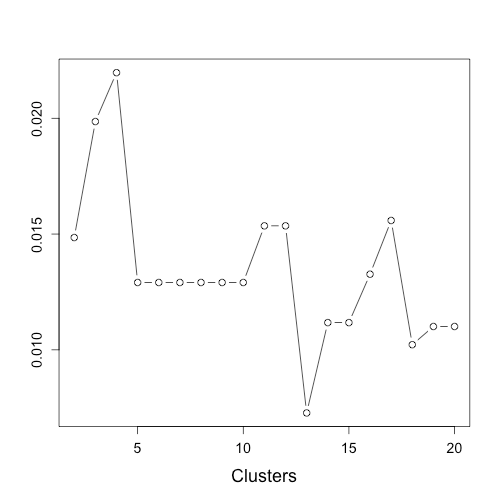}
		\caption{Dunn's index.}
	\end{subfigure}
	\begin{subfigure}[b]{.48\textwidth}
		\centering
		\includegraphics[width=\textwidth]{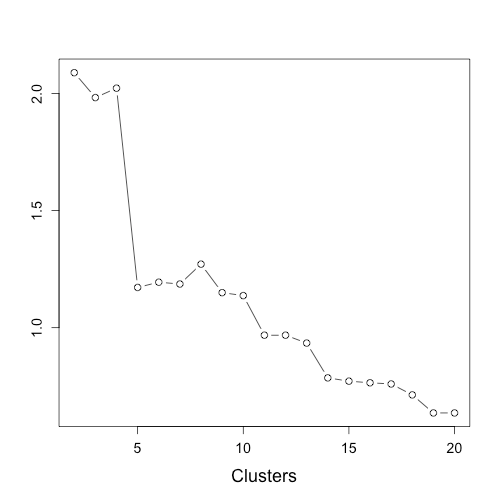}
		\caption{Dunn's modified index.}
	\end{subfigure}
	\caption{Expression data. Metrics used to choose the number of clusters.}
	\label{fig:expression-number-of-clusters}
\end{figure}

\begin{figure}
	\centering
	\begin{subfigure}[b]{.48\textwidth}
		\centering
		\includegraphics[width=\textwidth]{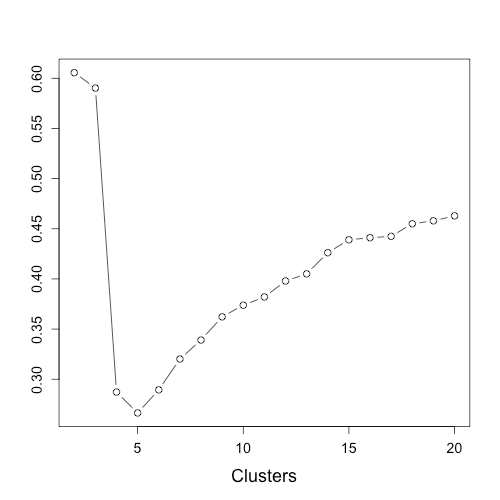}
		\caption{Average silhouette.}
	\end{subfigure}
	\begin{subfigure}[b]{.48\textwidth}
		\centering
		\includegraphics[width=\textwidth]{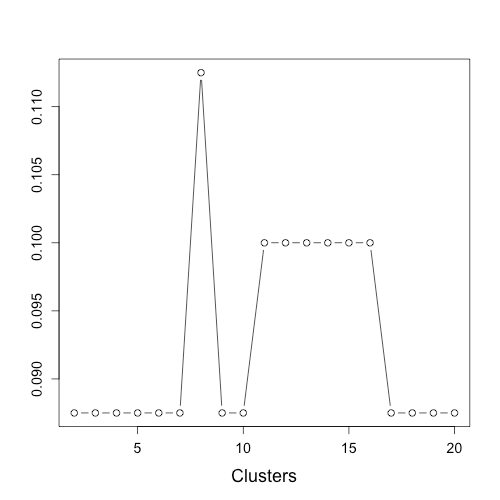}
		\caption{Widest gap.}
	\end{subfigure}
	\begin{subfigure}[b]{.48\textwidth}
		\centering
		\includegraphics[width=\textwidth]{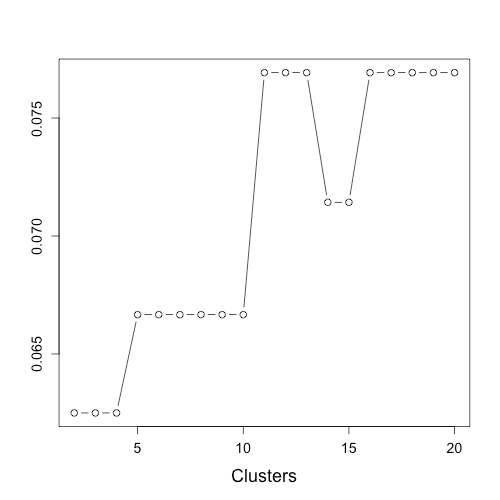}
		\caption{Dunn's index.}
	\end{subfigure}
	\begin{subfigure}[b]{.48\textwidth}
		\centering
		\includegraphics[width=\textwidth]{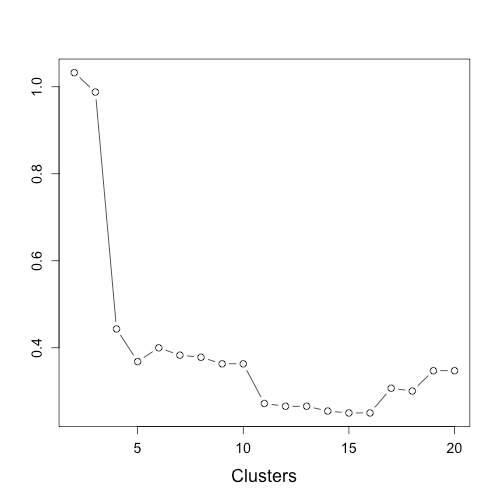}
		\caption{Dunn's modified index.}
	\end{subfigure}
	\caption{ChIP data. Metrics used to choose the number of clusters.}
	\label{fig:chip-number-of-clusters}
\end{figure}

In order to choose the number of clusters when using PAM, we considered multiple metrics: the average silhouette \citep{Rousseeuw1987}, the gap statistic \citep{Tibshirani2001estimating}, and the original and modified versions of Dunn's index \citep{Dunn1974well, Halkidi2001clustering}. We considered all number of clusters from two to 20. These are shown in Figures \ref{fig:expression-number-of-clusters} and \ref{fig:chip-number-of-clusters}. For the expression data, we chose four clusters since three of the chosen metrics have a peak at $K=4$. For the ChIP data, there is no consensus among the metrics, so we selected $K=8$ based on the gap metric.

\clearpage
\bibliographystyle{apalike}
\addcontentsline{toc}{section}{Bibliography}
\bibliography{supplement}
